\documentclass[10pt,twocolumn,letterpaper]{article}

\usepackage{wacv}
\usepackage{times}
\usepackage{epsfig}
\usepackage{color}
\usepackage{graphicx}
\usepackage{amsmath}
\usepackage{amssymb}

\usepackage{enumitem}
\usepackage{makecell}
\usepackage{multirow}
\usepackage{bbold}
\usepackage{adjustbox}
\usepackage{caption}
\usepackage{subcaption}
\usepackage{comment}
\usepackage{booktabs}

\usepackage{pifont}
\newcommand{\cmark}{\ding{51}}
\newcommand{\xmark}{\ding{55}}

\definecolor{ggreen}{RGB}{15,157,88}
\definecolor{gred}{RGB}{219,68,55}
\definecolor{black}{RGB}{0,0,0}

\wacvfinalcopy %

\ifwacvfinal
\usepackage[pagebackref=false,breaklinks=true,colorlinks,bookmarks=false]{hyperref}
\else
\usepackage[pagebackref=true,breaklinks=true,colorlinks,bookmarks=false]{hyperref}
\fi

\usepackage[capitalize]{cleveref}
\crefname{section}{Sec.}{Secs.}
\Crefname{section}{Section}{Sections}
\Crefname{table}{Table}{Tables}
\crefname{table}{Tab.}{Tabs.}

\begin{document}

\title{Intra-Batch Supervision for Panoptic Segmentation on High-Resolution Images}

\author{Daan de Geus \quad\quad Gijs Dubbelman\\
Eindhoven University of Technology\\
{\tt\small d.c.d.geus@tue.nl, g.dubbelman@tue.nl}
}

\maketitle
\thispagestyle{empty}
\begin{abstract}
Unified panoptic segmentation methods are achieving state-of-the-art results on several datasets. To achieve these results on high-resolution datasets, these methods apply crop-based training. In this work, we find that, although crop-based training is advantageous in general, it also has a harmful side-effect. Specifically, it limits the ability of unified networks to discriminate between large object instances, causing them to make predictions that are confused between multiple instances. To solve this, we propose Intra-Batch Supervision (IBS), which improves a network's ability to discriminate between instances by introducing additional supervision using multiple images from the same batch. We show that, with our IBS, we successfully address the confusion problem and consistently improve the performance of unified networks. For the high-resolution Cityscapes and Mapillary Vistas datasets, we achieve improvements of up to +2.5 on the Panoptic Quality for thing classes, and even more considerable gains of up to +5.8 on both the pixel accuracy and pixel precision, which we identify as better metrics to capture the confusion problem.

\end{abstract}

\section{Introduction}
\label{sec:introduction}

\begin{figure}[t]
\centering
\adjustbox{width=1.0\linewidth}{
\includegraphics[height=0.24\linewidth, trim={23.cm 6cm 0.cm 1.5cm},clip]{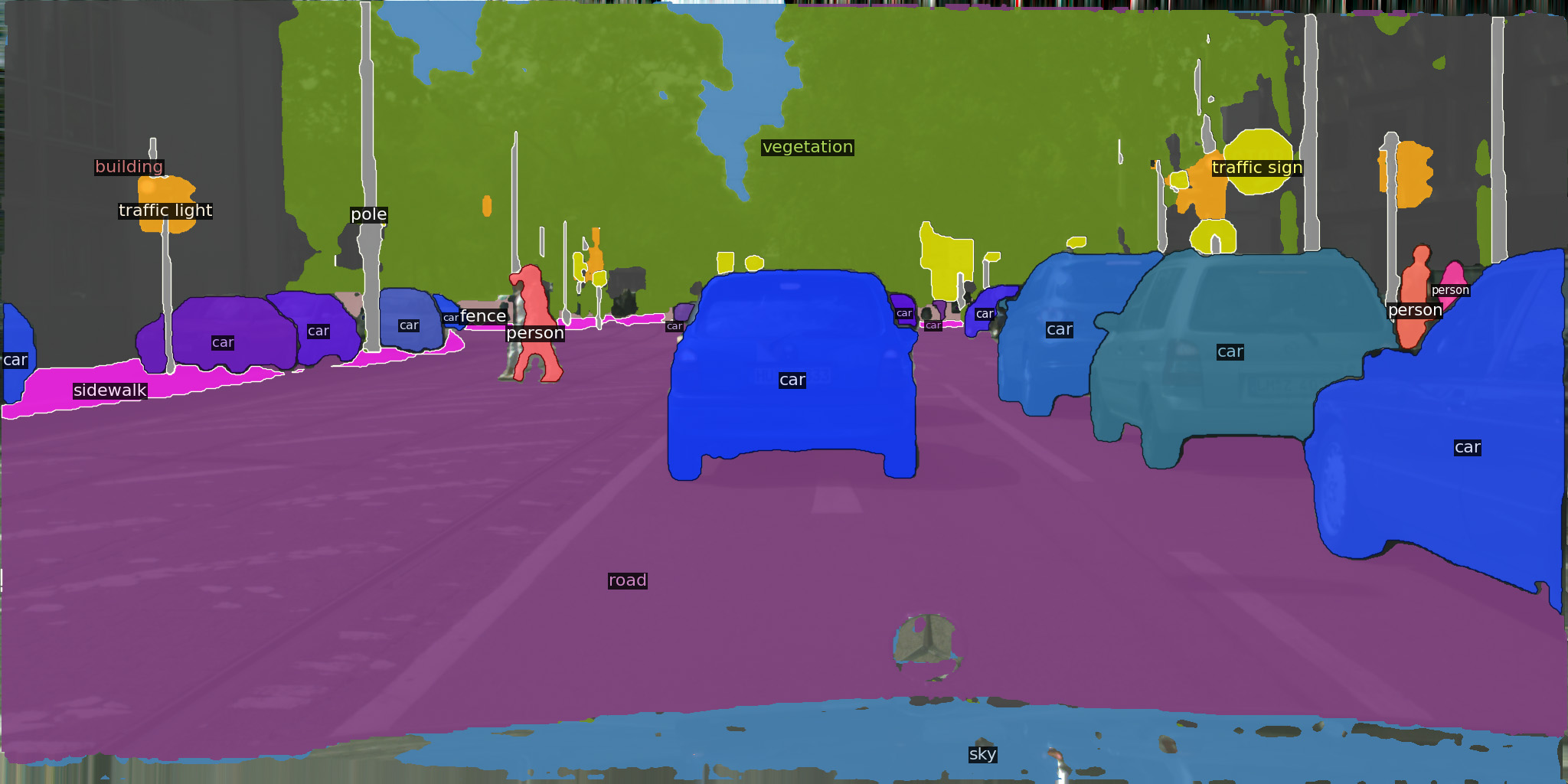}
\includegraphics[height=0.24\linewidth, trim={23.cm 6cm 0.cm 1.5cm},clip]{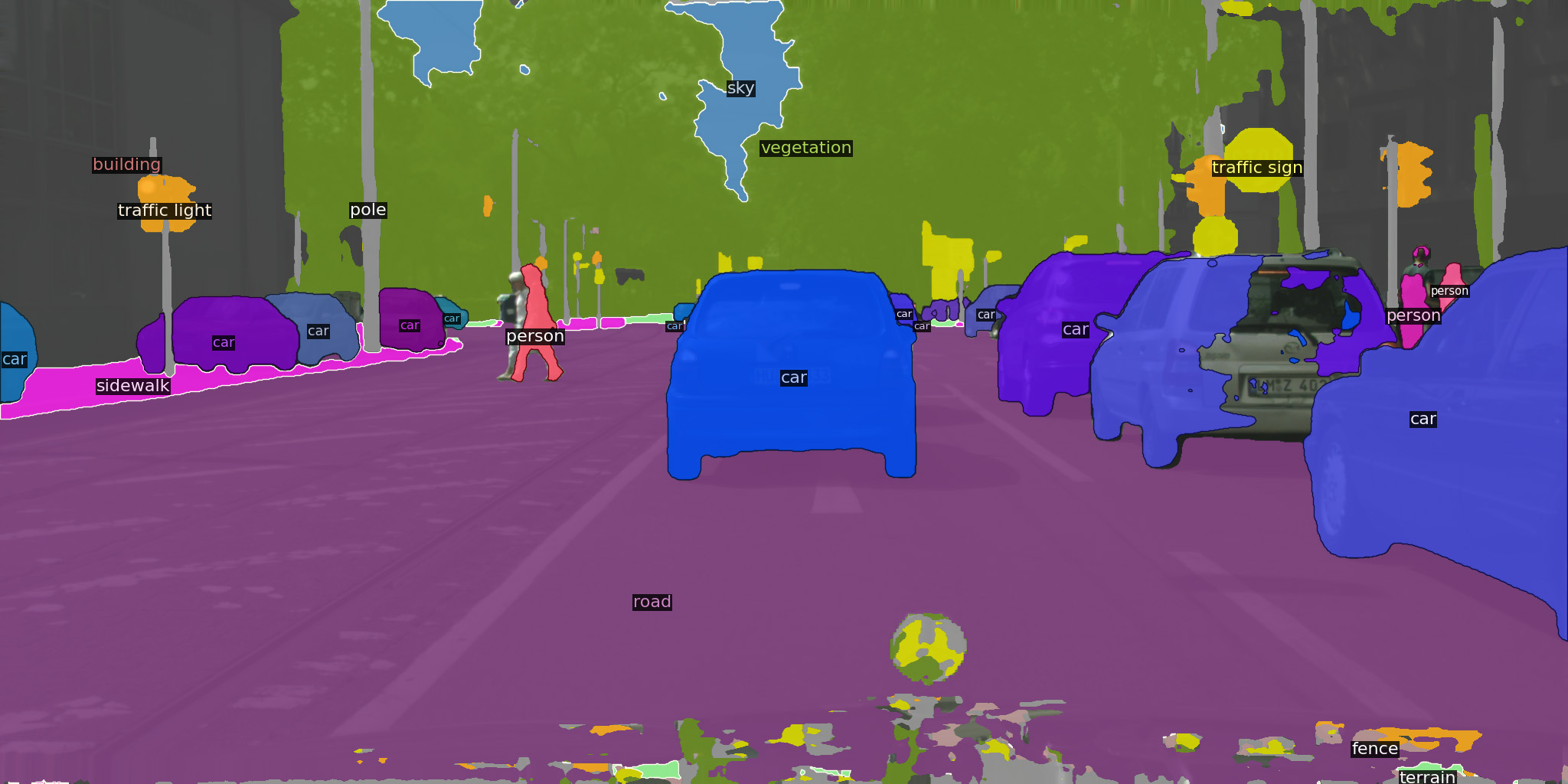}
\includegraphics[height=0.24\linewidth, trim={23.cm 6cm 0.cm 1.5cm},clip]{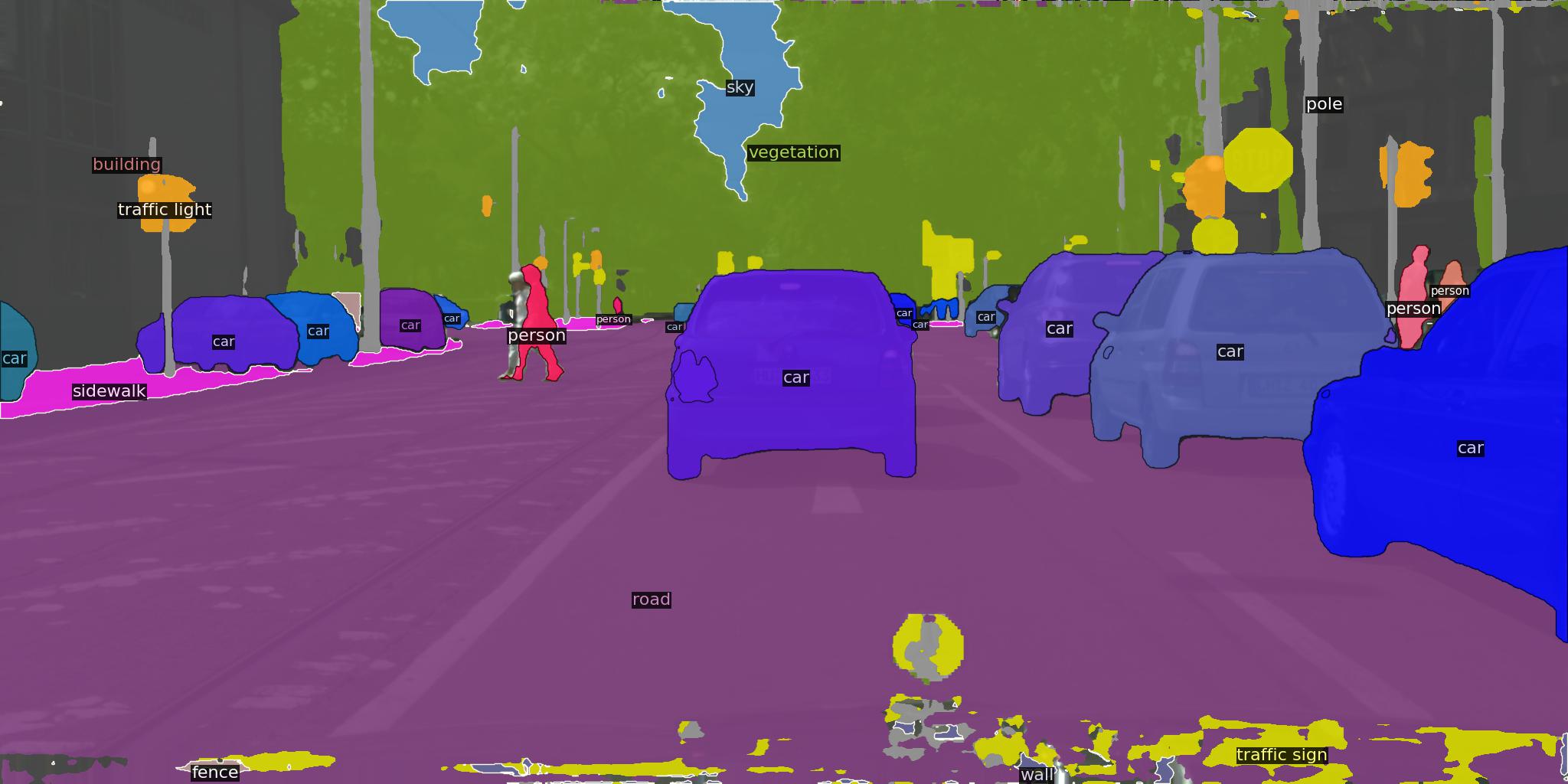}} \\
\adjustbox{width=1.0\linewidth}{
\includegraphics[height=0.24\linewidth, trim={12.cm 5cm 6.cm 2.5cm},clip]{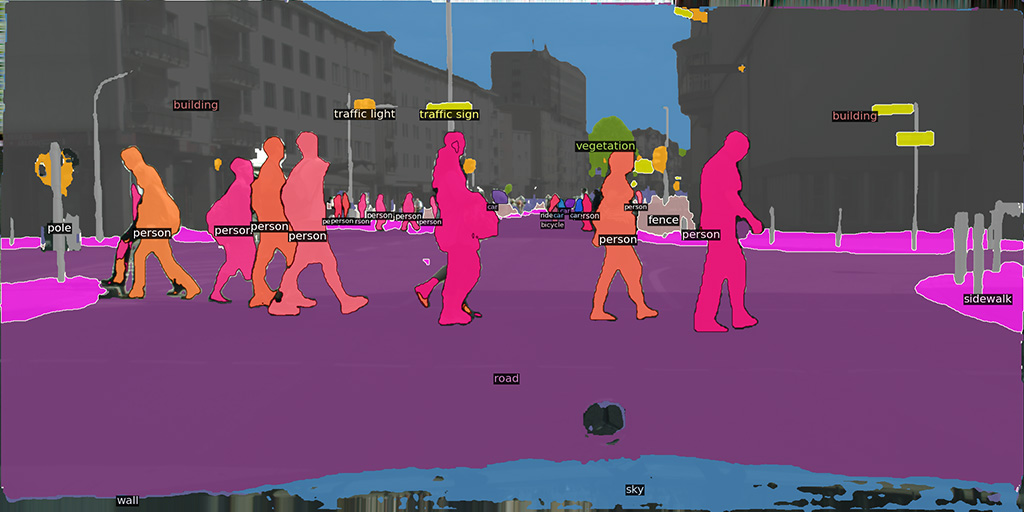}
\includegraphics[height=0.24\linewidth, trim={12.cm 5cm 6.cm 2.5cm},clip]{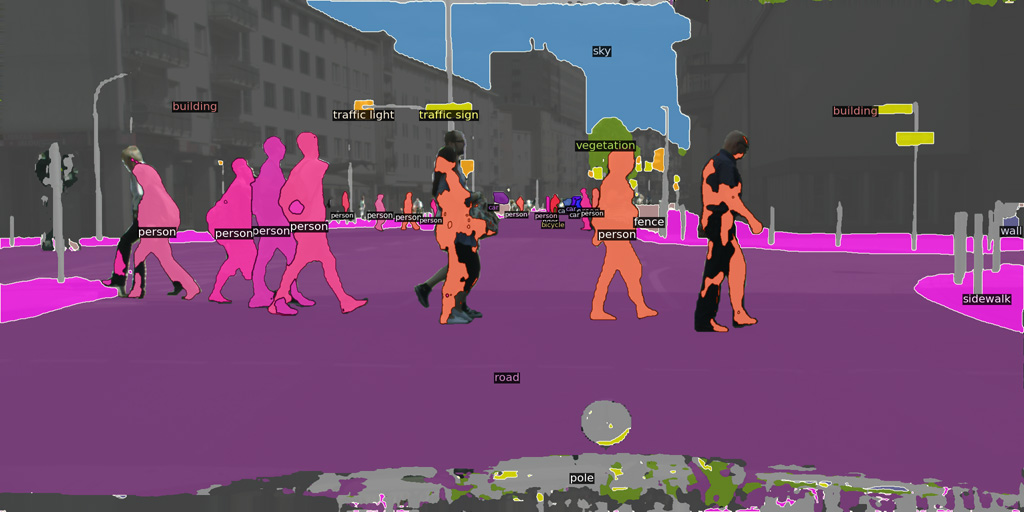}
\includegraphics[height=0.24\linewidth, trim={24.cm 10cm 12.cm 5cm},clip]{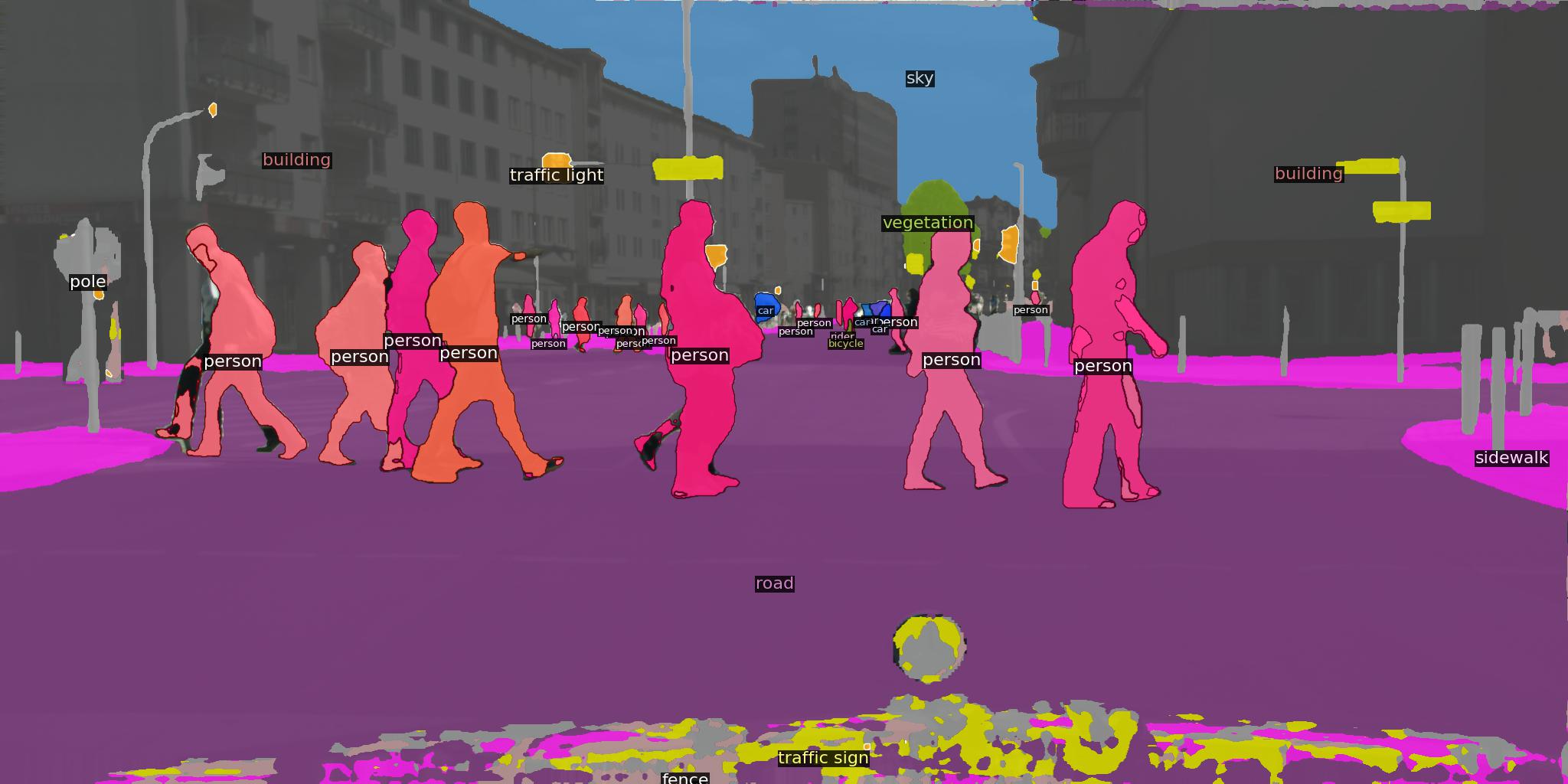}} \\

\adjustbox{width=1.0\linewidth}{
\includegraphics[height=0.3\linewidth]{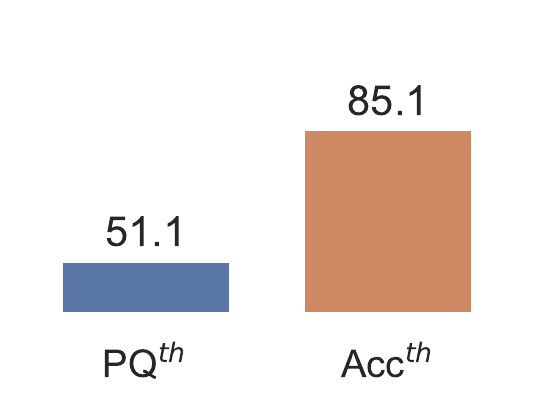}
\includegraphics[height=0.3\linewidth]{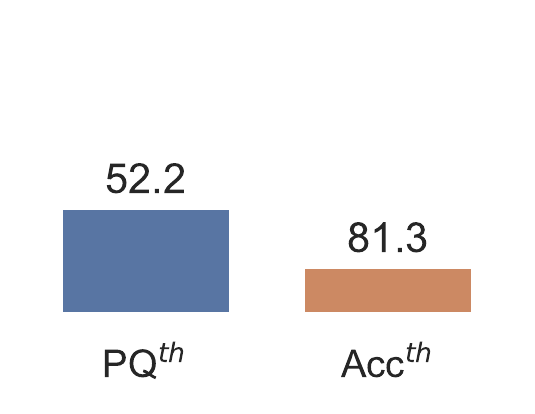}
\includegraphics[height=0.3\linewidth]{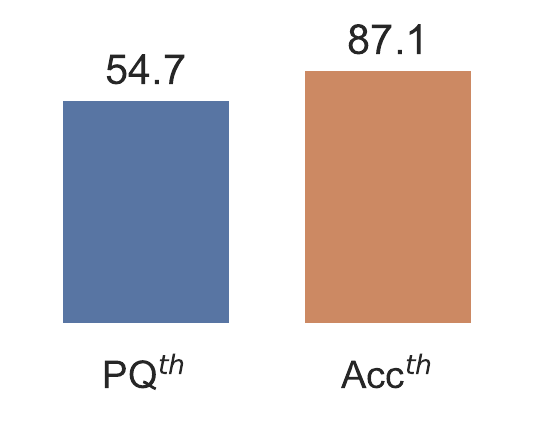}
}
\begin{subfigure}[b]{0.32\linewidth}

\caption{
Full-image \\ training
\centering}
\end{subfigure}
\begin{subfigure}[b]{0.32\linewidth}
\caption{Crop-based \\ training
\centering}
\end{subfigure}
\begin{subfigure}[b]{0.32\linewidth}
\caption{Crop-based \\ training + IBS (ours)
\centering}
\end{subfigure}
\vspace{-10pt}
\caption{\textbf{Results for different training strategies}. Crop-based training of unified networks, which is needed to achieve a good segmentation performance on high-resolution images, leads to inaccurate predictions for large \textit{thing} objects. We propose to solve this problem with Intra-Batch Supervision (IBS), and improve both the Panoptic Quality (PQ\textsuperscript{th})~\cite{kirillov2019ps} and pixel accuracy (Acc\textsuperscript{th}) for things.}
\vspace{-10pt}
\label{fig:eye_catcher}
\end{figure}

Panoptic segmentation~\cite{kirillov2019ps} is a scene understanding task that requires grouping all pixels of an image into segments with different semantic meaning, while also distinguishing between individual object instances. Where this task was first primarily tackled with multi-branch neural networks solving instance segmentation and semantic segmentation separately~\cite{kirillov2019panopticfpn,li2018tascnet,li2019aunet,mohan2020efficientps,xiong2019upsnet}, unified image segmentation approaches are now gaining popularity~\cite{cheng2021mask2former,cheng2021maskformer,li2021panopticfcn,wang2021maxdeeplab,zhang2021knet}. These unified approaches aim to use the same unified architecture for all types of image segmentation, and they outperform the previous state-of-the-art on various datasets. Despite their good performance, unified methods still suffer from a \textit{hitherto unidentified problem} when applied to high-resolution images, which is due to crop-based training (see \Cref{fig:eye_catcher}). In this work, we explore this problem, identify its cause, and we propose a method to solve it. 

Within panoptic segmentation, we distinguish between two types of classes: \textit{things} and \textit{stuff} \cite{kirillov2019ps}. Things are countable classes for which we can identify individual instances, such as car and person. For thing classes, panoptic segmentation requires segments for each unique instance. Stuff classes are uncountable, often amorphous regions, such as road and sky, for which panoptic segmentation requires just one segment per class. Recently, unified methods achieve state-of-the-art panoptic segmentation results by using a unified architecture to solve both stuff and thing segmentation. There are different variations to this approach, but the overall principle is the same: 1) it generates an \textit{embedding} vector for each stuff and thing segment, 2) it generates a tensor with \textit{features} for the entire image, and 3) it outputs a segmentation mask for each segment by taking the product of the embeddings and features.

\begin{figure}[t]
\centering
\adjustbox{width=\linewidth}{
\includegraphics[width=0.5\linewidth, trim={0cm 2.109375cm 0 0cm},clip]{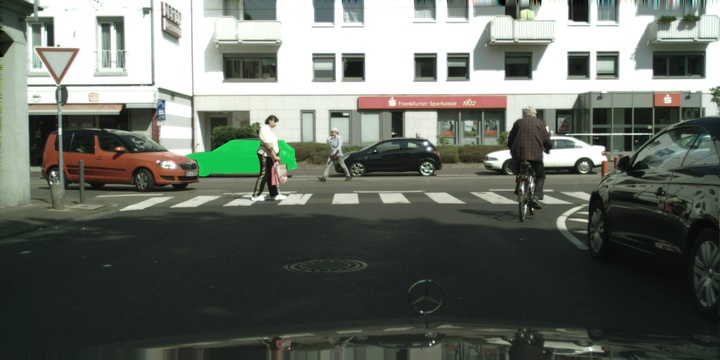}
\includegraphics[width=0.5\linewidth, trim={0cm 3cm 0 0cm},clip]{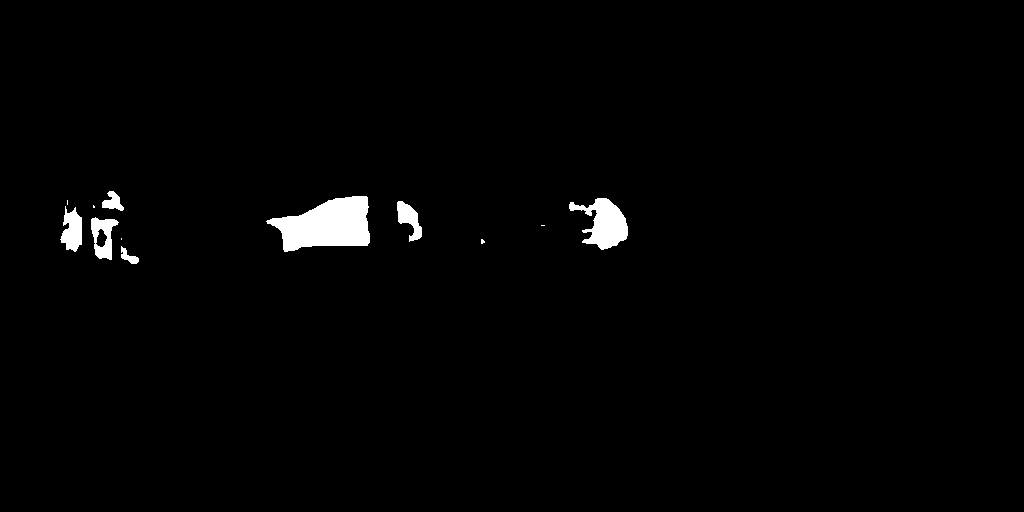}
}\\
\adjustbox{width=\linewidth}{
\includegraphics[width=0.5\linewidth, trim={0cm 2.109375cm 0 0cm},clip]{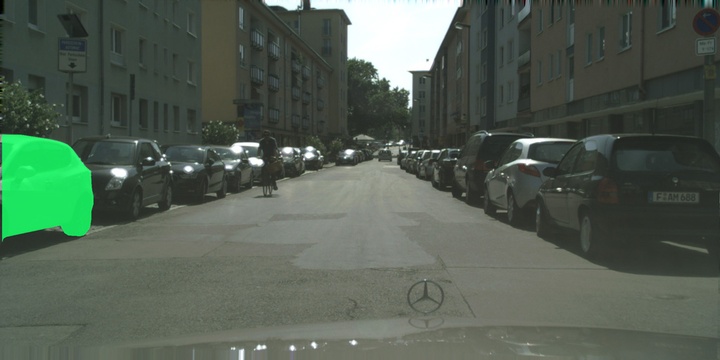}
\includegraphics[width=0.5\linewidth, trim={0cm 3cm 0 0cm},clip]{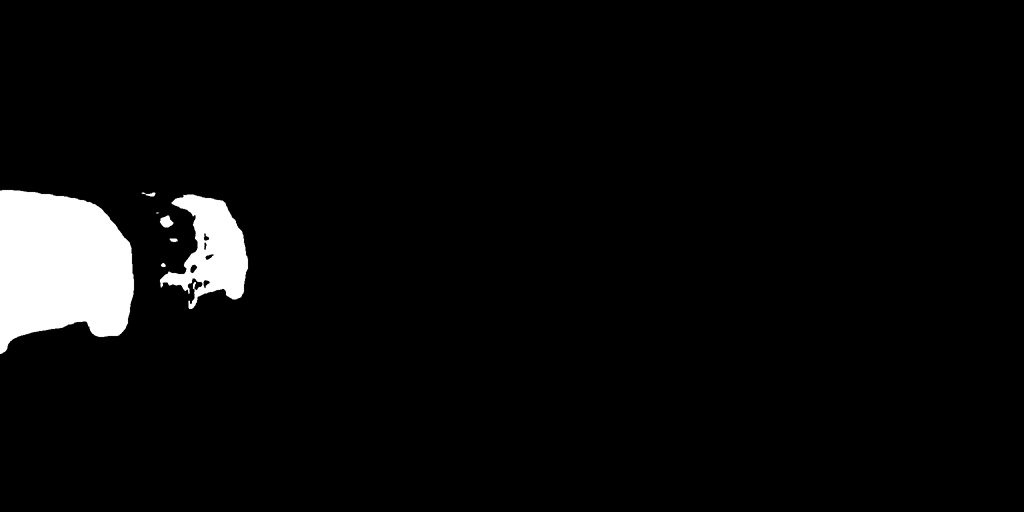}
}\\
\adjustbox{width=\linewidth}{
\includegraphics[width=0.5\linewidth, trim={0cm 2.109375cm 0 0cm},clip]{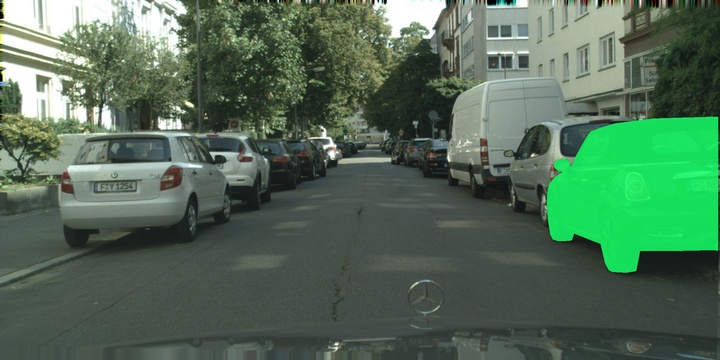}
\includegraphics[width=0.5\linewidth, trim={0cm 3cm 0 0cm},clip]{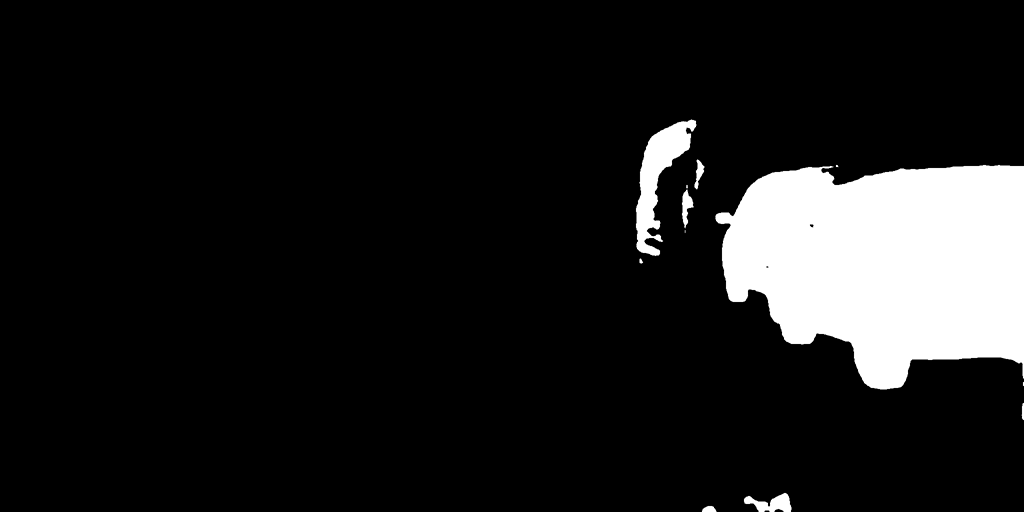}
}
\begin{subfigure}[b]{0.49\linewidth}
\caption{}
\end{subfigure}
\begin{subfigure}[b]{0.49\linewidth}
\caption{}
\end{subfigure}
\vspace{-15pt}
\caption{\textbf{Confusion between thing segments.} These are typical examples of the \textit{confusion} in predicted thing segments for Panoptic FCN~\cite{li2021panopticfcn} when trained on image crops. (a) Input image with ground-truth thing mask in green. (b) The corresponding prediction confused between instances.}
\label{fig:problem_examples}
\vspace{-5pt}
\end{figure}

\begin{figure}[t]
\centering
\includegraphics[width=\linewidth]{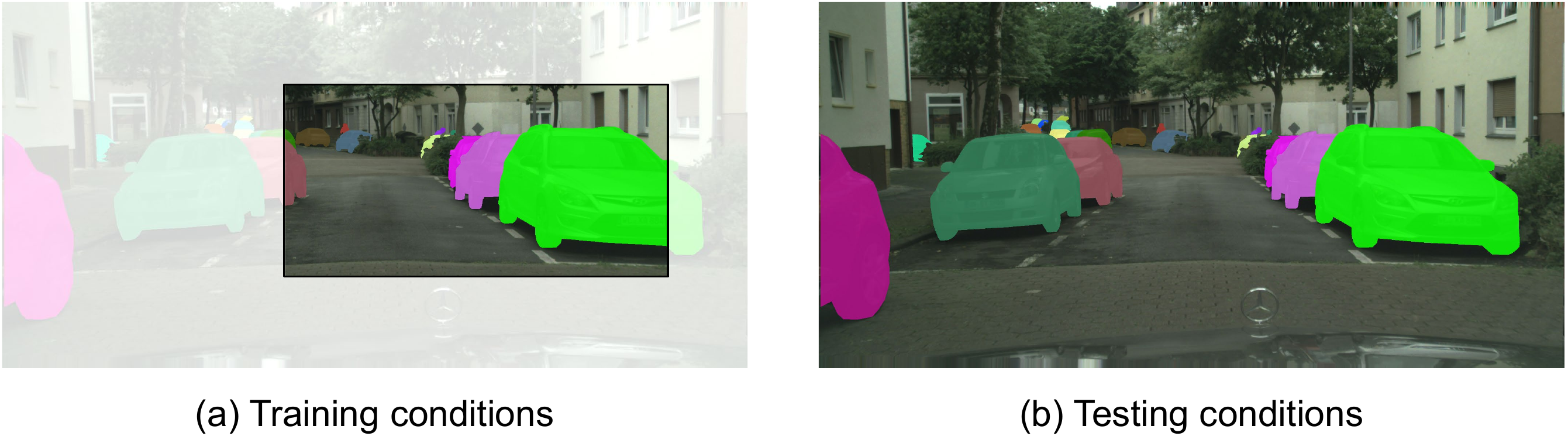}
\vspace{-12pt}
\caption{\textbf{Crop-based training and full-image inference}. (a) When training with image crops, the network can only learn to discriminate between instances within a small part of the input image. (b) During testing, which happens on full images, the network needs to discriminate between many more instances. As it has not learnt to do so, this will lead to confusion between instances (see \Cref{fig:problem_examples}).}
\label{fig:training_testing_conditions}
\vspace{-5pt}
\end{figure}

Despite the improvements made by these unified approaches, they still suffer from a problem that –- to the best of our knowledge -- has not been reported in literature. Specifically, when trained and tested on high-resolution images, their predictions for individual thing segments often overlap multiple thing instances, especially for easily recognizable large objects, see \Cref{fig:problem_examples}. In this work, we find out that this problem is caused by crop-based training, which is customary and often necessary for panoptic segmentation methods to be able to train on high-resolution images and achieve state-of-the-art results~\cite{porzi2021allscales}. Specifically, unified networks generate thing segmentation masks by taking the product of the \textit{embedding} for those segments, and the overall \textit{features}. By supervising these masks during training, the network learns to generate embeddings and features that are sufficiently discriminative to generate an accurate, unique mask for each thing instance. However, when training on crops, the network sees considerably fewer instances in an image than during testing, as illustrated in \Cref{fig:training_testing_conditions}. Because the network is trained to distinguish only a relatively small number of objects, it is not sufficiently discriminative for the large number of objects encountered during testing. This causes the problem visualized in \Cref{fig:problem_examples}, where segmentation masks are confused between multiple thing instances. We further explore this confusion problem in \Cref{sec:problem_description}. We show that the standard Panoptic Quality metric~\cite{kirillov2019ps} for panoptic segmentation does not adequately reflect the confusion problem because it is biased towards small segments and infrequent classes. We find that the pixel accuracy and pixel precision metrics  better reflect the confusion problem, and we propose to use these metrics together with the PQ.

To solve the confusion problem caused by crop-based training, the network should receive more adequate supervision for thing segments.
To be generally applicable, this should be achieved without using additional data or changing the network architecture. In this paper, we propose to achieve this by teaching the network to distinguish instances not only from other instances in the same image crop, but also from instances in other image crops in the same training batch. Specifically, we take the product of the thing embeddings from one crop and the features from another crop, and supervise the outputs to teach the network to be more discriminative.
We call our improvement \textit{Intra-Batch Supervision} (IBS), as we apply additional supervision to the product of thing embeddings and features of different image crops in the same batch. Because IBS only helps if there is meaningful supervision, we also introduce a new crop sampling strategy that makes sure that IBS always occurs between crops for which the confusion problem is likely to occur. In \Cref{sec:method}, we motivate and describe IBS in detail.

In \Cref{sec:results}, we extensively evaluate our proposed solution on multiple networks and datasets, using the experimental setup as described in~\Cref{sec:experiments}. We find that our IBS significantly decreases the confusion between thing segments, as intended, resulting in a consistent increase of the PQ for thing classes, and a significant improvement on the pixel accuracy. This is also shown in \Cref{fig:eye_catcher}. Further discussions and conclusions are provided in \Cref{sec:discussion,sec:conclusion}. All code is available through \url{https://ddegeus.github.io/intra-batch-supervision/}.

In summary, we make the following contributions:
\setlist{nolistsep}
\begin{itemize}
    \item We demonstrate that crop-based training of unified segmentation networks has a harmful side-effect that leads to predictions with \textit{confusion} between thing segments, and we show that standard metrics for panoptic segmentation do not capture this problem.
    \item We propose \textit{intra-batch supervision} (IBS) to solve the confusion problem, which introduces additional supervision to allow the network to better discriminate between thing instances.
    \item We show the effectiveness of IBS on state-of-the-art unified panoptic segmentation and instance segmentation networks, on two different datasets.
\end{itemize}

\section{Related work}
\label{sec:related_work}
\paragraph{Multi-branch panoptic segmentation.}
Following the formalization of the panoptic segmentation task~\cite{kirillov2019ps}, many deep neural network architectures have been proposed to solve it. As panoptic segmentation is essentially a combination of semantic segmentation and instance segmentation, most proposed methods initially used neural networks that make separate predictions for these two tasks~\cite{degeus2018jsisnet,kirillov2019panopticfpn,li2019aunet,liu2019end,mohan2020efficientps,porzi2019seamless,xiong2019upsnet}. Concretely, this means that there is a joint backbone which extracts features from the image, followed by two separate branches: 1) one that predicts instance masks and classes for \textit{thing} classes, often based on Mask R-CNN~\cite{he2017maskrcnn}, and 2) another that applies per-pixel classification to segment the remaining \textit{stuff} classes. Over the past years, many variations and improvements to this multi-branch framework have been introduced~\cite{cheng2020pandeeplab,degeus2020fpsnet,gao2019ssap,li2020unifying,sofiiuk2019adaptis,yang2019deeperlab,yang2020sognet}, but all these networks use separate, task-specific architectural components for instance and semantic segmentation.

\paragraph{Unified panoptic segmentation.}
Recently, there has been an effort to develop universal or \textit{unified} approaches for the panoptic segmentation task. The proposed unified networks~\cite{cheng2021mask2former,cheng2021maskformer,li2021panopticfcn,wang2021maxdeeplab,zhang2021knet} treat all segments, \ie, stuff and things, as equally as possible. Although there are clear differences between the unified methods (\eg, fully-convolutional~\cite{li2021panopticfcn} vs. transformer-based~\cite{cheng2021mask2former,cheng2021maskformer}), the overall principle is similar: images are segmented by taking the product of per-segment \textit{embeddings} and the overall pixel-level \textit{features} from the images. This principle is similar to an approach used by recent instance segmentation methods~\cite{tian2020condinst,wang2020solov2}, but they do not apply stuff segmentation. By following this approach, the panoptic networks do not need to use multiple, specialized branches, and achieve state-of-the-art results. In this work, we show that despite the state-of-the-art performance, these models still make specific inaccurate predictions when applied on high-resolution images, and we propose a method to solve it.

\paragraph{Crop-based training.}
Many panoptic segmentation networks use crop-based training when applied to high-resolution datasets~\cite{cheng2020pandeeplab,cheng2021mask2former,kirillov2019panopticfpn,li2021panopticfcn,mohan2020efficientps}, and Porzi \etal~\cite{porzi2021allscales} show that crop-based training on high-resolution images outperforms full-image training, \eg, because it allows for more stable training and facilitates identifying small objects. At the same time, Porzi~\etal also highlight that crop-based training has negative side-effects for multi-branch panoptic segmentation methods that apply box-based detection. They show that 1) bounding boxes are truncated due to cropping, and 2) large images are underrepresented in cropped images, which harms the segmentation performance. To solve this, they propose a new loss and a new image sampling strategy, respectively. 
In this work, we find that crop-based training is also advantageous for unified panoptic segmentation networks, but that it has a specific negative side-effect, which does not apply for multi-branch methods, and is not solved by the solutions proposed by Porzi \etal~\cite{porzi2021allscales}. Therefore, this problem requires a new solution, which we propose and call \textit{intra-batch supervision}.

\section{Confusion problem analysis}
\label{sec:problem_description}
When testing unified panoptic segmentation networks on high-resolution images, we observe a large number of inaccurately predicted segmentation masks for things. Specifically, these networks predict masks that overlap \textit{multiple} thing instances, usually from the same semantic class, as if they are \textit{confused}. Examples of individual confused thing predictions are shown in \Cref{fig:problem_examples}, and the effect on the final panoptic segmentation result is visualized in \Cref{fig:results_main}. We hypothesize that this problem is caused by the way these networks are trained, \ie, with image crops. In this section, we explore the \textit{confusion} problem in more detail.

\subsection{Unified panoptic segmentation}
\label{sec:problem_description:unified_networks}

Unified panoptic segmentation methods~\cite{cheng2021mask2former,cheng2021maskformer,li2021panopticfcn,wang2021maxdeeplab,zhang2021knet} treat thing and stuff segments as equally as possible. This means that the output masks for all thing and stuff segments are generated in a unified way, using the same operations. 

The existing unified panoptic segmentation methods use the same high-level principle. In one part of the network, they generate \textit{embeddings} $E \in \mathbb{R}^{{N}\times{C}}$ for all $N$ \textit{thing} and \textit{stuff} segments in the input image, where $C$ is the embedding dimension. These embeddings should encode the properties of the segment they are representing. In another part of the network, they generate a single tensor of \textit{features} $F \in \mathbb{R}^{{C}\times{H}\times{W}}$, where $H$, $W$ are the height and width of the image. These features should represent the content of each pixel as well as possible. Together, these per-segment embeddings and the per-image, pixel-level features are used to generate the panoptic segmentation output. Specifically, the output segmentation masks $M \in \mathbb{R}^{{N}\times{H}\times{W}}$ are the result of a matrix multiplication between embeddings $E$ and features $F$, followed by a sigmoid activation. This process is visualized for thing segments in~\Cref{fig:overview} (top). To get the output in the panoptic segmentation format, each predicted mask should also have a corresponding class prediction. Each unified method treats this problem differently, \eg, MaskFormer~\cite{cheng2021maskformer} applies a linear layer to the embeddings to predict the classes, and Panoptic FCN~\cite{li2021panopticfcn} already predicts a class for each embedding earlier in the network.

\subsection{Effect of crop-based training}
\label{sec:problem_description:crop_training}
Panoptic segmentation requires that there is a \textit{unique} segmentation mask for each thing instance in an input image. In unified networks, the output mask for each thing instance is the product of the per-segment \textit{embedding} $E$ and per-image \textit{features} $F$. Therefore, to get a unique mask for each instance, it is important that both of these components are sufficiently discriminative, \ie, each per-segment embedding should be clearly distinguishable from the other generated embeddings, and the features at the pixels where an instance is present should be sufficiently different from the features at pixels with other instances. If this were not the case, the matrix multiplication between $E$ and $F$ would result in segmentation masks that have high activations at the pixels of more than one instance, \ie, \textit{confusion} between instances.

When training on full-resolution images, the network is able to learn embeddings and features that are sufficiently discriminative, as it sees many instances of different sizes and classes at different locations, which are representative of the situations it could encounter during testing. When training on image crops, the situation is different. Specifically, the `field of view' of the network is limited. Instead of seeing full-resolution images, representative of those that it will encounter during testing, it will see only a fraction of the input images (see \Cref{fig:training_testing_conditions}). Consequently, the network sees significantly fewer instances during training than during testing. This hinders the network's ability to discriminate between many different instances, as it only learns to be discriminative among the small set of instances in a crop. We expect that this lack of sufficient `negative' supervision causes the network to output \textit{thing} segmentation masks with confusion between different instances, when applied to full-resolution images during testing. Because only few large segments fit in a crop, the network sees few large segments during training, so we expect that this problem mainly affects these large segments.

\begin{table}[t]
\centering
\adjustbox{width=\linewidth}{
\setlength{\tabcolsep}{6pt} %
    \begin{tabular}{lcccccc}
    \toprule
    Network & Crop & PQ$\uparrow$ & PQ\textsuperscript{th}$\uparrow$ & PQ\textsuperscript{st}$\uparrow$ & Acc\textsuperscript{th}$\uparrow$ & Prec\textsuperscript{th}$\uparrow$\\
    \midrule
    \multicolumn{7}{c}{Cityscapes \textit{val}} \\
    \midrule
    Panoptic FCN~\cite{li2021panopticfcn} & \xmark & 57.2 & 51.1 & 61.7 & \textbf{85.1} & \textbf{91.4} \\
    Panoptic FCN~\cite{li2021panopticfcn} & \cmark & \textbf{59.5} & \textbf{52.2} & \textbf{64.8} & 81.3 & 86.8 \\
    \midrule
    \multicolumn{7}{c}{Mapillary Vistas \textit{validation}} \\
    \midrule
    Panoptic FCN~\cite{li2021panopticfcn} & \xmark & 32.7 & 29.8 & 36.6 & 73.8 & \textbf{80.8} \\
    Panoptic FCN~\cite{li2021panopticfcn} & \cmark & \textbf{35.5} & \textbf{31.8} & \textbf{40.3} & \textbf{74.7} & 77.9 \\
    \bottomrule
    \end{tabular}
}
\caption{\textbf{Full-image vs. crop-based training.}}
\label{tab:confusion_problem}
\vspace{-10pt}
\end{table}

\subsection{Impact on metrics}
\label{sec:problem_description:quantification}

To verify our hypothesis that crop-based training causes the \textit{confusion} problem, and to measure the effect of the problem on the overall panoptic segmentation results, we train unified networks on (a) full images and (b) image crops, and report the performance in \Cref{tab:confusion_problem}. As a first metric, we use the Panoptic Quality (PQ)~\cite{kirillov2019ps}, the standard metric for panoptic segmentation, defined as

\noindent
\begin{equation}
\label{eq:pq_metric}
    \textrm{PQ} = \frac{1}{|C|} \sum_{c=1}^{|C|}\frac{\sum_{(p,g)\in{TP_c}}{\textrm{IoU}{(p,g)}}}{{|TP_c|}+\frac{1}{2}{|FP_c|}+\frac{1}{2}{|FN_c|}} ,
\end{equation}
where $|C|$ is the set of all classes, $p$ and $g$ are the predictions and ground-truth segments, and $TP_c$, $FP_c$ and $FN_c$ are the sets of true positives, false positives and false negatives per class $c$, respectively. True positives are matched prediction/ground-truth pairs that have an intersection-over-union (IoU) larger than 0.5. False positives are predictions not matched to a ground-truth segment, and false negatives are ground-truth segments not matched to a prediction. By design, this metric treats all classes equally, as it takes the average over all classes, and it treats all segments equally, as they all count as a single $TP$, $FP$ or $FP$.

\begin{figure*}[t]
\centering
\includegraphics[width=0.8\linewidth]{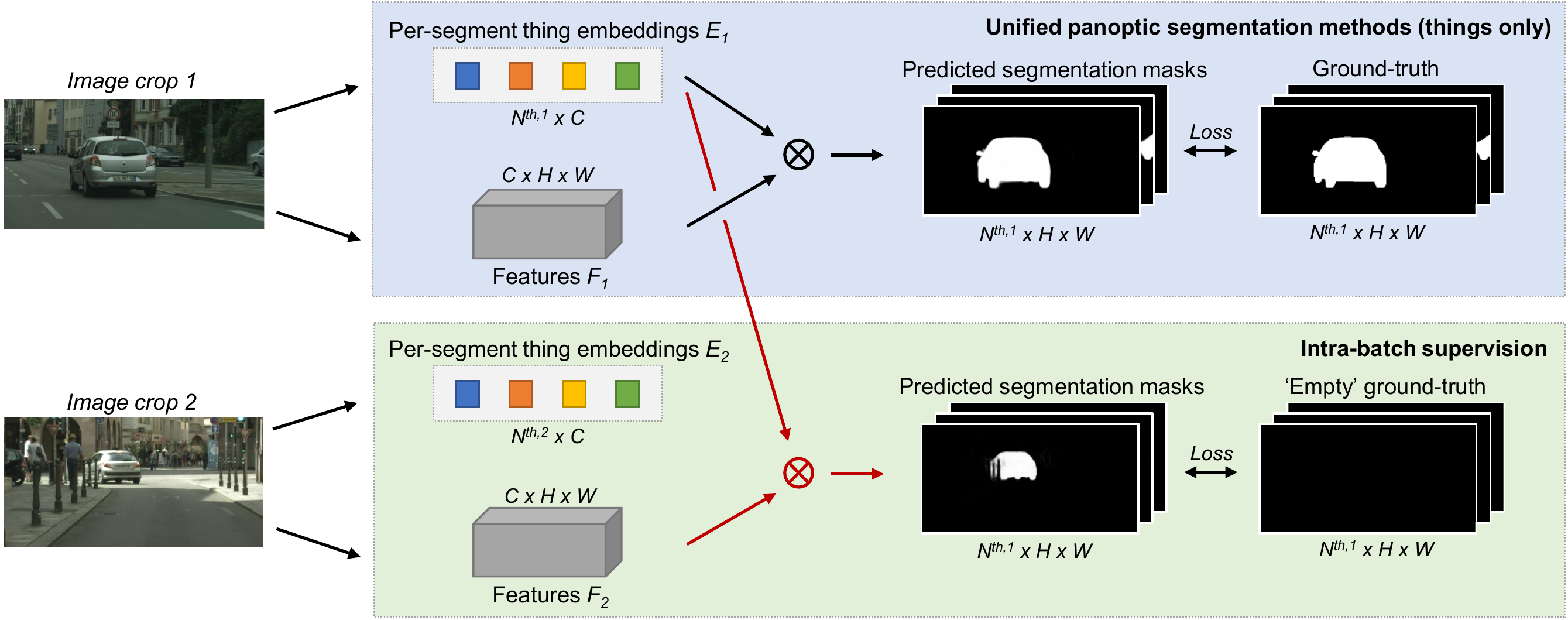}
\caption{\textbf{Intra-Batch Supervision}. We take the product of thing embeddings $E$ from one image crop and features $F$ from another image crop in the batch (indicated in red), and as the thing instances belonging to the embeddings are not present in the other image crop, the result should be `empty' predicted masks. By applying a loss to these predictions, we teach the network to generate more discriminative embeddings and features, leading to predicted masks with less \textit{confusion}.}
\label{fig:overview}
\vspace{-5pt}
\end{figure*}

From the results in \Cref{tab:confusion_problem}, we observe that the PQ increases when we use crop-based training instead of full-image training, for both thing (PQ\textsuperscript{th}) and stuff classes (PQ\textsuperscript{st}). This is as expected, because crop-based training facilitates more stable training and allows for better identification of small and underrepresented segments~\cite{porzi2021allscales}, thereby improving the quality of the predictions. However, from the qualitative results in \Cref{fig:eye_catcher,fig:problem_examples}, it is also obvious that there are many inaccurate predictions for large objects when using crop-based training, which can be harmful for downstream tasks. Because this problem predominantly occurs for large segments and classes with multiple segments per image, and the PQ treats all segments and classes equally, the segments affected by the confusion problem have only a limited impact on the PQ, compared to the many small segments and the segments from other classes.
Therefore, we also report the performance on the previously commonly-used pixel accuracy metric~\cite{hurtado2022pa}, and the related pixel precision. Because we are mainly interested in the thing classes, we report on these metrics for things only, with Acc\textsuperscript{th} and Prec\textsuperscript{th}, respectively. They are defined as
\begin{equation}
\label{acc_metric}
    \textrm{Acc\textsuperscript{th}} = \frac{|TP^{px}_{th}|}{N^{px}_{th,g}};
\end{equation}
\begin{equation}
\label{prec_metric}
    \textrm{Prec\textsuperscript{th}} = \frac{|TP^{px}_{th}|}{|TP^{px}_{th}|+|FP^{px}_{th}|},
\end{equation}
where $TP^{px}_{th}$ and $FP^{px}_{th}$ are the sets of true positive and false positive thing pixels, respectively, and $N^{px}_{th,g}$ is the total number of ground-truth thing pixels. We count a pixel as a true positive if 1) it belongs to a predicted thing segment that has an $\textrm{IoU} > 0.5$ with a ground-truth segment of the same class (like for the PQ), and 2) if this pixel intersects with the matched ground-truth segment. In other words, the Acc\textsuperscript{th} captures how many ground-truth thing pixels have been correctly predicted, and the Prec\textsuperscript{th} captures how many thing predictions are correct. Ideally, panoptic segmentation methods should score well on \textit{all three} metrics, the PQ, the Acc\textsuperscript{th} and the Prec\textsuperscript{th}, as this means that A) they work well on average, over all classes and segment sizes, and B) they make correct predictions for most pixels in an image.

When looking at the results for crop-based training in \Cref{tab:confusion_problem} in terms of the Acc\textsuperscript{th}, we observe a drop or much more modest increase than for the PQ, and we observe a consistent, significant drop for the Prec\textsuperscript{th}. This shows that, as expected and as observed in \Cref{fig:problem_examples}, crop-based training has a harmful side-effect that causes a large number of pixels to be predicted inaccurately. This side-effect can clearly cause severe problems for downstream tasks such as autonomous driving because of its impact on the identification of large, nearby objects, and thus it needs to be solved.

\section{Intra-Batch Supervision}
\label{sec:method}
In \Cref{sec:problem_description}, we analyzed the harmful side-effect of crop-based training that causes \textit{confused predictions}, and hypothesized that it is caused by a lack of sufficient supervision during training. We aim to solve this problem by proposing a method that is easily and generally applicable, so it should not require a) changes to the network architecture, or b) the use of any additional training data. To achieve this, we propose to address this problem for unified networks by introducing additional supervision, using the \textit{embeddings} and \textit{features} that are already being generated within each training batch. We call our proposed solution \textit{Intra-Batch Supervision} (IBS). Because our method is only applied during training, it does not add any computational complexity during testing.

\subsection{Additional supervision}
\label{sec:method:ibs}
The principle of IBS is straightforward: take the product of the per-segment \textit{embeddings} $E$ of one image crop and the per-image-crop \textit{features} $F$ of another crop in the batch, and apply a loss to the resulting segmentation masks. Under the assumption that each object instance only occurs in one crop, this means that for embeddings $E$ belonging to thing segments, matrix multiplication with features $F$ from another crop should always result in empty segmentation masks, as the instances represented by $E$ are not present in the other crop. When this is supervised, the network will learn to generate embeddings and features that are more unique to the segments they correspond to. By applying IBS to random \textit{embedding}-\textit{feature} combinations in the batch, the network receives supervision for a large number of instances which are outside its `field-of-view', solving the supervision problem caused by crop-based training. The key principle of IBS is visualized in~\Cref{fig:overview}. Note that we expect that the assumption on each thing instance occurring only in a single crop per batch is violated so infrequently, due to the large dataset size and the random batch and crop sampling methods, that its effect is negligible.

\paragraph{Predictions.}
The first step of intra-batch supervision consists of making the intra-batch predictions, between the \textit{embeddings} and \textit{features} of different image crops. Consider a training scenario where the batch size is $B$. With a given unified method, we generate embeddings ${E_1}, ..., {E_B}$ and features ${F_1}, ..., {F_B}$ for $B$ image crops. Then, for each \textit{source} crop, \ie, the crop from which we take the embeddings, we randomly pick another crop from the batch to be the \textit{target} crop, from which we pick the features. For the embeddings $E$ from each source crop $i$ we then have corresponding features $F$ from target crop $tgt_{i}$, resulting in $E$,$F$ pairs $\{{E_1},{F_{tgt_{1}}}\},...,\{{E_B},{F_{tgt_{B}}}\}$. Subsequently, each of these pairs are subjected to matrix multiplication and sigmoid activation to generate the segmentation masks of segments from source crop $i$ in target crop $tgt_{i}$, \ie, $M_{i\rightarrow{tgt_{i}}}$.

For example, when $B=3$, we could randomly pick pairs $\{{E_1},{F_{2}}\}$, $\{{E_2},{F_{3}}\}$, $\{{E_3},{F_{1}}\}$. We then apply a matrix multiplication between $E_1$ and $F_2$, to generate the segmentation masks for embeddings from crop 1 on crop 2, $M_{1\rightarrow{2}}$. The example of pair $\{{E_1},{F_{2}}\}$ is visualized in~\Cref{fig:overview}.

\paragraph{Supervision.}
The activations in segmentation mask $M_{i\rightarrow{tgt_{i}}}$ represent the presence of segments from source crop $i$ in target crop $tgt_{i}$. Under the assumption that each instance is present in only one crop, this means that the predictions for thing segment embeddings should always be zero. If it is not zero, this means that the network confuses a thing instance from $i$ with another instance in $tgt_i$, which is the \textit{confusion} we are trying to solve. Therefore, we supervise these segmentation masks in such a way that the network learns to output \textit{empty} segmentation masks for all things in $M_{i\rightarrow{tgt_{i}}}$, creating unique embeddings and features for each instance. As we only want to apply the loss to the masks that indicate thing instances, we identify and retrieve the masks generated by embeddings $E$ for which there is an associated \textit{thing} ground-truth segment. For the segmentation masks of the $N^{th,i}$ things in the source crop $i$, $M^{th}_{i\rightarrow{tgt_{i}}} \in \mathbb{R}^{{N^{th,i}}\times{H}\times{W}}$, we generate an artificial segmentation ground-truth $S^{th}$, which has the shape of $M^{th}_{i\rightarrow{tgt_{i}}}$, but is zero at all pixels. We supervise this by applying the focal loss~\cite{lin2017focal} to the generated ground-truth and mask logits, to force the predictions to zero. We choose the focal loss because it is designed for data with large class imbalances, and we expect the majority of the predictions to be zero.

\subsection{Crop sampling}
For intra-batch supervision to work, it is critical that the supervision that the network receives is meaningful. When solving the confusion problem, supervision is meaningful when the \textit{intra-batch predictions} $M_{i\rightarrow{tgt_{i}}}$ contain positive, `confused' masks. When this is the case, the embeddings from image A activate features in image B, meaning that they are not sufficiently discriminative, and we can apply a loss to counter this. As we observe that confusion mainly happens between thing instances of the same class, we expect that meaningful supervision mostly occurs when both crops contain objects from the same class. When images are sampled and cropped randomly, it is unlikely that this is always the case, especially when the images are very large and when there are many classes in the dataset. 

Therefore, we propose a new crop sampling strategy, loosely inspired by the Instance Scale-Uniform Sampling (ISUS) strategy presented by Porzi \etal~\cite{porzi2021allscales}. Where they use a uniform sampling strategy to make sure that the network sees segments of all classes and sizes at a somewhat consistent rate, we want to sample the crops in such a way that there are always two crops that contain a segment of the same class and a similar size. Specifically, we 1) sample a class, 2) pick two images that contain a segment of this class, 3) pick one random ground-truth segment from that class from each image, 4) resize the images so that these segments have a similar size, and 5) crop the images in such a way that the crops contain the sampled ground-truth segments. We then make sure that IBS is applied between the crops that contain different segments of the same class. This way, we aim to accomplish meaningful supervision as often as possible, enhancing the effect of IBS.

\section{Experimental setup}
\label{sec:experiments}
We conduct several experiments to show the effectiveness of our IBS approach to solve the confusion problem. Concretely, we use state-of-the-art, unified panoptic segmentation methods Panoptic FCN~\cite{li2021panopticfcn} and Mask2Former~\cite{cheng2021mask2former}, and evaluate them with and without IBS, on Cityscapes and Mapillary Vistas, which are \textit{de facto} benchmarks for high-resolution panoptic segmentation. We use the evaluation metrics described in \Cref{sec:problem_description:quantification}.

\subsection{Datasets}
\label{sec:experiments:datasets}
We train and evaluate on two datasets with high-resolution street scene images: Cityscapes~\cite{cordts2016cityscapes} and Mapillary Vistas~\cite{neuhold2017mapillary}. The Cityscapes dataset contains 5k images, captured in and around Germany. All images have a resolution of $1024\times2048$ pixels, and there are panoptic labels for 19 classes (8 \textit{thing}, 11 \textit{stuff}). We train on the \textit{train} set (2975 images) and evaluate on the \textit{val} set (500 images). Mapillary Vistas~\cite{neuhold2017mapillary} is much larger: it has 25k images of different, large resolutions ($8.9\times10^6$ pixels on average), and has labels for 65 classes (37 \textit{thing}, 28 \textit{stuff}). We train on the \textit{training} set (18k images) and evaluate on the \textit{validation} set (2k images).

\subsection{Implementation details}
\label{sec:experiments:impl_details}
All networks are trained on Nvidia A100 GPUs with 40GB memory each. For Cityscapes, we use 4 GPUs for training; for Mapillary Vistas we use 8. All networks use a ResNet-50 backbone~\cite{he2016resnet}, for a fair comparison. For both Panoptic FCN~\cite{li2021panopticfcn} and Mask2Former~\cite{cheng2021mask2former}, we extend their official published code and training settings, and crop-based training is conducted exactly like in the original works. Below, we specify the most relevant hyperparameters and settings. For more details, we refer to the original work~\cite{cheng2021mask2former,li2021panopticfcn} and \Cref{sec:app:impl_details}.

\textbf{Panoptic FCN.} For crop-based training on Cityscapes, we train for 65k steps with batches of 32 crops of 512$\times$1024 pixels, after randomly resizing the image with a factor between 0.5 and 2.0. For full-image training, we train the network for 100k steps on batches of 4 images, after randomly resizing them with a factor between 0.5 and 2.0. For Mapillary Vistas, we train for 150k steps with 32 crops of 1024$\times$1024 pixels, taken from images which are first randomly resized such that the shortest side is between 1024 and 2048 pixels. For full-image training, we train with batches of 8 images, for 150k steps, after randomly resizing the shortest side between 1024 and 2048 pixels.

\textbf{Mask2Former.} For Mask2Former, we make an important improvement to the code which is relevant for crop-based training, \ie, removing ground-truth segments outside the cropped region. Due to bipartite matching, these regions required a prediction, which the network could not make. By removing these segments during training, the baseline Panoptic Quality for Mapillary Vistas is significantly improved (36.3~\cite{cheng2021mask2former} vs.~41.5 in~\Cref{tab:main_results}). For Cityscapes, we train for 90k steps on batches of 16 crops of 512$\times$1024 pixels. For Mapillary Vistas, we train for 300k steps on batches of 16 crops of 1024$\times$1024 pixels. 

\section{Results}
\label{sec:results}

\begin{table}[t]
\centering
\adjustbox{width=\linewidth}{
\setlength{\tabcolsep}{5pt} %
    \begin{tabular}{lccccc}
    \toprule
    Network & PQ$\uparrow$ & PQ\textsuperscript{th}$\uparrow$ & PQ\textsuperscript{st}$\uparrow$ & Acc\textsuperscript{th}$\uparrow$ & Prec\textsuperscript{th}$\uparrow$ \\
    \midrule
    \multicolumn{6}{c}{Cityscapes \textit{val}} \\
    \midrule
    Panoptic FCN~\cite{li2021panopticfcn} & 59.5 & 52.2 & 64.8 & 81.3 & 86.8 \\
    Panoptic FCN~\cite{li2021panopticfcn} + IBS (ours) & \textbf{60.8} & \textbf{54.7} & \textbf{65.3} & \textbf{87.1} & \textbf{92.6}\\
    \midrule
    Mask2Former~\cite{cheng2021mask2former} & 62.1 & 55.2 & 67.2 & 87.1 & 93.3 \\
    Mask2Former~\cite{cheng2021mask2former} + IBS (ours) & \textbf{62.4} & \textbf{55.7} & \textbf{67.3} & \textbf{87.6} & \textbf{94.1} \\
    \midrule
    \multicolumn{6}{c}{Mapillary Vistas \textit{validation}} \\
    \midrule
    Panoptic FCN~\cite{li2021panopticfcn}  & 35.5 & 31.8 & \textbf{40.3} & 74.7 & 77.9 \\
    Panoptic FCN~\cite{li2021panopticfcn} + IBS (ours) & \textbf{36.3} & \textbf{33.6} & 40.0 & \textbf{77.0} & \textbf{82.2} \\
    \midrule
    Mask2Former~\cite{cheng2021mask2former} & 41.5 & 33.3 & \textbf{52.4} & 71.7 & 78.8 \\
    Mask2Former~\cite{cheng2021mask2former}+ IBS (ours)  & \textbf{42.2} & \textbf{34.9} & 52.0 & \textbf{75.7} & \textbf{84.1} \\
    \bottomrule
    \end{tabular}
}
\caption{\textbf{Main results.} We apply our IBS to Panoptic FCN~\cite{li2021panopticfcn} and Mask2Former~\cite{cheng2021mask2former}, and train and evaluate on the Cityscapes and Mapillary Vistas datasets~\cite{cordts2016cityscapes,neuhold2017mapillary}.
}
\label{tab:main_results}
\vspace{-5pt}
\end{table}

\subsection{Intra-Batch Supervision}
\begin{figure*}[t]
\centering
\adjustbox{width=\linewidth}{
\includegraphics[height=0.24\linewidth]{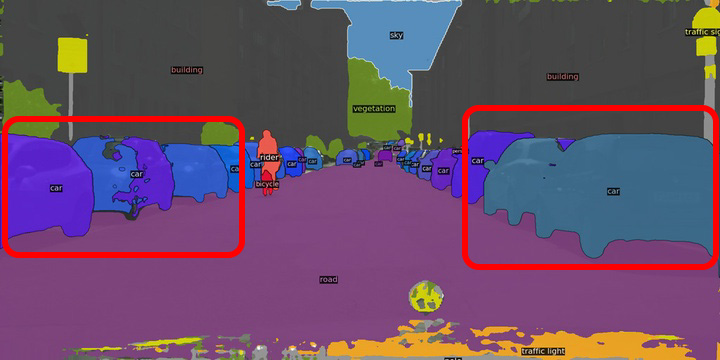}
\includegraphics[height=0.24\linewidth]{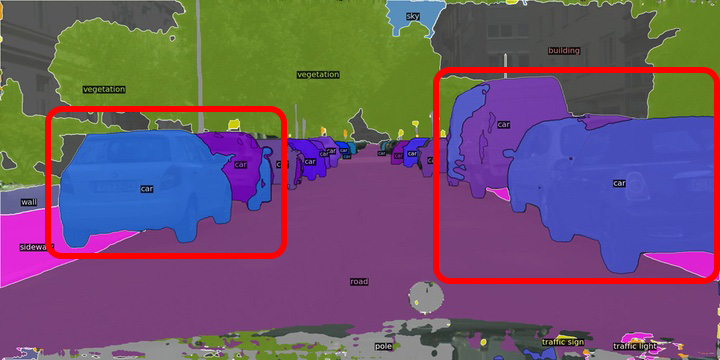}
\includegraphics[height=0.24\linewidth, trim={0.9cm 0 4.5cm 4.5cm},clip]{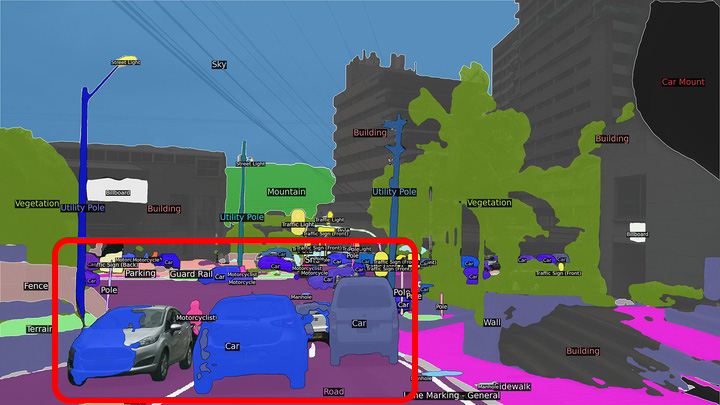}
\includegraphics[height=0.24\linewidth, trim={1.4cm 0.9cm 0.9cm 4.4cm},clip]{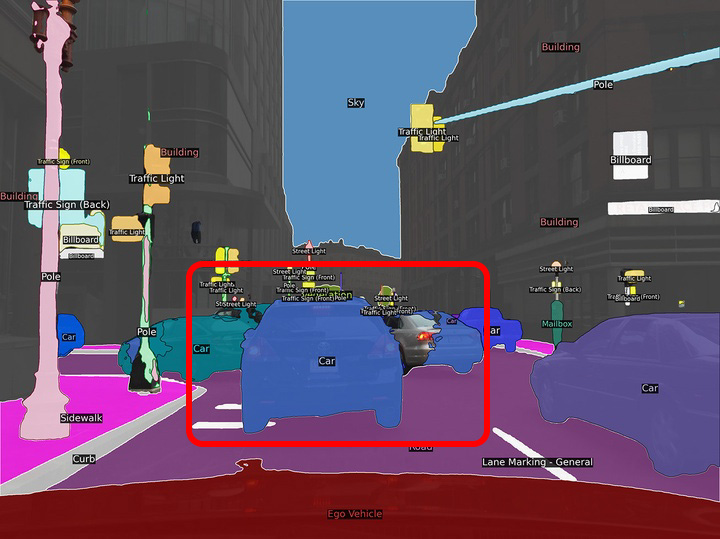}
}
\\
\adjustbox{width=\linewidth}{
\includegraphics[height=0.24\linewidth]{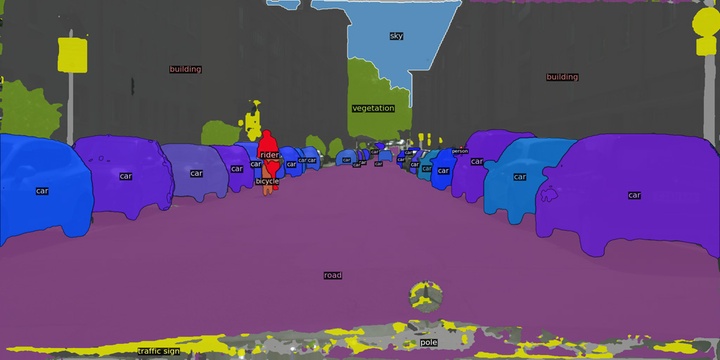}
\includegraphics[height=0.24\linewidth]{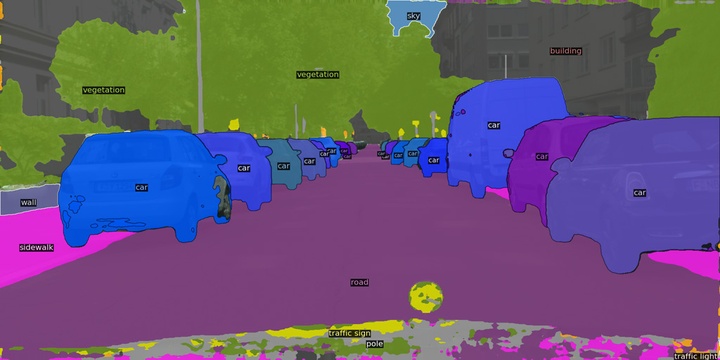}
\includegraphics[height=0.24\linewidth, trim={0.9cm 0 4.5cm 4.5cm},clip]{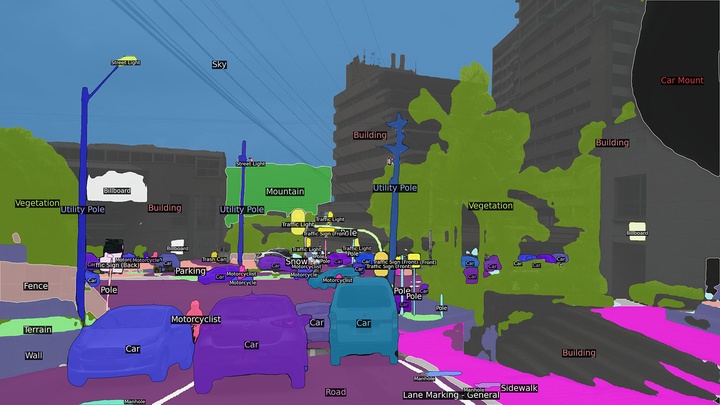}
\includegraphics[height=0.24\linewidth, trim={1.4cm 0.9cm 0.9cm 4.4cm},clip]{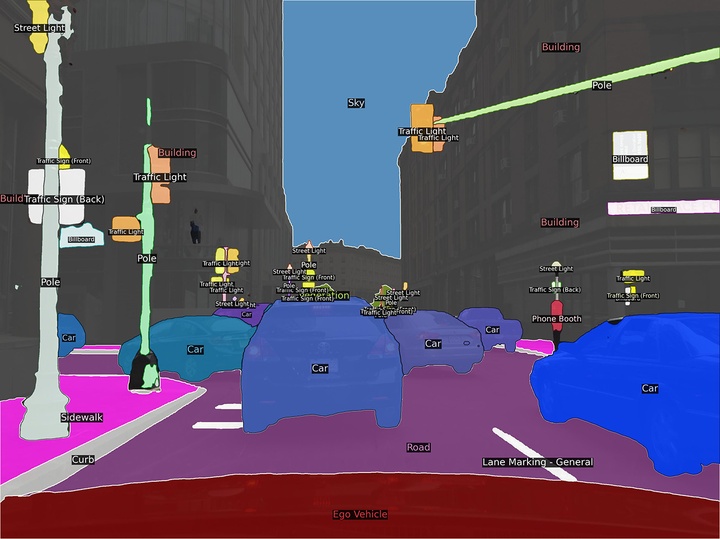}
}
\\

\caption{\textbf{Results for crop-based training on Panoptic FCN.} Top: without IBS, bottom: with IBS. From left to right, first two images: Cityscapes \textit{val} set~\cite{cordts2016cityscapes}; final two: Mapillary Vistas \textit{validation} set~\cite{neuhold2017mapillary}. Each segment gets a unique color and text label. The most prominent examples of predictions with \textit{confusion} are indicated with red rectangles. Best viewed digitally.
} 
\label{fig:results_main}
\vspace{0pt}
\end{figure*}

We apply our Intra-Batch Supervision (IBS) to two state-of-the-art unified panoptic segmentation methods, and evaluate the performance on two high-resolution datasets in \Cref{tab:main_results}. From these results, we find that IBS yields a consistent but modest improvement on the Panoptic Quality (PQ) metric~\cite{kirillov2019ps}. This is mainly caused by a boost in the PQ for things classes, PQ\textsuperscript{th}, which consistently shows considerable improvements. This is expected, because the aim of our IBS is to improve the confusion problem that occurs for thing segments. As explained in~\Cref{sec:problem_description:quantification}, however, the ability of the PQ metric to capture errors for large objects of frequently-occuring classes, such as the \textit{confusion} problem that we are addressing, is limited. To better capture this, we also report on the pixel accuracy (Acc\textsuperscript{th}) and pixel precision (Prec\textsuperscript{th}) for things classes. For these metrics, we observe much more significant improvements, \eg, a +5.8 improvement for both metrics for Panoptic FCN on Cityscapes, and +4.0 and +5.3 for Mask2Former on Mapillary Vistas. In other words, IBS  causes the network to correctly predict the panoptic segment for significantly more pixels than methods that used the naive crop-based training. This is also clearly visible in the qualitative examples shown in~\Cref{fig:results_main} and \Cref{sec:app:qual_results}. Where naive crop-based training causes inaccurate thing segmentation masks, IBS solves this problem.

\begin{figure}[t]
\centering
     \begin{subfigure}[b]{0.49\linewidth}
         \centering
         \includegraphics[width=\textwidth]{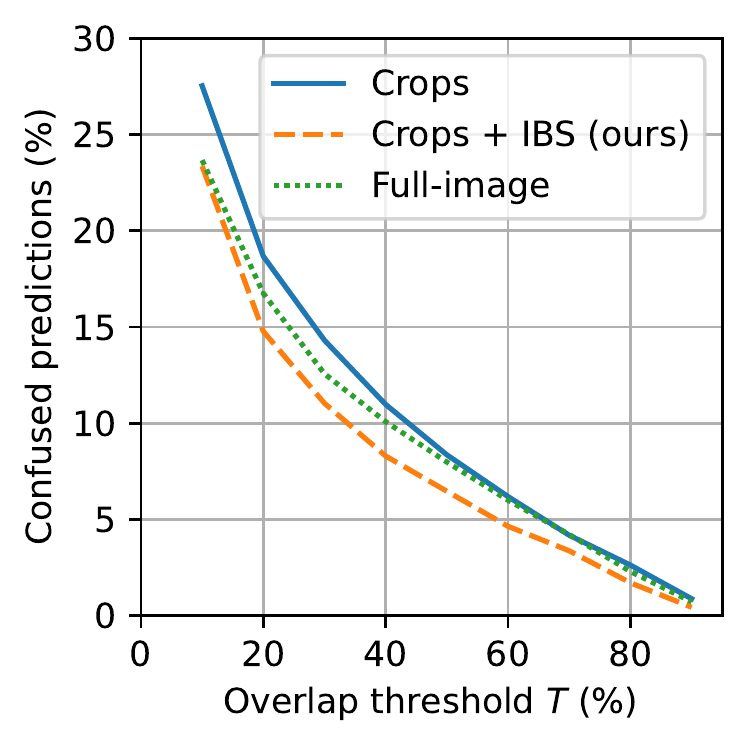}
         \vspace{-18pt}
         \caption{All thing segments}
         \label{fig:rcp_curves:cityscapes}
     \end{subfigure}
          \begin{subfigure}[b]{0.49\linewidth}
         \centering
         \includegraphics[width=\textwidth]{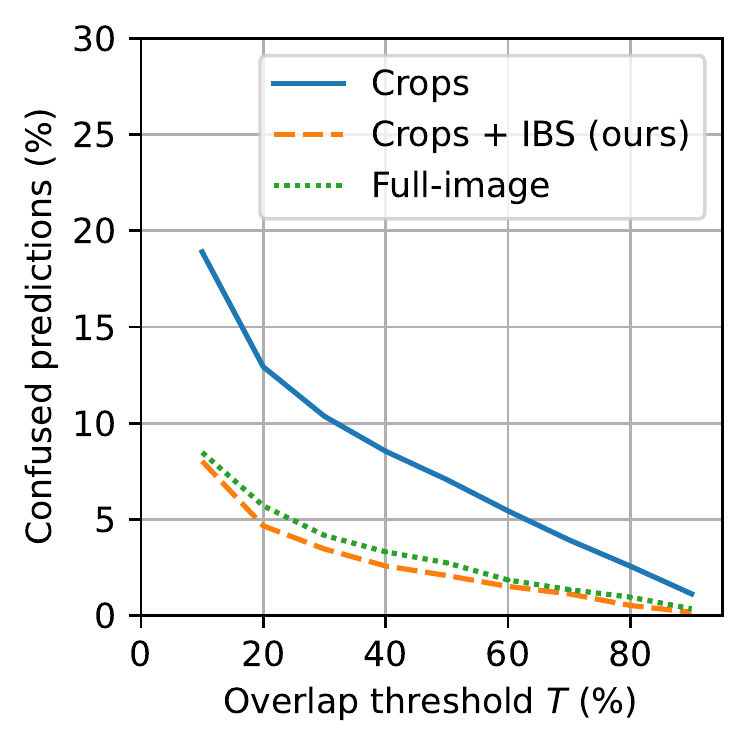}
         \vspace{-18pt}
         \caption{Thing segments $>4096$ px}
         \label{fig:rcp_curves:mapillary}
     \end{subfigure}
\vspace{-5pt}
\caption{\textbf{Confusion.} Ratio of predicted thing segments that intersect with at least \textit{two} ground-truth thing segments for more than $T\%$ of the ground-truth area, on Cityscapes \textit{val}~\cite{cordts2016cityscapes}, for different training settings. Lower is better.}
\label{fig:rctp_curves}
\vspace{0pt}
\end{figure}

\begin{table}[t]
\centering
\adjustbox{width=0.85\linewidth}{
\setlength{\tabcolsep}{4pt} %
    \begin{tabular}{ccc|ccccc}
    \toprule
    Crop & IBS & Sampling & PQ$\uparrow$ & PQ\textsuperscript{th}$\uparrow$ & PQ\textsuperscript{st}$\uparrow$& Acc\textsuperscript{th}$\uparrow$& Prec\textsuperscript{th}$\uparrow$\\

    \midrule
    \xmark & \xmark & random & 32.7 & 29.8 & 36.6 & 73.8 & 80.8 \\
    \cmark & \xmark & random & 35.5 & 31.8 & \textbf{40.3} & 74.7 & 77.9 \\
    \cmark & \cmark & random & 35.9 & 32.9 & 39.9 & 76.3 & \textbf{82.5} \\
    \cmark & \cmark & ours & \textbf{36.3} & \textbf{33.6} & 40.0 & \textbf{77.0} & 82.2 \\
    \bottomrule
    \end{tabular}
}
\caption{\textbf{Ablations.} Impact of different IBS components on Panoptic FCN~\cite{li2021panopticfcn} on Mapillary Vistas \textit{validation}~\cite{neuhold2017mapillary}.}
\label{tab:ablations}
\vspace{-5pt}
\end{table}

\subsection{Confusion}
To assess the effect of IBS on the specific confusion problem that we identified, instead of the more general metrics, we conduct an additional experiment where we calculate how often confusion occurs for predicted thing segments. Concretely, for different thresholds $T$, we calculate how many predicted thing segments intersect with \textit{at least 2 ground-truth thing segments} for more than $T\%$ of the ground-truth area. The resulting number will indicate \textit{how many predicted thing segments are confused between at least two segments} given an overlap threshold $T$. In~\Cref{fig:rctp_curves}, we show the results for this measure for full-image training, naive crop-based training and crop-based training with IBS. We find that, as we hypothesized, naive crop-based training leads to considerably more confusion than full-image training, especially for large thing segments. When we introduce our IBS, however, this confusion drops significantly. These results confirm that the confusion problem caused by crop-based training mainly impacts large segments, and that our IBS effectively solves this problem.

\subsection{Ablations}
In \Cref{tab:ablations}, we show the effect of the different components of our proposed IBS. From these results, we find that vanilla IBS with random crop sampling already leads to a considerable improvement on the pixel accuracy and pixel precision for things, and a moderate improvement on the PQ. When we introduce our new sampling strategy, which makes sure that IBS is applied to images containing segments from the same semantic class, both the pixel accuracy and the PQ increase even further. This shows that IBS is effective, and that an effective crop sampling strategy, which aims to boost the amount of \textit{meaningful} supervision, is required to unlock its full potential.

\subsection{Instance segmentation}
Because unified networks can also be used for instance segmentation, we expect the confusion problem to occur here as well. To show the more general applicability of our IBS to other computer visions tasks, we apply IBS to Mask2Former~\cite{cheng2021mask2former} for instance segmentation on Mapillary Vistas~\cite{neuhold2017mapillary}. In~\Cref{tab:results_inst_seg}, we report the results on the Average Prediction (AP), the standard COCO metric~\cite{lin2014coco}.
We find that our IBS leads to an improvement on the AP, especially for the large instances (AP\textsuperscript{L}) that are most impacted by confusion. Like for panoptic segmentation, we also find significant boosts for the Acc\textsuperscript{th} and Prec\textsuperscript{th}, with +4.3 for both metrics. This shows that IBS can also effectively solve the confusion problem for instance segmentation.

\section{Discussion}
\label{sec:discussion}
With experiments, we found that the confusion problem is not effectively captured by the Panoptic Quality (PQ) metric, because large thing segments have only limited impact.
Instead, we found that other established segmentation metrics, \ie, the pixel accuracy and pixel precision, are better suited to measure the confusion problem. Therefore, although the PQ is an essential metric -- as it is important that a method works well on segments of all categories and different sizes -- we advocate for the use of \textit{multiple} metrics to evaluate methods, and not just a single metric like the PQ, to properly assess their performance for a variety of downstream tasks.
We hypothesize that focusing only on a single metric is one of the reasons that the confusion problem has not been addressed until now, despite its relevance for downstream tasks such as autonomous driving and robotics because of its impact on the performance for large objects.

\begin{table}[t]
\centering
\adjustbox{width=1.0\linewidth}{
\setlength{\tabcolsep}{3pt} %
    \begin{tabular}{lcccccc}
    \toprule
    Network & AP$\uparrow$ & AP\textsuperscript{S}$\uparrow$ & AP\textsuperscript{M}$\uparrow$ & AP\textsuperscript{L}$\uparrow$ & Acc\textsuperscript{th}$\uparrow$ & Prec\textsuperscript{th}$\uparrow$ \\
    \midrule
    Mask2Former~\cite{cheng2021mask2former} & 17.4 & \textbf{5.3} & 18.2 & 34.9 & 73.4 & 74.9 \\
    Mask2Former~\cite{cheng2021mask2former} + IBS (ours) & \textbf{18.1} & 4.8 & \textbf{18.5} & \textbf{36.8} & \textbf{77.7} & \textbf{79.2} \\
    \bottomrule
    \end{tabular}
}
\caption{\textbf{Instance segmentation.} IBS results for instance segmentation with Mask2Former~\cite{cheng2021mask2former} evaluated on the Mapillary Vistas \textit{validation} set~\cite{neuhold2017mapillary}.}
\label{tab:results_inst_seg}
\vspace{0pt}
\end{table}

\section{Conclusions}
\label{sec:conclusion}
In this work, we have shown that the generally advantageous crop-based training strategy has a negative side-effect that causes
state-of-the-art, unified panoptic segmentation networks such as Panoptic FCN~\cite{li2021panopticfcn} and Mask2Former~\cite{cheng2021mask2former} to predict inaccurate segmentation masks for thing instances. To solve this, we have proposed Intra-Batch Supervision (IBS), which introduces additional supervision for thing segments by teaching the network to distinguish thing instances from instances in other crops. With experiments, we have shown that our IBS solves the confusion problem for multiple unified panoptic segmentation networks on two high-resolution datasets, causing consistent improvements on the Panoptic Quality for things, and large gains on the pixel accuracy and pixel prediction metrics. Moreover, we have demonstrated that IBS is effective on the instance segmentation task as well. These results demonstrate that our IBS is necessary to make accurate predictions on high-resolution datasets with state-of-the-art, unified networks.

\vspace{-10pt}
\paragraph{Acknowledgements}
This work is supported by Eindhoven Engine, NXP Semiconductors and Brainport Eindhoven. This work made use of the Dutch national e-infrastructure with the support of the SURF Cooperative using grant no.~EINF-2748, which is financed by the Dutch Research Council (NWO).

{\small
\bibliographystyle{ieee_fullname}
\bibliography{egbib.bib}
}

\clearpage
\appendix 

\section*{Appendix}

\noindent In the appendix, we provide the following supplementary material:
\begin{itemize}
    \item More extensive implementation details (\Cref{sec:app:impl_details}).
    \item Additional qualitative results, showing the \textit{confusion} problem and the effect of our IBS to solve it (\Cref{sec:app:qual_results}).
\end{itemize}

All code is available through \url{https://ddegeus.github.io/intra-batch-supervision/}.

\section{Implementation details}
\label{sec:app:impl_details}
We provide more extensive implementation details for the neural networks we use for the experiments in our main manuscript. For both Panoptic FCN~\cite{li2021panopticfcn} and Mask2Former~\cite{cheng2021mask2former}, we use and extend the official code repositories, which are both based on PyTorch~\cite{paszke2019pytorch} and Detectron2~\cite{wu2019detectron2}. All networks use a ResNet-50 backbone~\cite{he2016resnet}, which is initialized with weights pre-trained on ImageNet~\cite{deng2009imagenet}. In general, we use the original implementation details of both Panoptic FCN and Mask2Former. Whenever we use different settings, we indicate this explicitly.

For both networks, we add the focal loss~\cite{lin2017focal} for IBS to the original losses of the network, and calculate the total loss as

\begin{equation}
    L_{total} = L_{orig} + \lambda_{IBS}L_{IBS},
\end{equation}

\noindent where $L_{orig}$ are the original losses, $L_{IBS}$ is the focal loss for IBS, $\lambda_{IBS}$ is the loss weight for the IBS loss, and $L_{total}$ is the resulting total loss.

\subsection{Panoptic FCN}
Panoptic FCN is optimized with stochastic gradient descent, and uses a polynomial learning rate schedule with initial learning rate $lr_0$ and decay 0.9. The weight decay is $10^{-4}$. We use $\lambda_{IBS} = 1$, and we empirically find that the network is robust to relatively small variations to this value ([0.5; 5.0]). The embedding dimension is set to $C = 256$, instead of 64 as in the original work. We find that this improves the performance both with and without IBS. In all experiments, we apply random horizontal flipping to the images and ground-truth before feeding them to the network.

\paragraph{Cityscapes.}
For crop-based training on Cityscapes~\cite{cordts2016cityscapes}, $lr_0 = 0.02$. Following the original settings, we train for 65k steps with batches of 32 crops of $512\times1024$ pixels, after they are randomly resized with a factor between 0.5 and 2.0. For full-image training, we use $lr_0 = 0.005$. We train for 100k steps with batches of 4 images, after randomly resizing them with a factor between 0.5 and 2.0.

\paragraph{Mapillary Vistas.}
For crop-based training on Mapillary Vistas~\cite{neuhold2017mapillary}, $lr_0 = 0.02$. We train for 150k steps with batches of 32 crops of $1024\times1024$ pixels, after the images are randomly resized such that the shortest side is between 1024 and 2048 pixels. For full-image training, $lr_0 = 0.005$. We train for 150k steps with batches of 8 images, after the images are randomly resized such that the shortest side is between 1024 and 2048 pixels.

\subsection{Mask2Former}
Mask2Former is optimized using AdamW~\cite{loshchilov2019adamw}, and uses an initial learning rate $lr_0 = 0.0001$. The weight decay is 0.05. IBS is applied to the predictions made at each transformer decoder layer, except the first one. To apply IBS, we identify and extract the per-segment \textit{embeddings} belonging to thing segments based on the ground-truth segment they are matched to with the bipartite matching algorithm that is also used for the other losses. We use $\lambda_{IBS} = 100$ to balance the losses in such a way that ratio between the magnitudes of $L_{IBS}$ and $L_{orig}$ is similar to the one for Panoptic FCN. For Mask2Former, we only do experiments with crop-based training. In all experiments, we apply random horizontal flipping to the images and ground-truth before feeding them to the network.

\paragraph{Cityscapes.}
Following the original settings, we train for 90k steps using batches of 16 crops of $512\times1024$ pixels, which are taken from the original images after first resizing them with a random factor between 0.5 and 2.0.

\paragraph{Mapillary Vistas.}
We train for 300k steps using batches of 16 crops of $1024\times1024$ pixels, which are taken from the original images after first randomly resizing them such that the shortest side is between 1024 and 2048 pixels. The training settings for instance segmentation are equal to those for panoptic segmentation, except that no stuff predictions are made for instance segmentation.

\section{Qualitative results}
\label{sec:app:qual_results}
We provide additional qualitative results to illustrate both the \textit{confusion} problem with crop-based training, and the effectiveness of IBS to solve this problem. 

\begin{figure*}[t]
\centering
\includegraphics[width=0.28\linewidth]{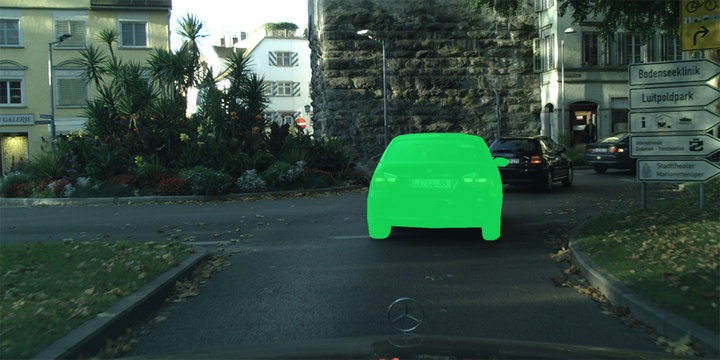}
\includegraphics[width=0.28\linewidth]{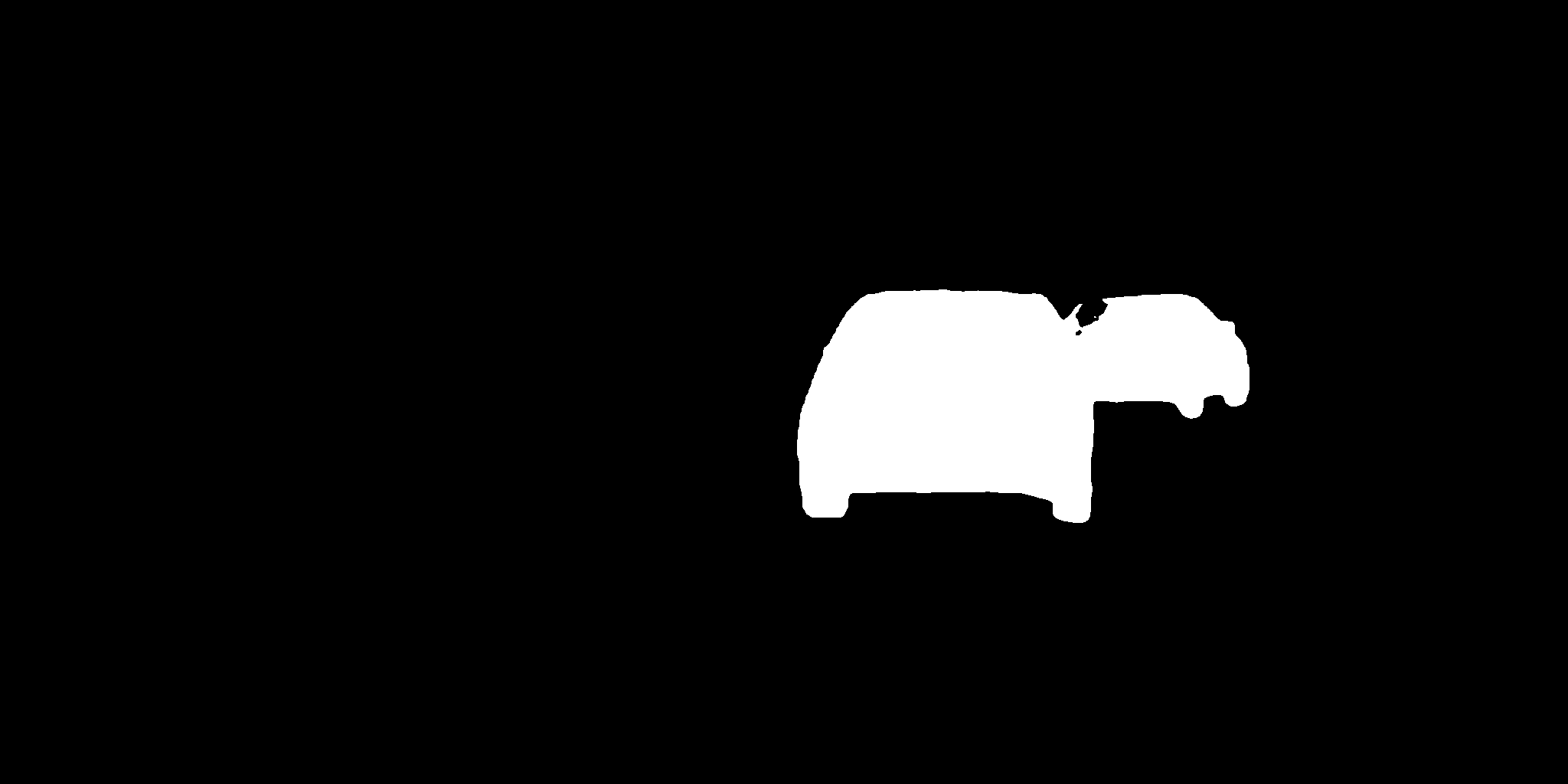}
\includegraphics[width=0.28\linewidth]{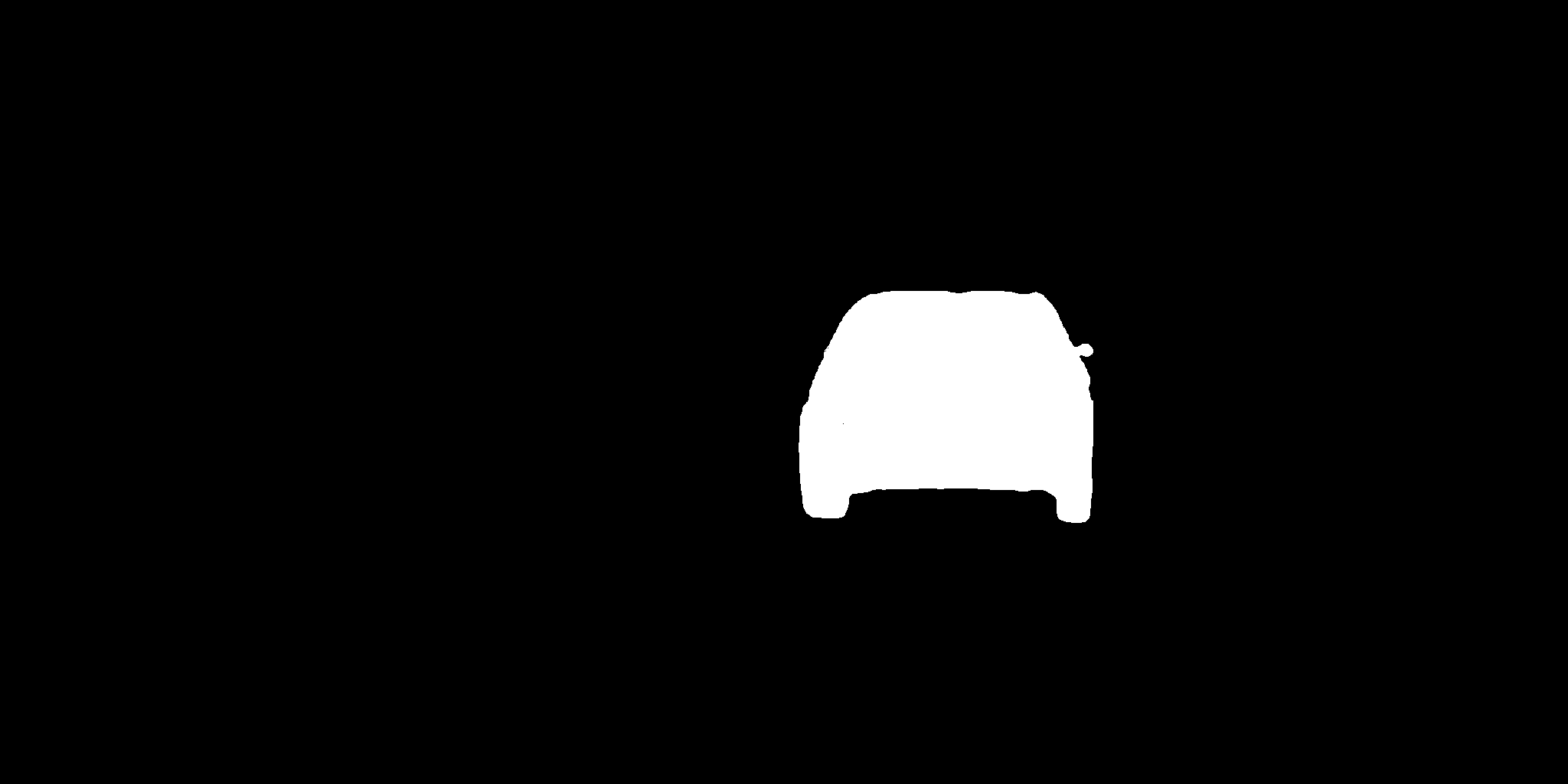}\\

\includegraphics[width=0.28\linewidth]{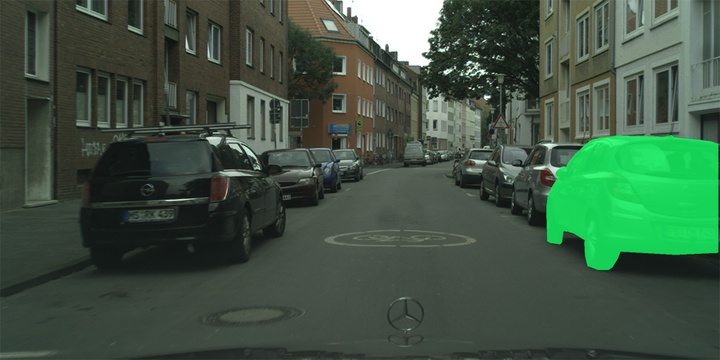}
\includegraphics[width=0.28\linewidth]{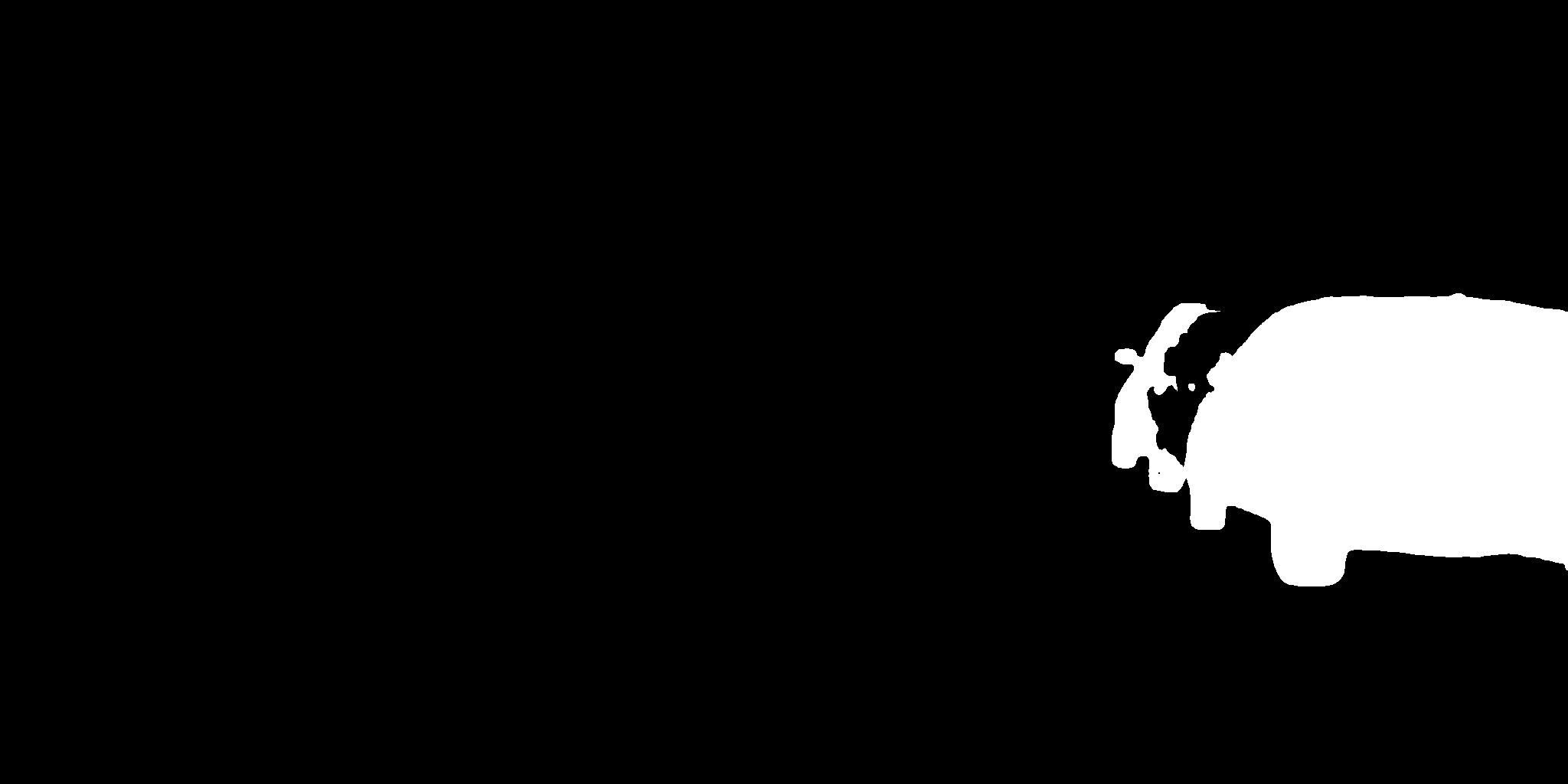}
\includegraphics[width=0.28\linewidth]{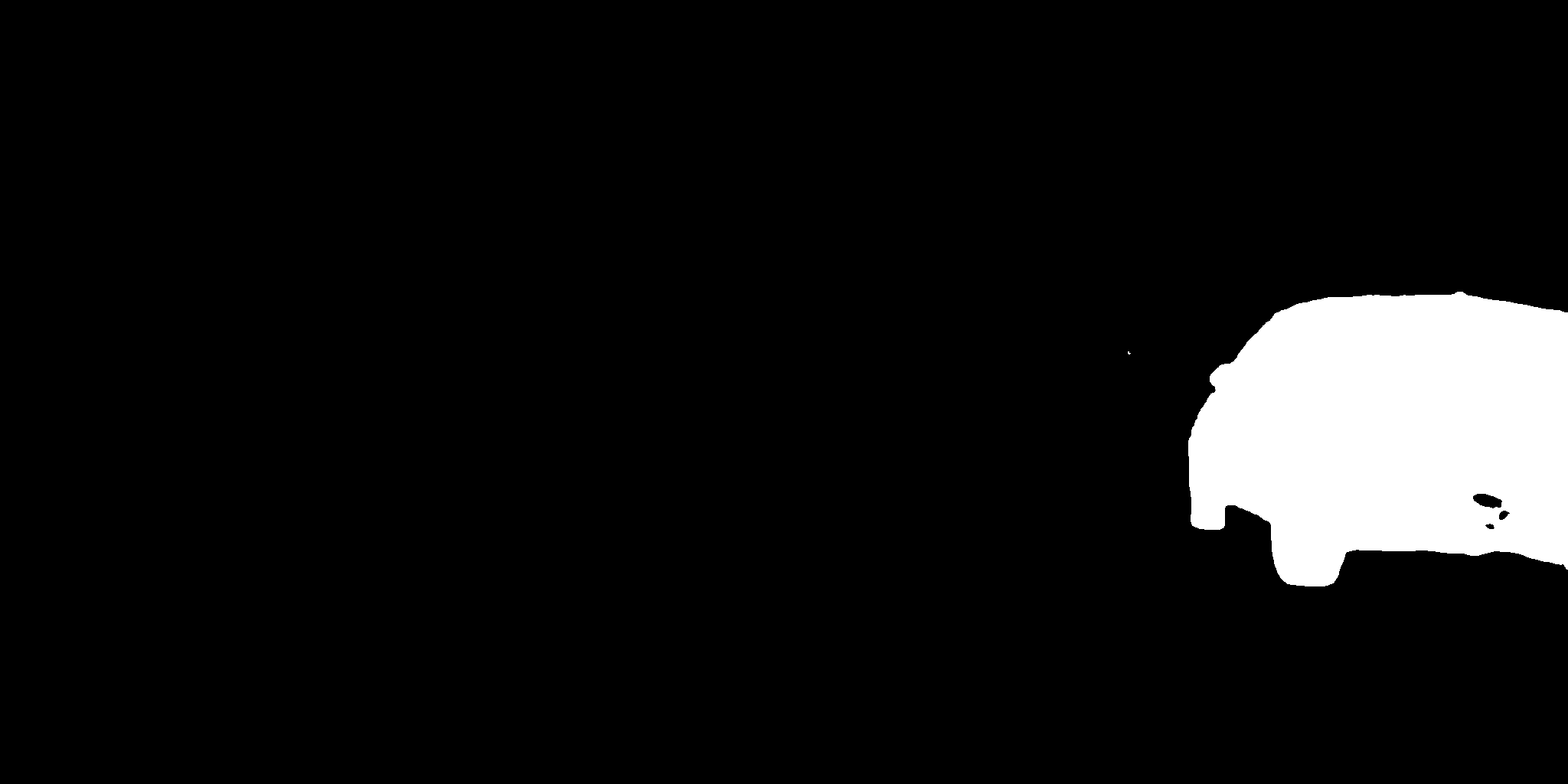}\\

\includegraphics[width=0.28\linewidth]{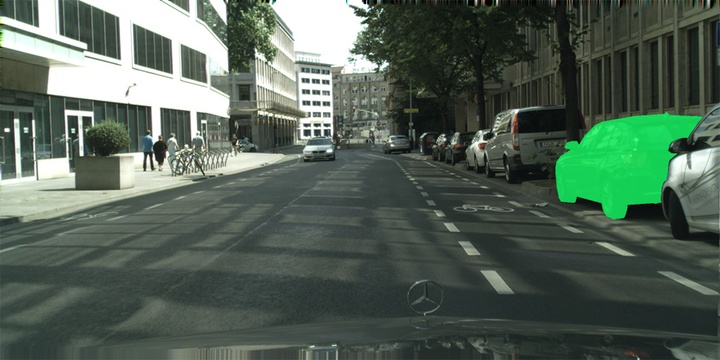}
\includegraphics[width=0.28\linewidth]{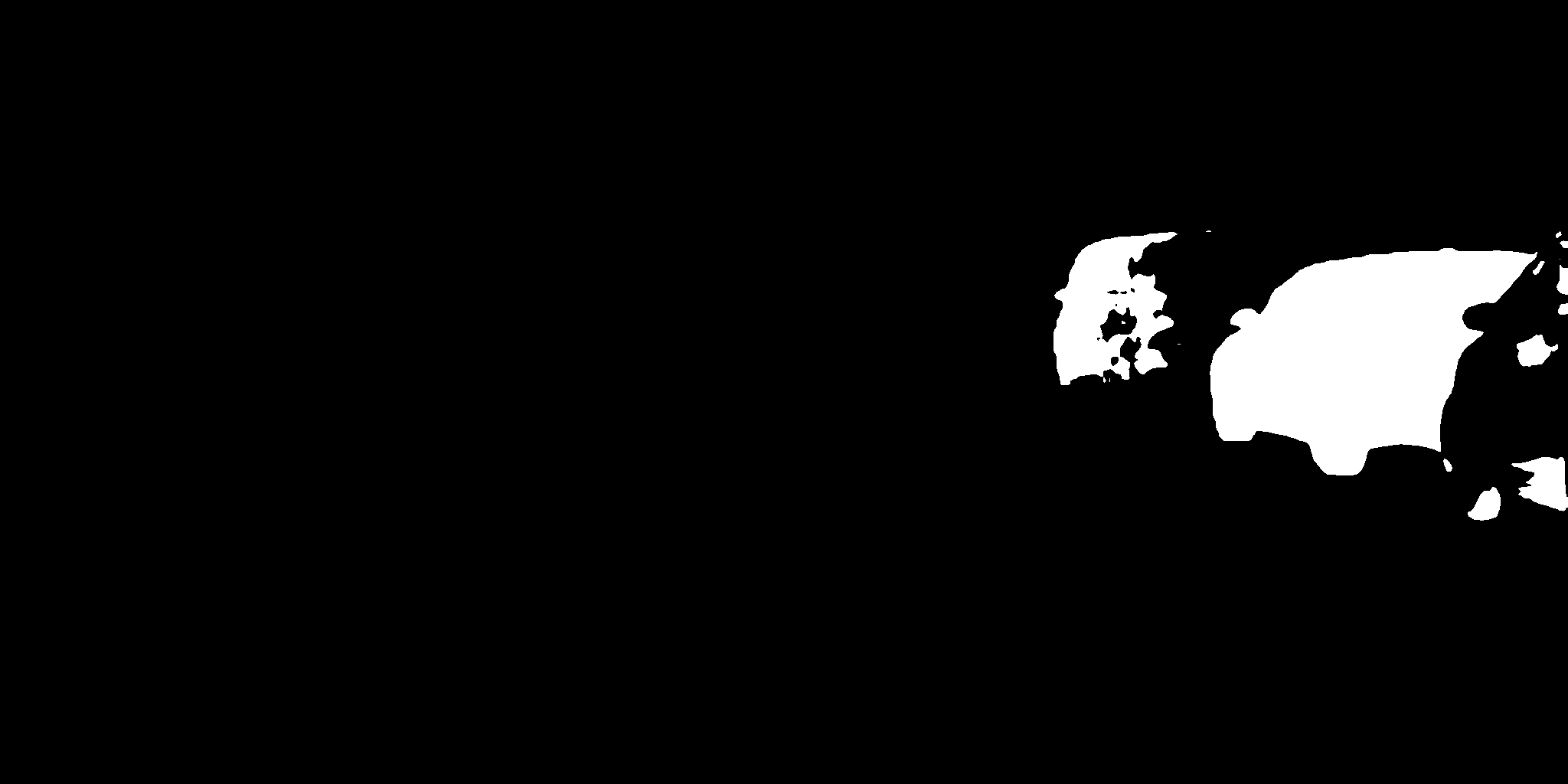}
\includegraphics[width=0.28\linewidth]{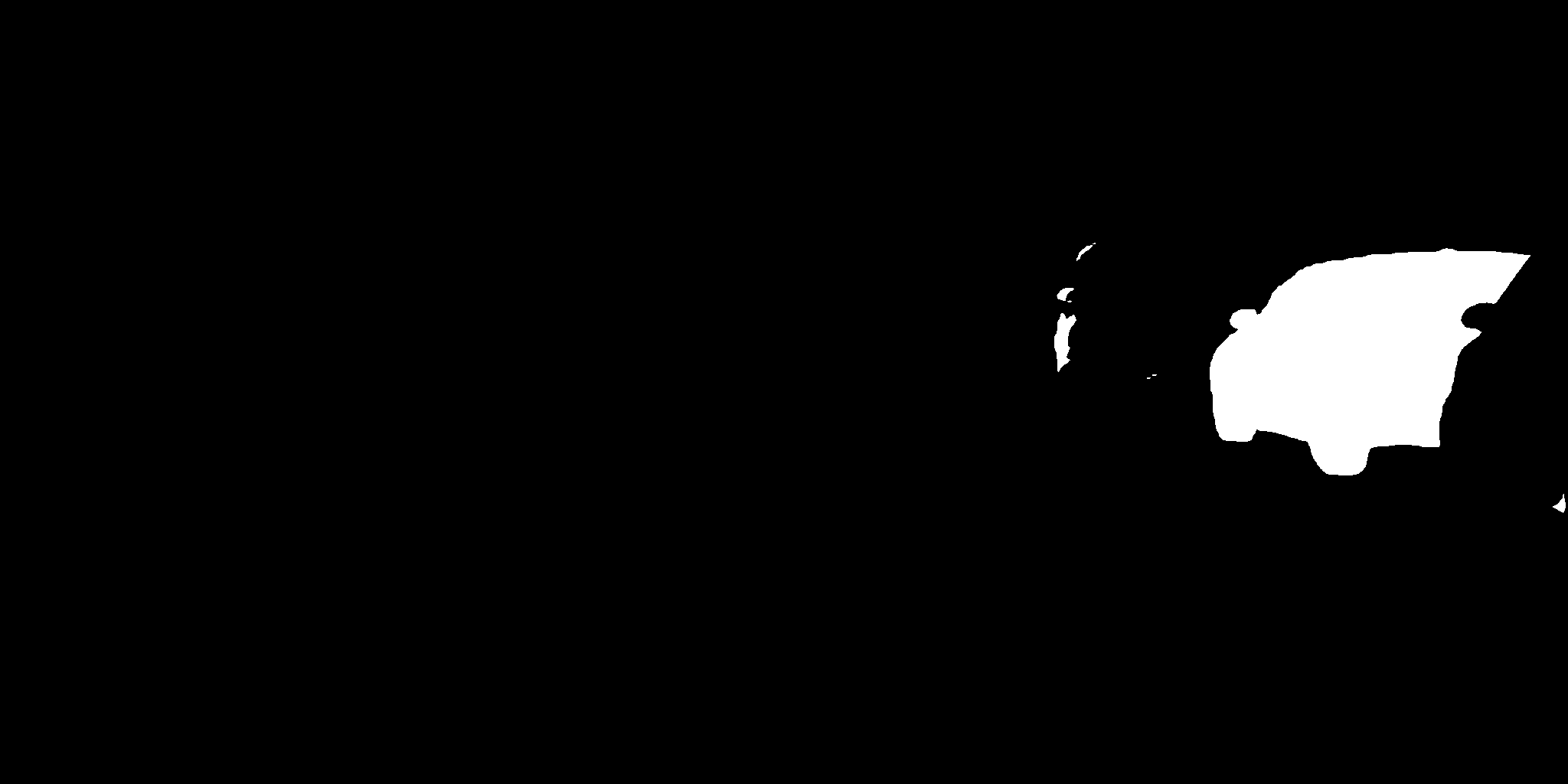}\\

\includegraphics[width=0.28\linewidth]{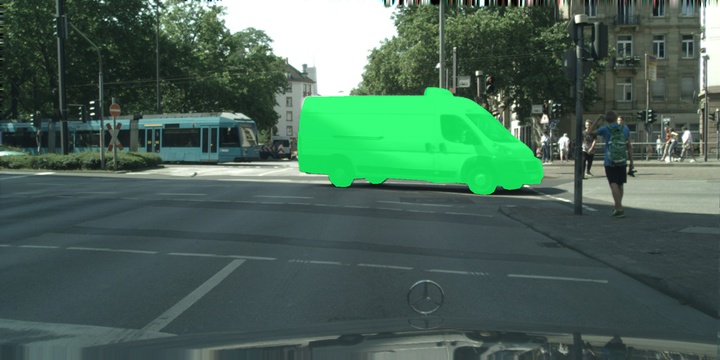}
\includegraphics[width=0.28\linewidth]{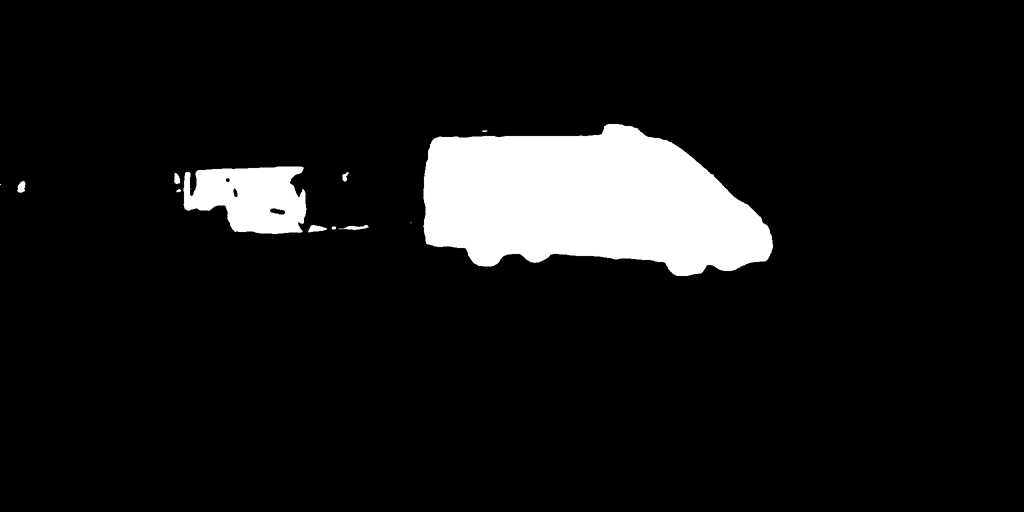}
\includegraphics[width=0.28\linewidth]{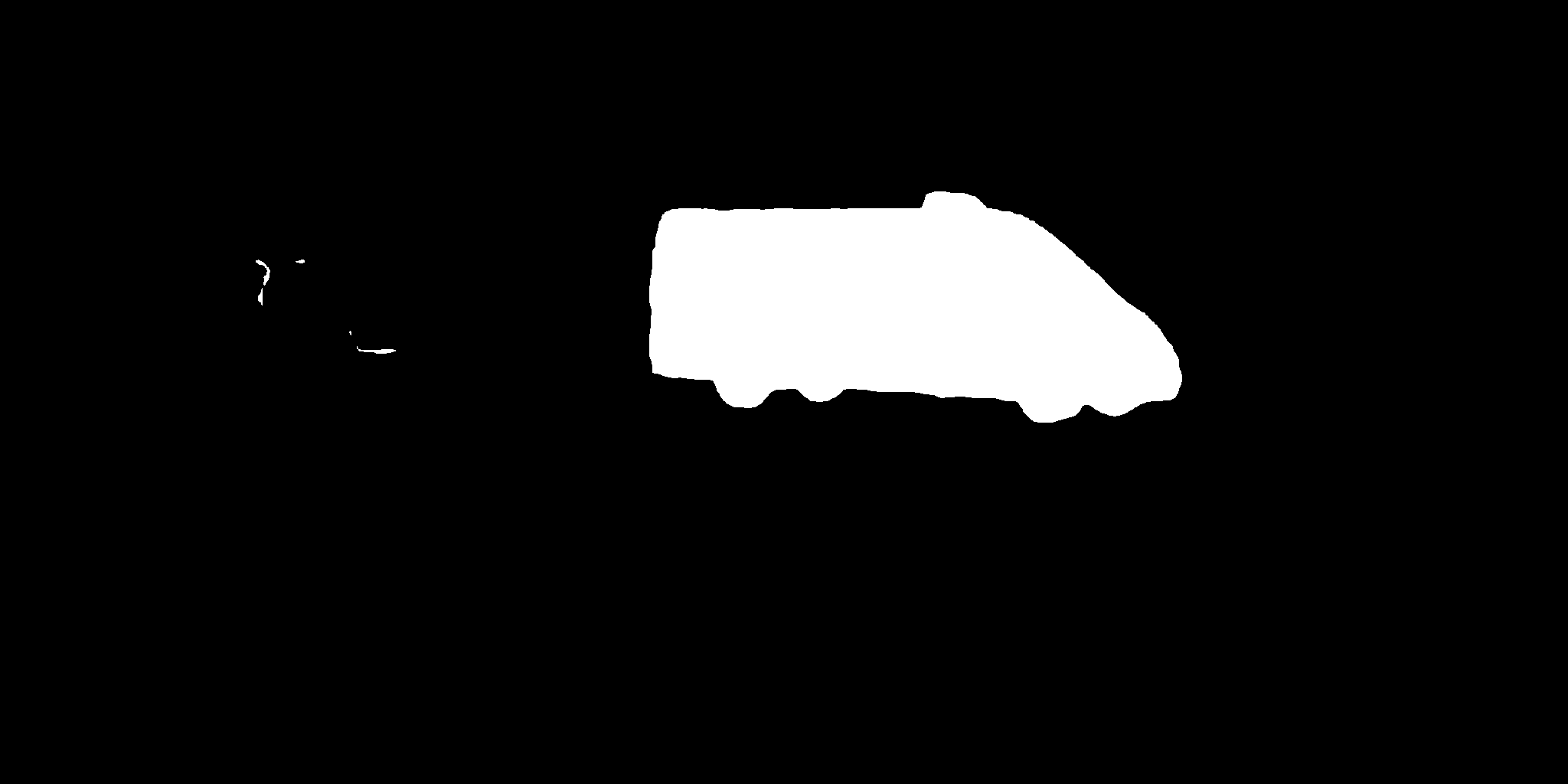}\\

\includegraphics[width=0.28\linewidth]{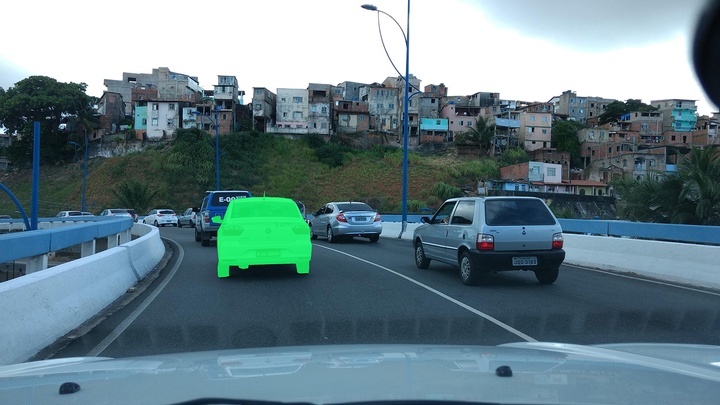}
\includegraphics[width=0.28\linewidth]{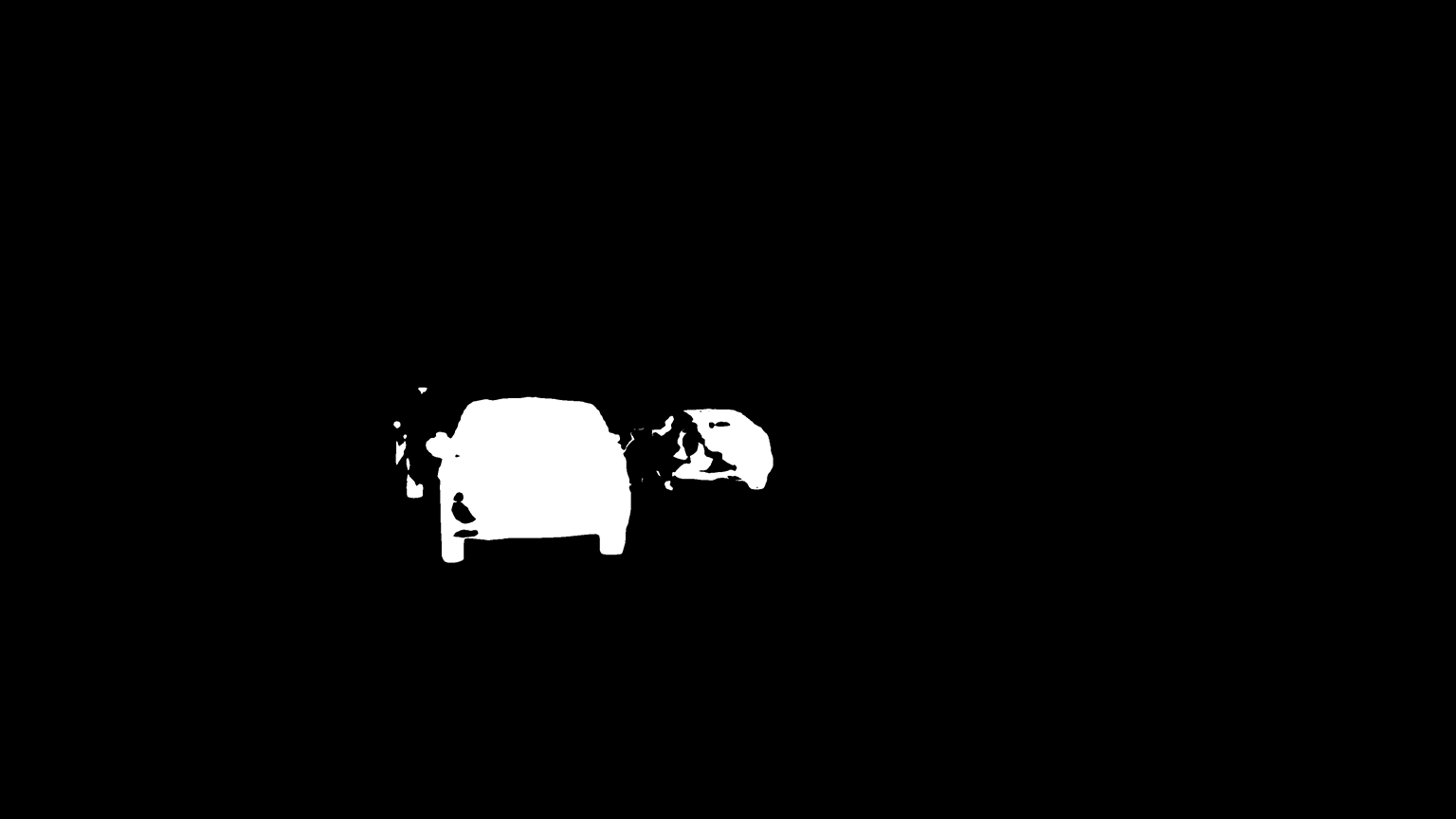}
\includegraphics[width=0.28\linewidth]{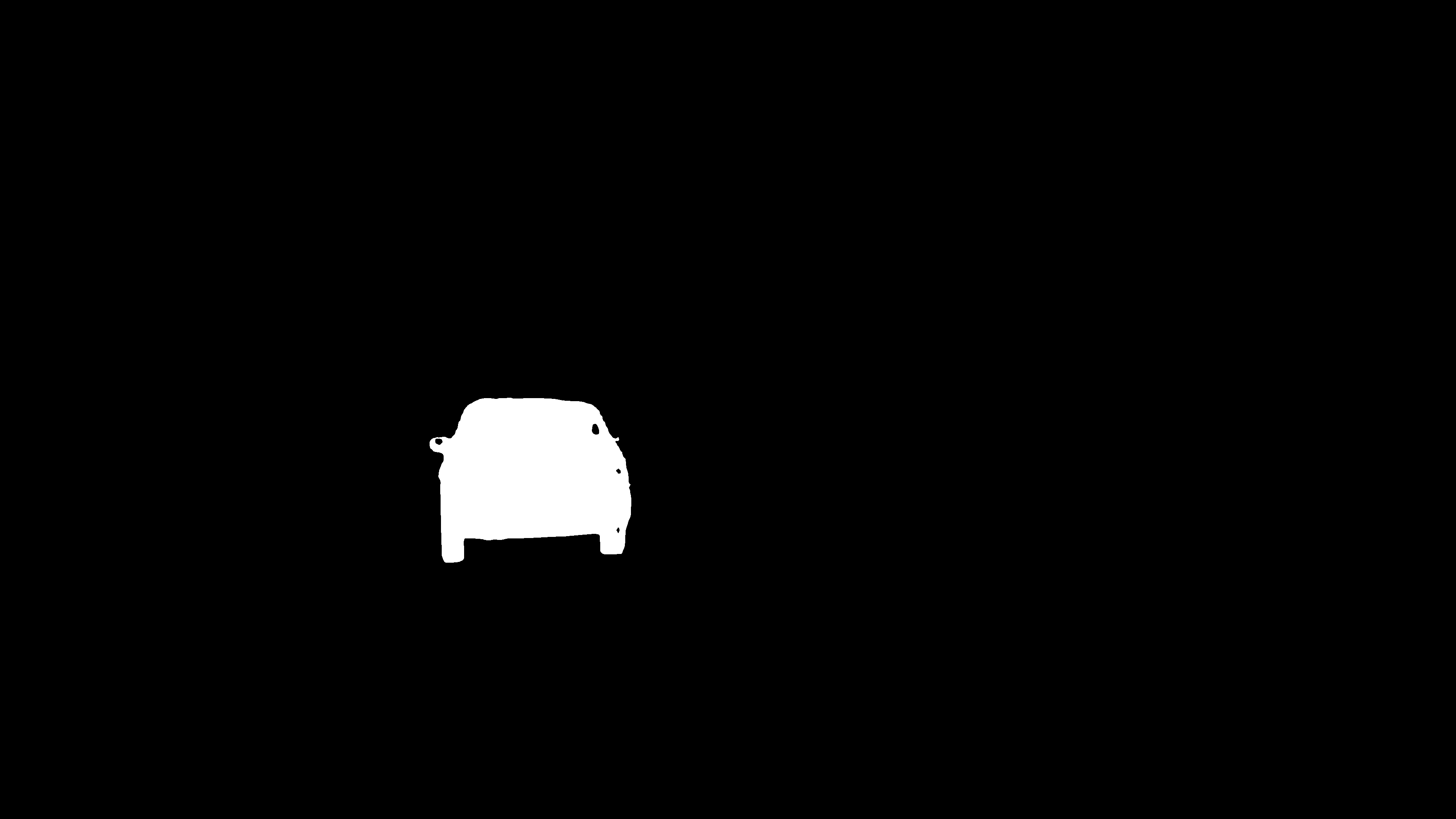}\\

\includegraphics[width=0.28\linewidth, trim={0 0 0 5.5130cm},clip]{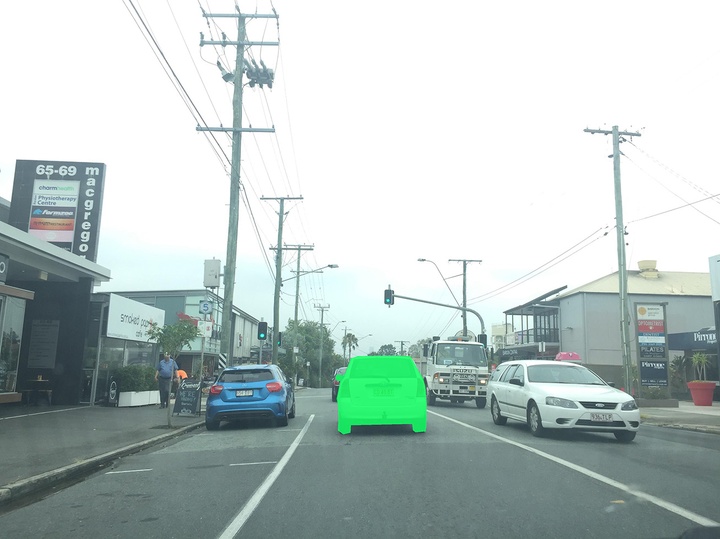}
\includegraphics[width=0.28\linewidth, trim={0 0 0 10cm},clip]{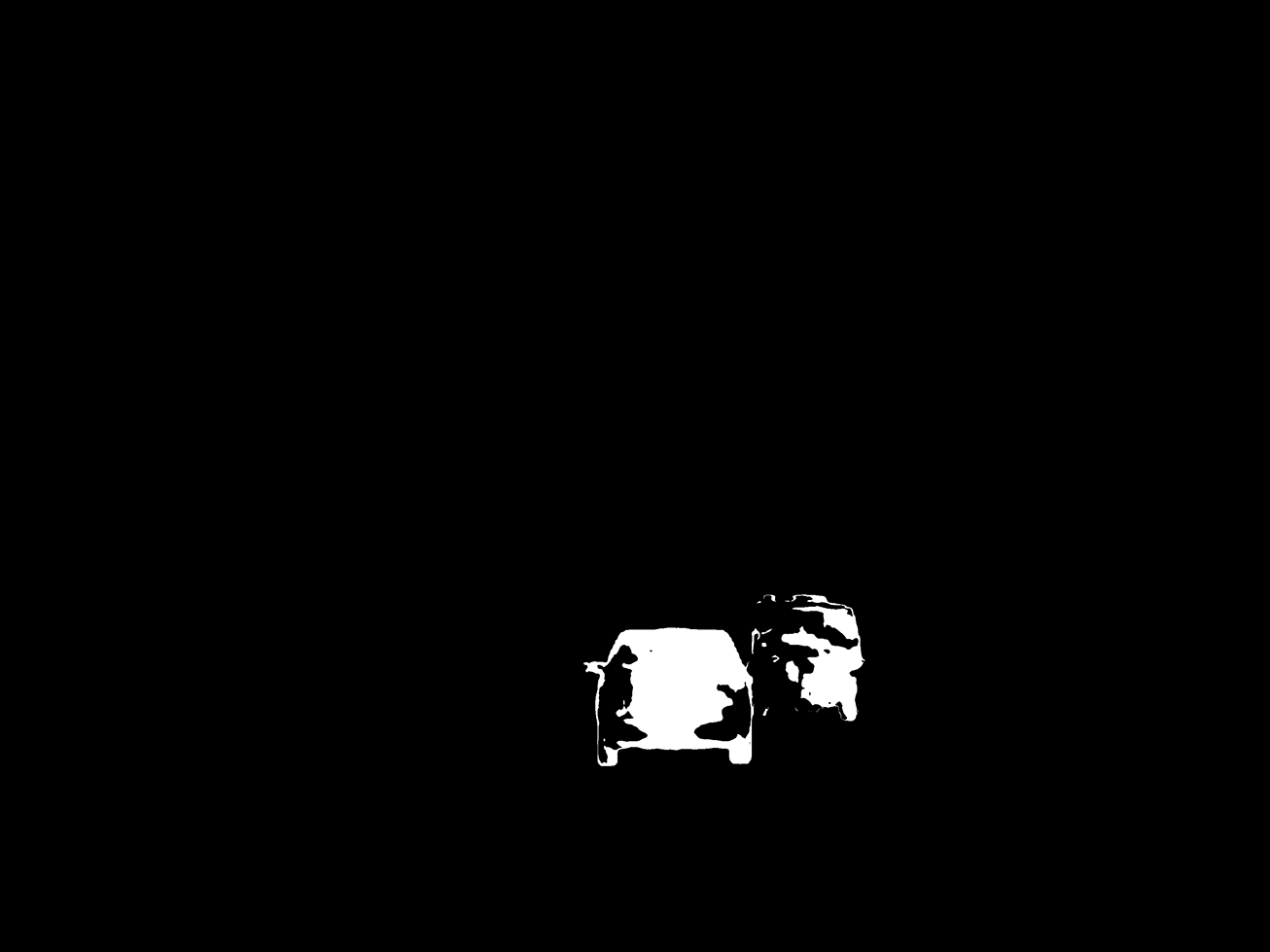}
\includegraphics[width=0.28\linewidth, trim={0 0 0 25cm},clip]{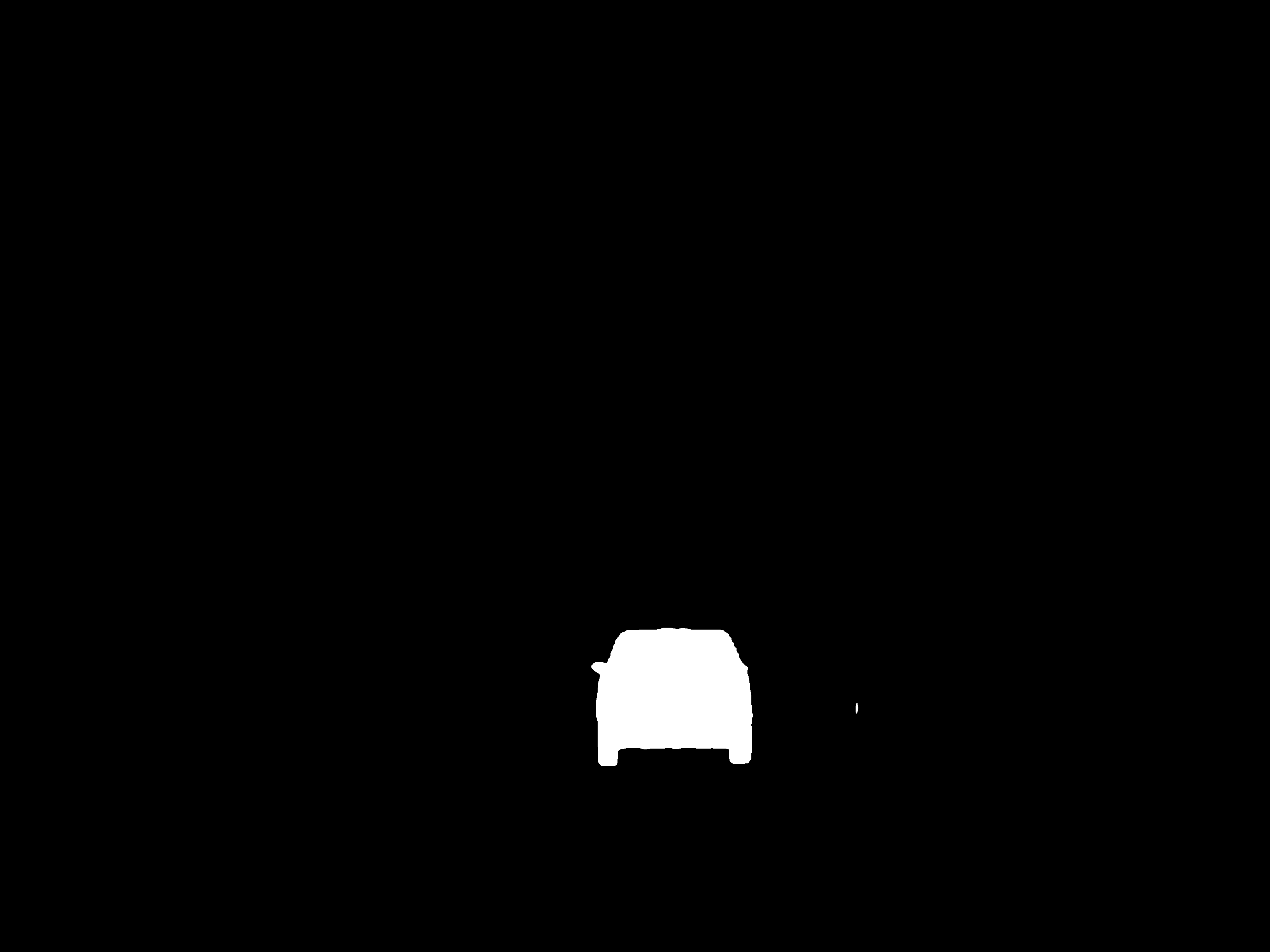}\\

\includegraphics[width=0.28\linewidth]{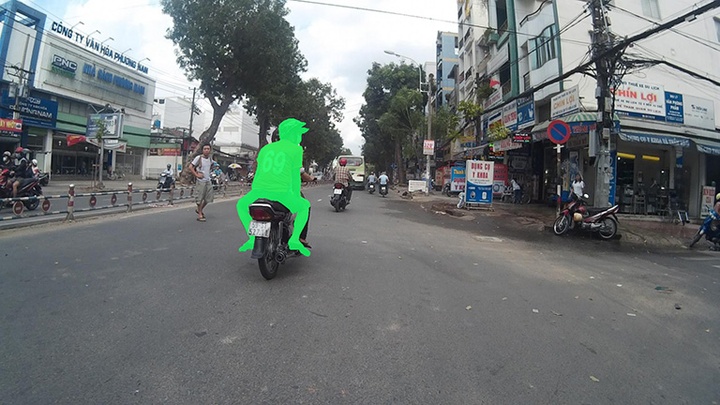}
\includegraphics[width=0.28\linewidth]{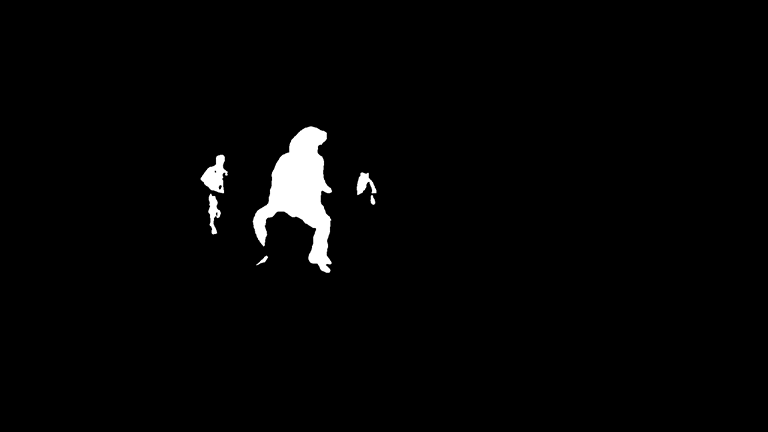}
\includegraphics[width=0.28\linewidth]{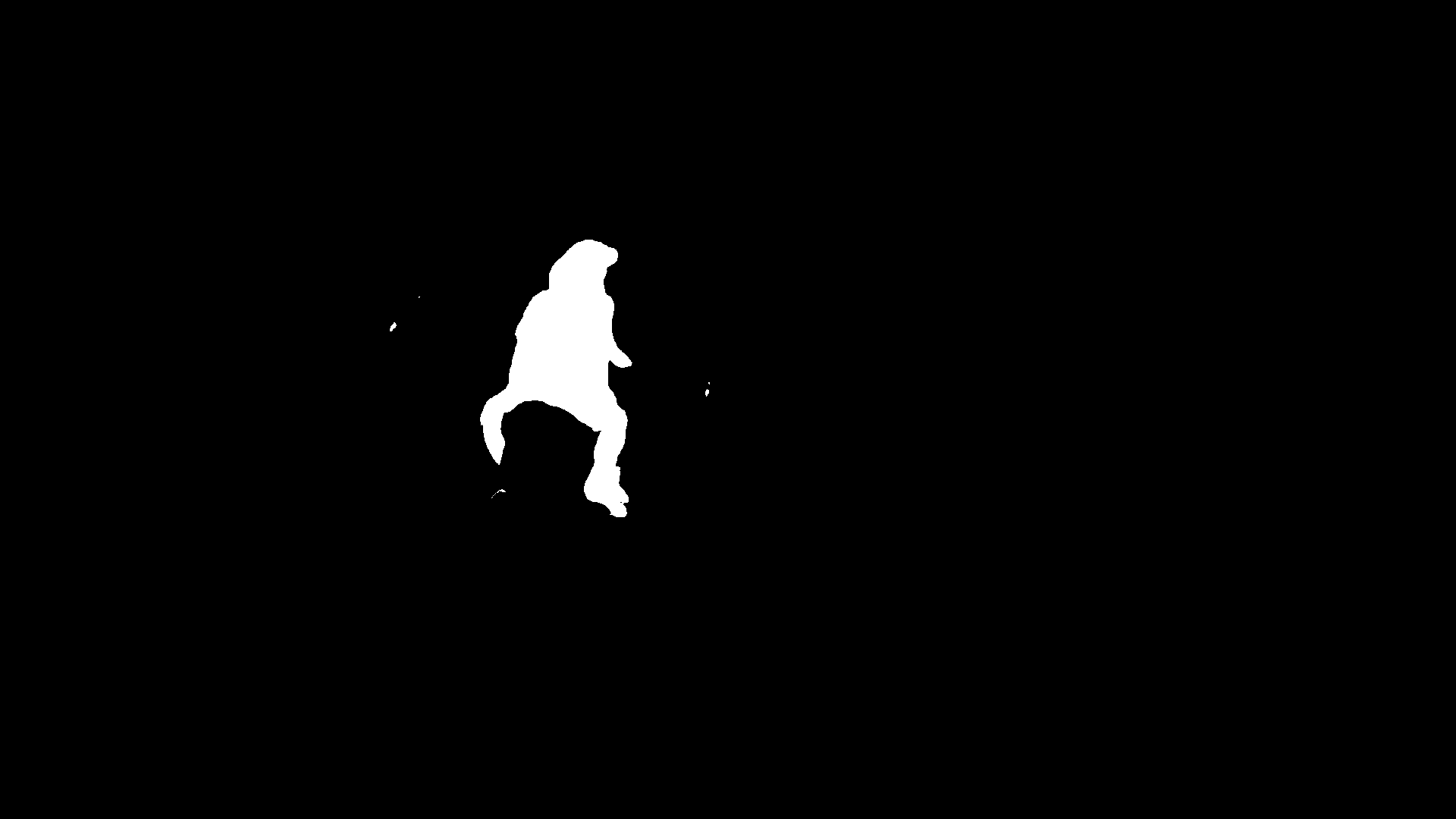}\\


\includegraphics[width=0.28\linewidth, trim={0 1.3542cm 0 4.514cm},clip]{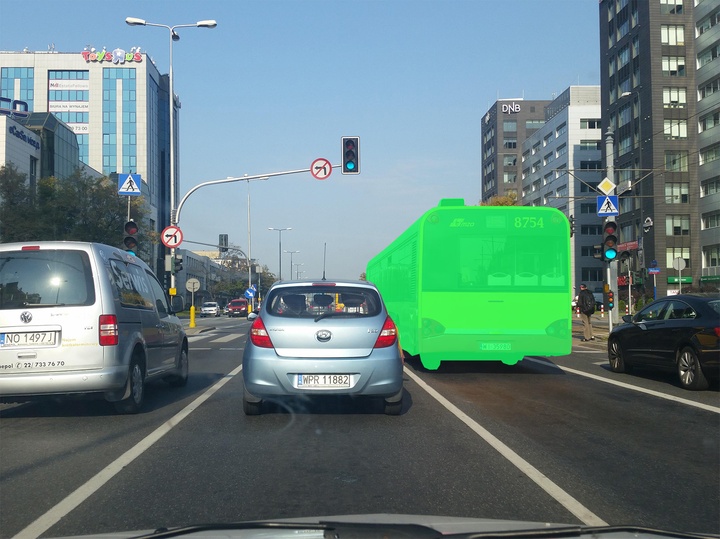}
\includegraphics[width=0.28\linewidth, trim={0 7.5cm 0 25cm},clip]{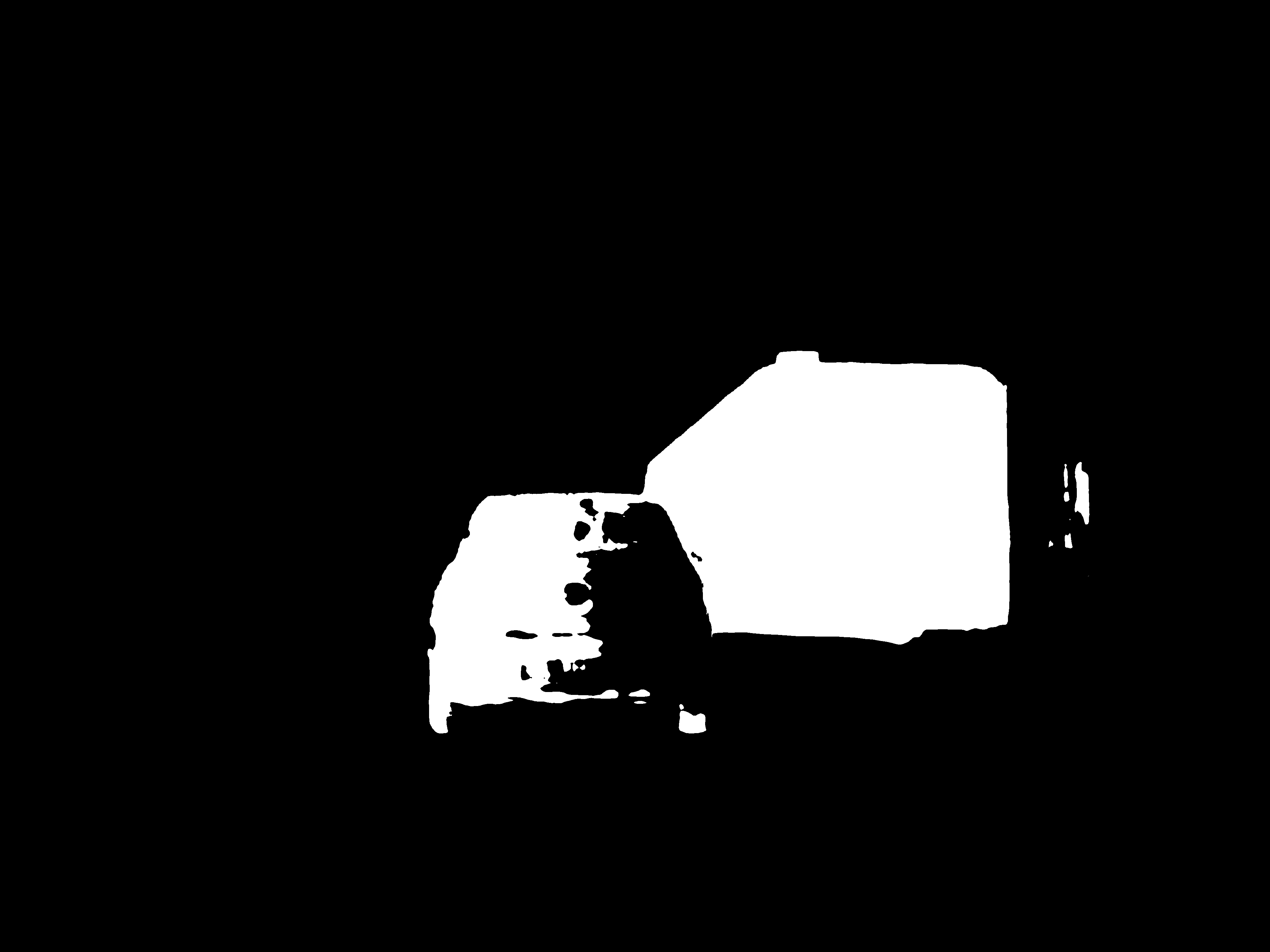}
\includegraphics[width=0.28\linewidth, trim={0 7.5cm 0 25cm},clip]{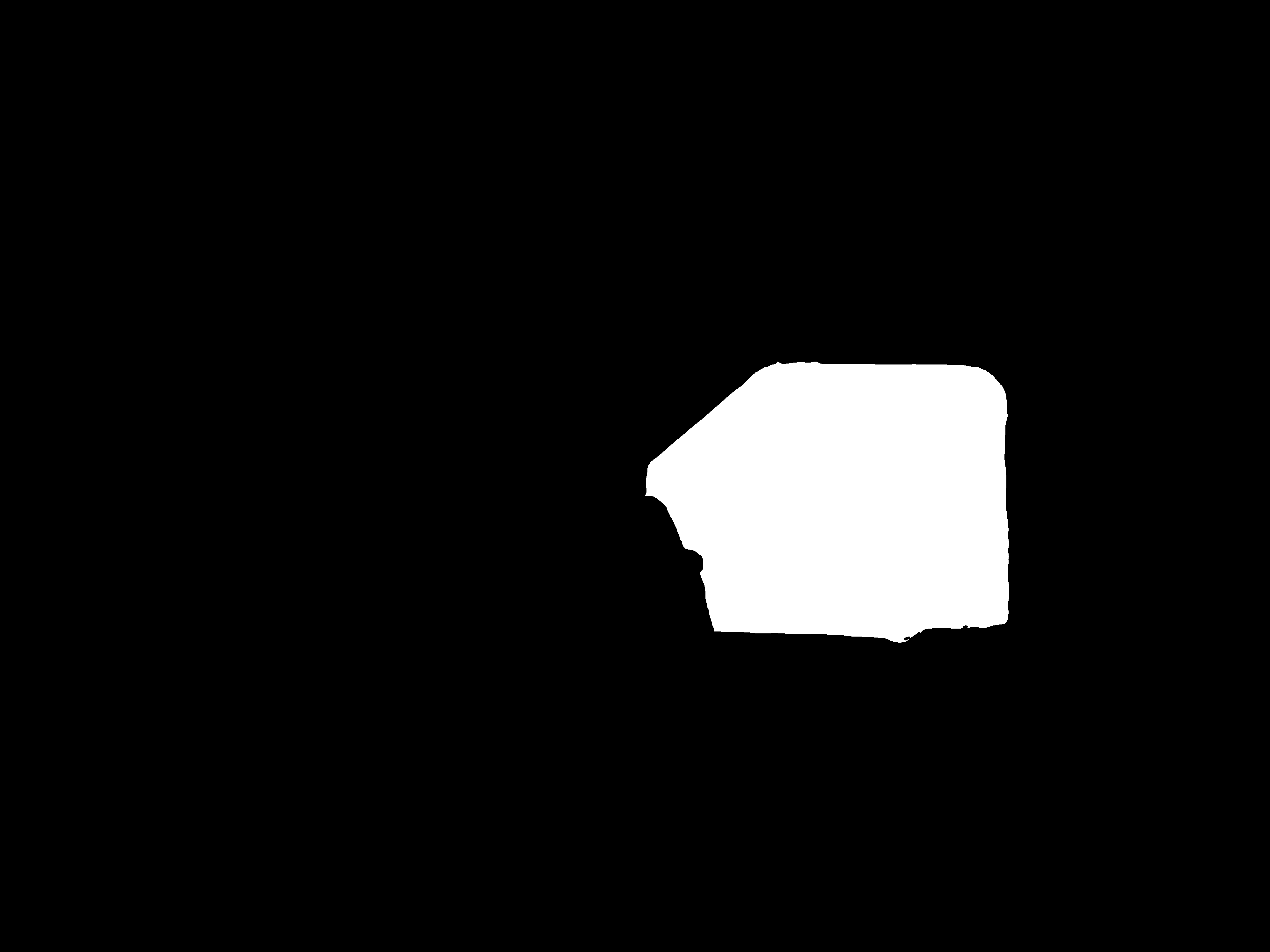}\\

 \begin{subfigure}[b]{0.28\textwidth}
     \centering
     \caption{Image with ground-truth segment}
 \end{subfigure}
 \begin{subfigure}[b]{0.28\textwidth}
     \centering
     \caption{Predicted segment \textbf{without IBS}}
 \end{subfigure}
  \begin{subfigure}[b]{0.28\textwidth}
     \centering
     \caption{Predicted segment \textbf{with IBS} (ours)}
 \end{subfigure}

\vspace{-10pt}
\caption{\textbf{Confusion problem for crop-based training of Panoptic FCN.} Predictions for individual thing instances with and without IBS. Top four images: Cityscapes \textit{val}; bottom four: Mapillary Vistas \textit{validation}. (b) The predictions by Panoptic FCN without IBS suffer from confusion, and (c) IBS largely solves this problem, leading to more accurate predictions. Full panoptic results for these images are shown in \Cref{fig:results_panfcn_overall}.} 
\label{fig:results_panfcn_problem}
\end{figure*}

\begin{figure*}[t]
\centering
\includegraphics[width=0.28\linewidth]{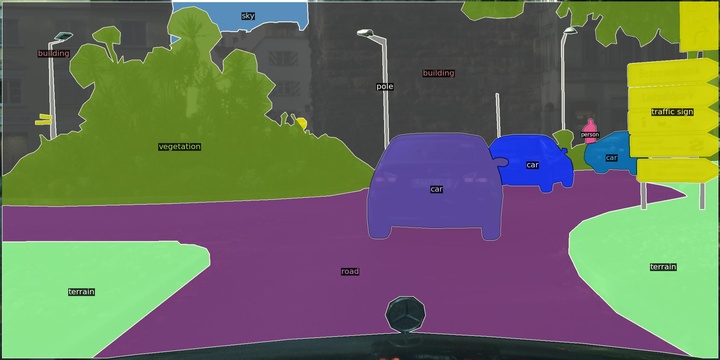}
\includegraphics[width=0.28\linewidth]{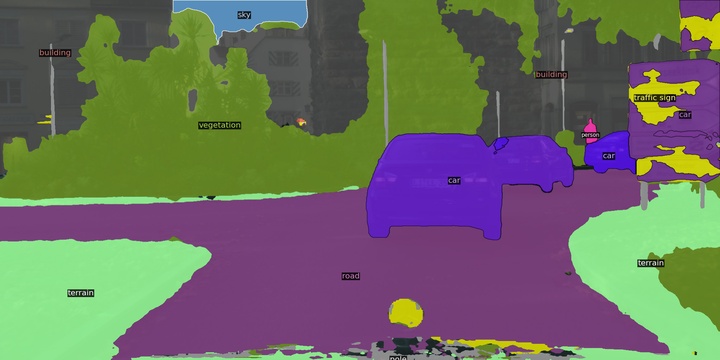}
\includegraphics[width=0.28\linewidth]{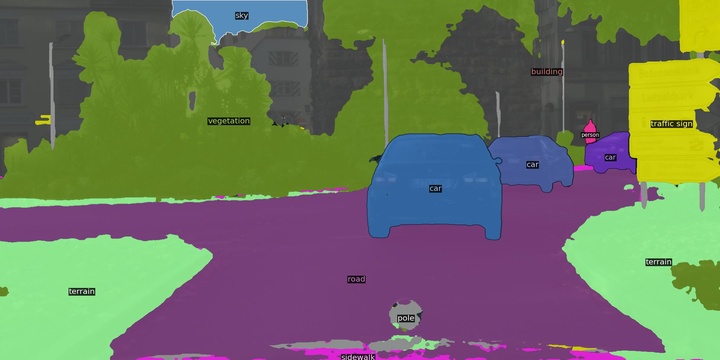}
\\

\includegraphics[width=0.28\linewidth]{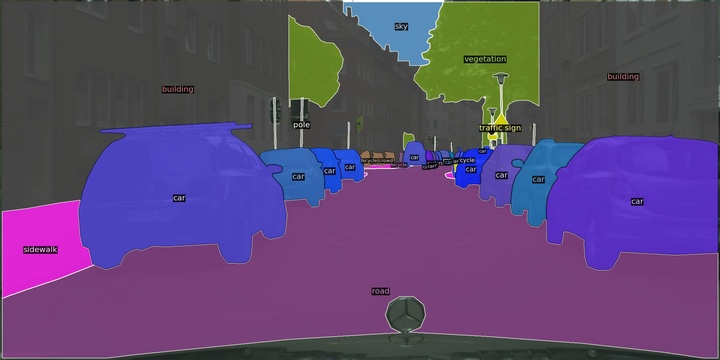}
\includegraphics[width=0.28\linewidth]{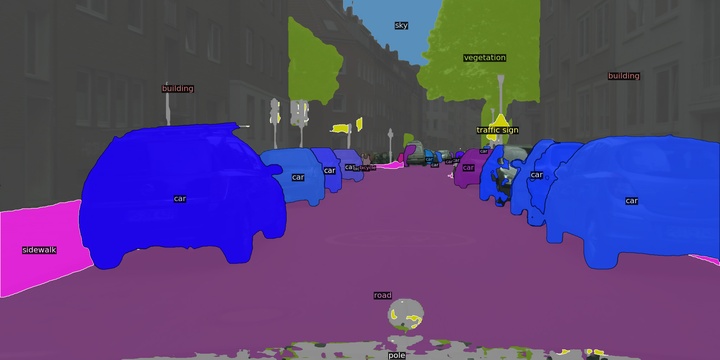}
\includegraphics[width=0.28\linewidth]{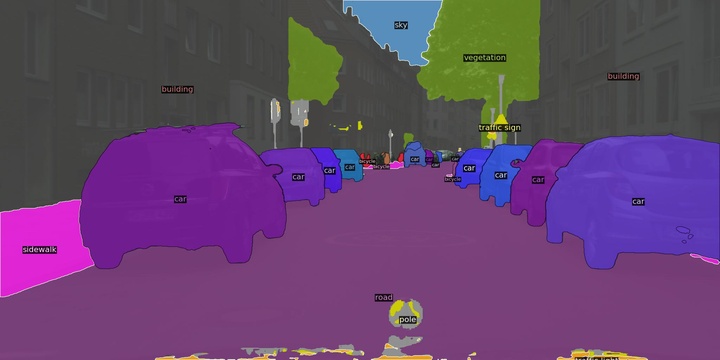}
\\

\includegraphics[width=0.28\linewidth]{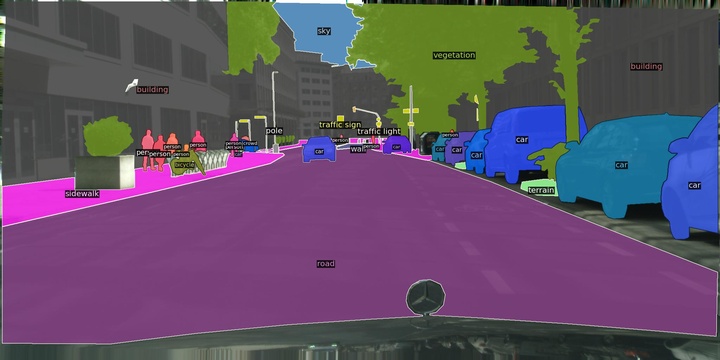}
\includegraphics[width=0.28\linewidth]{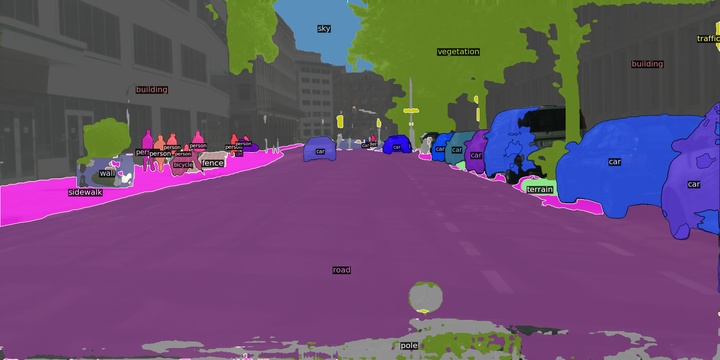}
\includegraphics[width=0.28\linewidth]{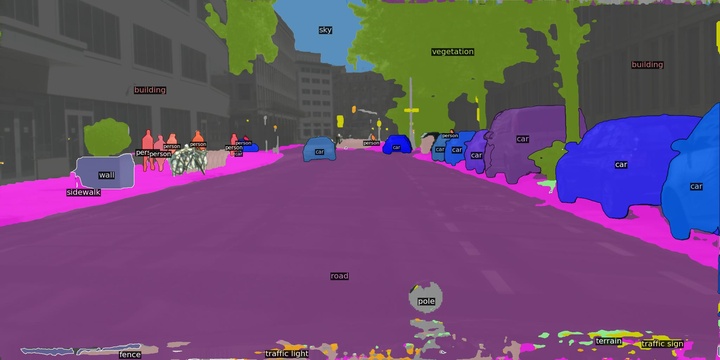}
\\

\includegraphics[width=0.28\linewidth]{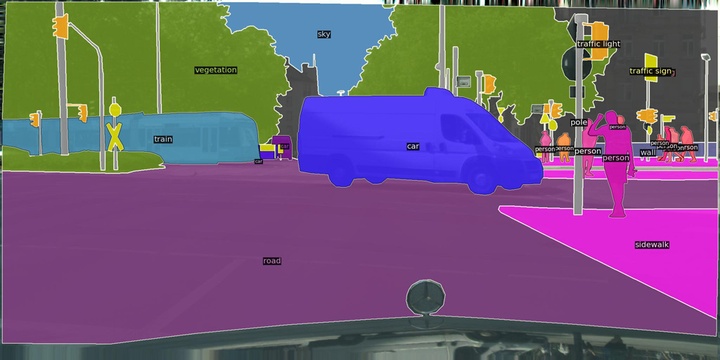}
\includegraphics[width=0.28\linewidth]{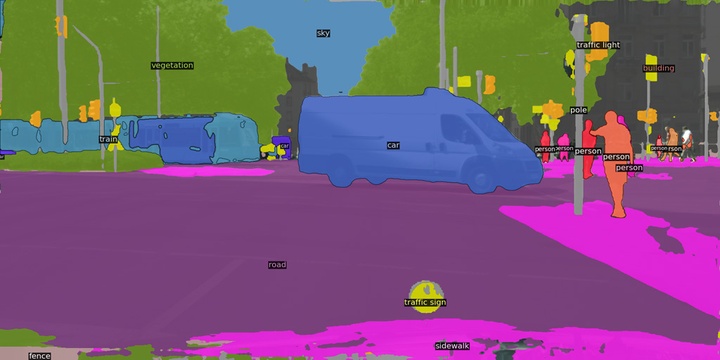}
\includegraphics[width=0.28\linewidth]{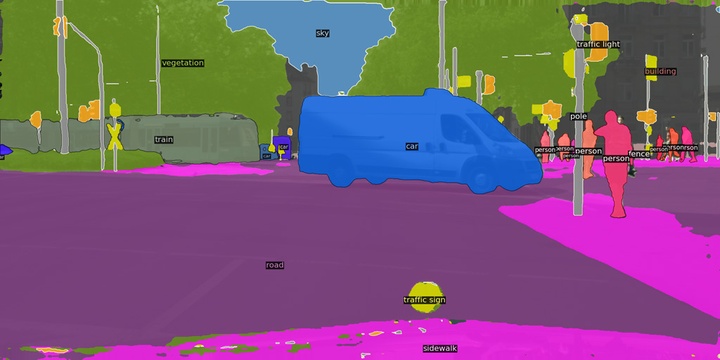}
\\

\includegraphics[width=0.28\linewidth]{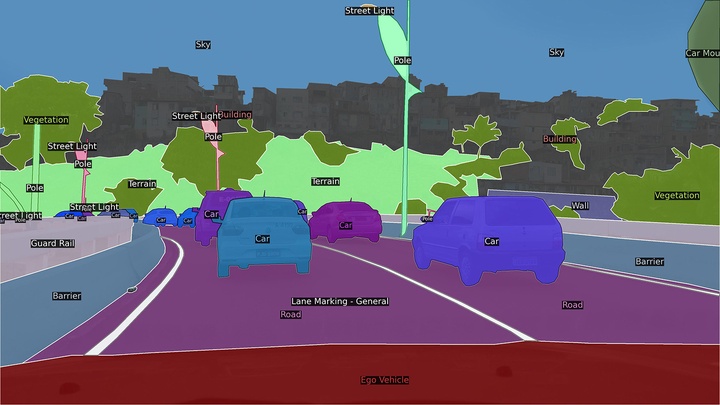}
\includegraphics[width=0.28\linewidth]{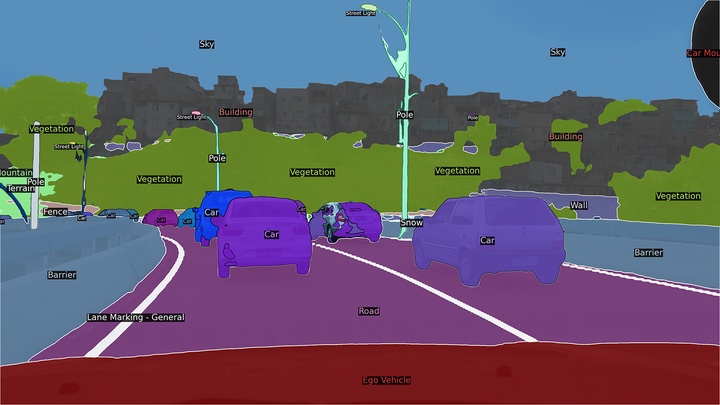}
\includegraphics[width=0.28\linewidth]{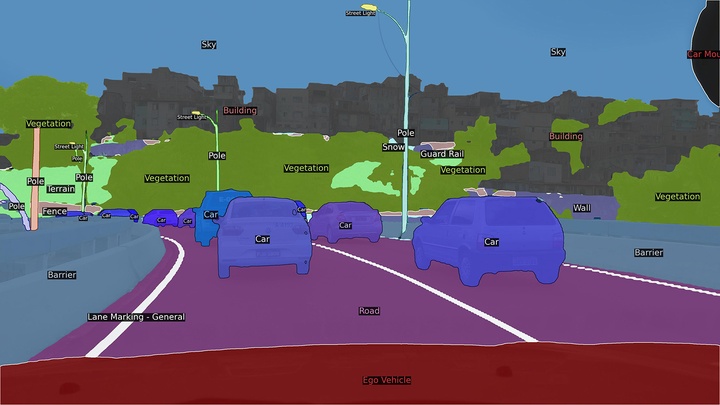}
\\

\includegraphics[width=0.28\linewidth, trim={0 0 0 5.5130cm},clip]{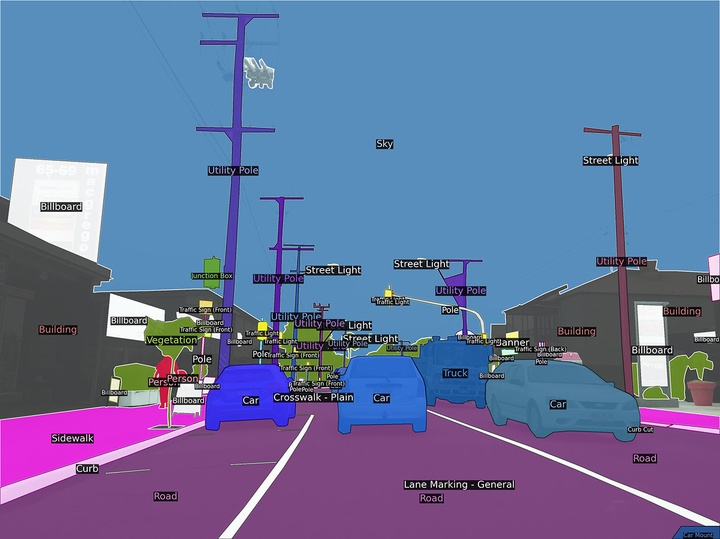}
\includegraphics[width=0.28\linewidth, trim={0 0 0 5.5130cm},clip]{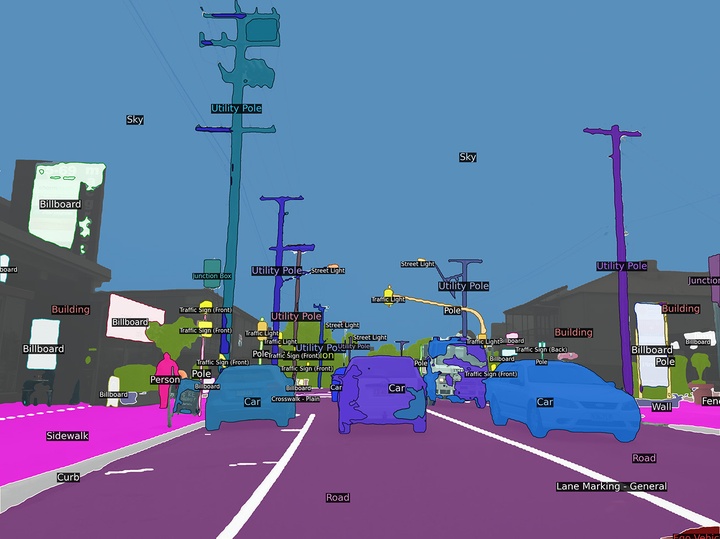}
\includegraphics[width=0.28\linewidth, trim={0 0 0 5.5130cm},clip]{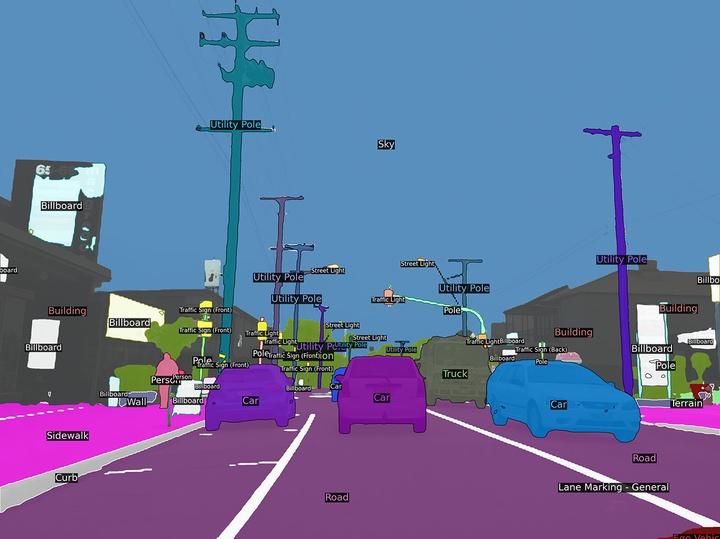}
\\

\includegraphics[width=0.28\linewidth]{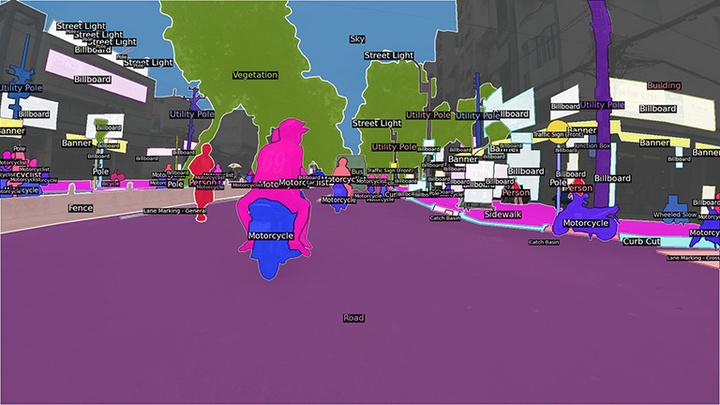}
\includegraphics[width=0.28\linewidth]{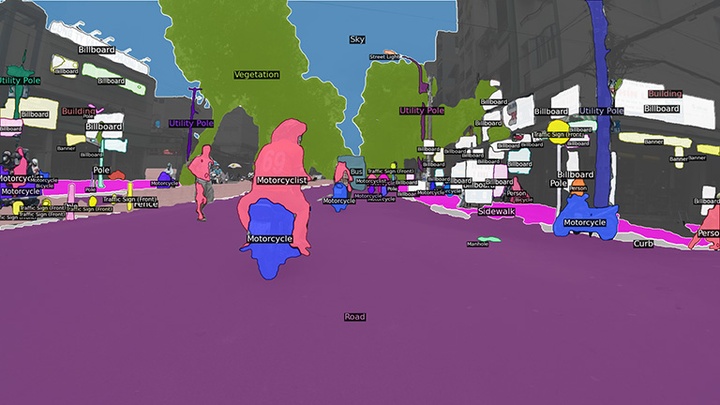}
\includegraphics[width=0.28\linewidth]{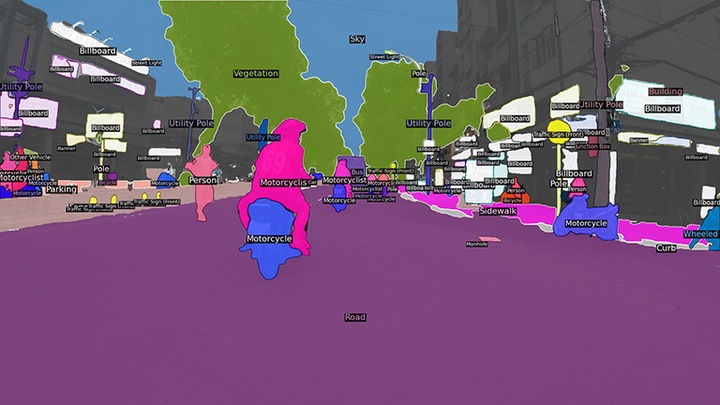}
\\

\includegraphics[width=0.28\linewidth, trim={0 1.3542cm 0 4.514cm},clip]{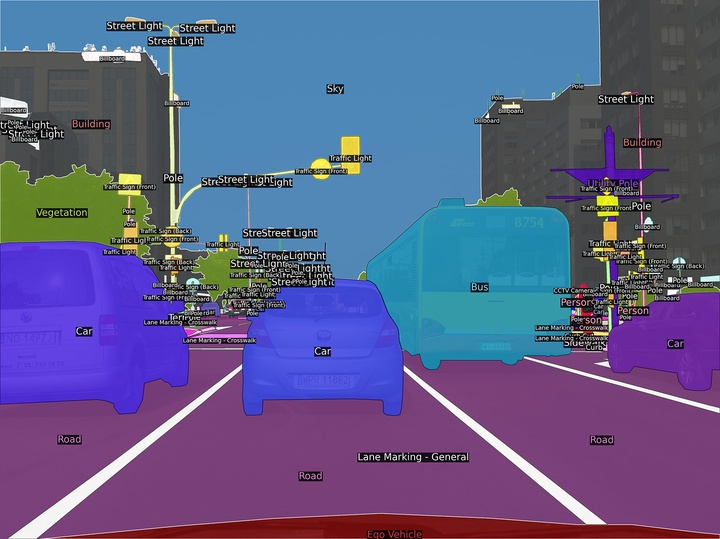}
\includegraphics[width=0.28\linewidth, trim={0 1.3542cm 0 4.514cm},clip]{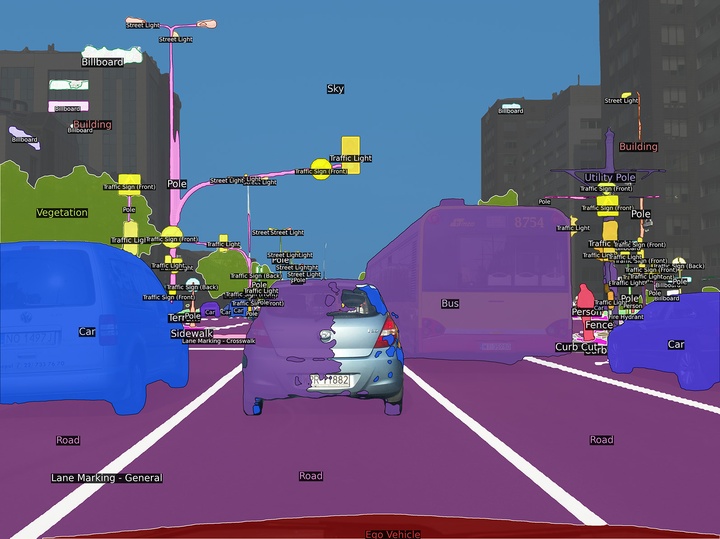}
\includegraphics[width=0.28\linewidth, trim={0 1.3542cm 0 4.514cm},clip]{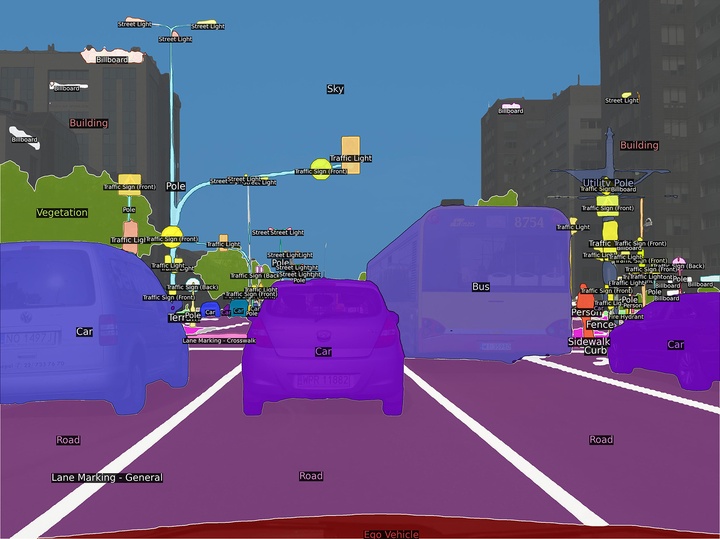}
\\

\begin{subfigure}[b]{0.28\textwidth}
 \centering
 \caption{Ground truth}
\end{subfigure}
\begin{subfigure}[b]{0.28\textwidth}
 \centering
 \caption{Without IBS}
\end{subfigure}
\begin{subfigure}[b]{0.28\textwidth}
 \centering
 \caption{With IBS (ours)}
\end{subfigure}

\vspace{-10pt}
\caption{\textbf{Intra-Batch Supervision on Panoptic FCN.} Top four images: Cityscapes \textit{val}; bottom four: Mapillary Vistas \textit{validation}. Each segment is indicated with a unique color and text label, so confusion can be observed when multiple thing instances share a color or text label. Individual thing predictions for these images are shown in \Cref{fig:results_panfcn_problem}. Best viewed digitally.}  
\label{fig:results_panfcn_overall}
\end{figure*}

\begin{figure*}[t]
\centering
\includegraphics[width=0.28\linewidth]{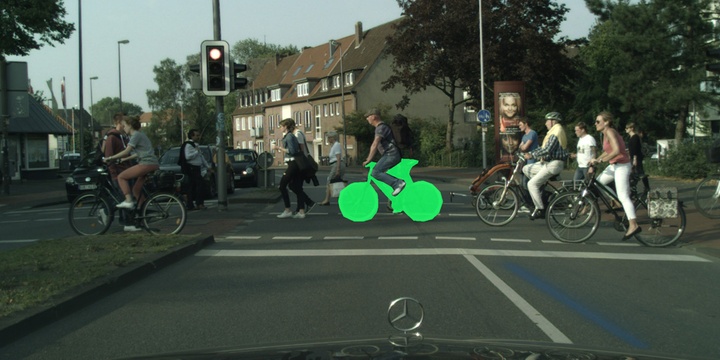}
\includegraphics[width=0.28\linewidth]{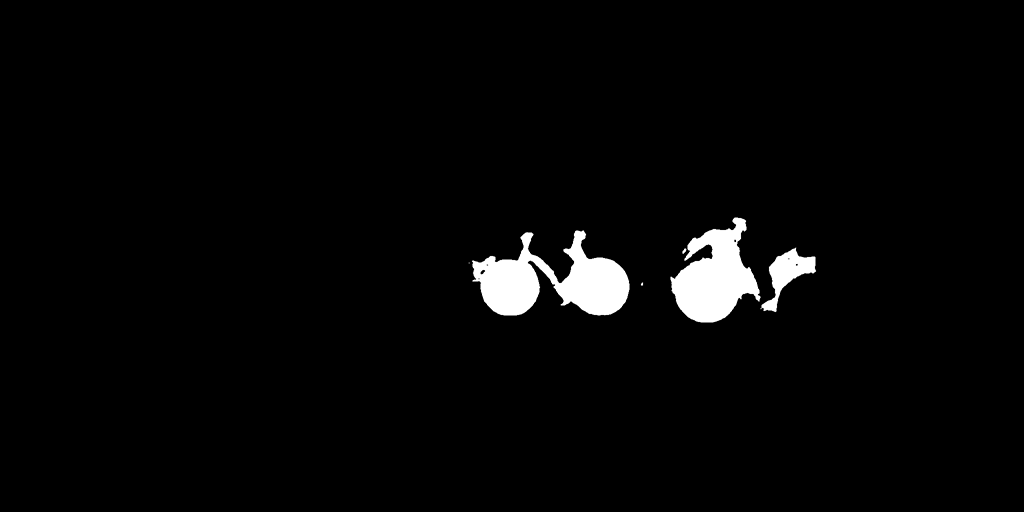}
\includegraphics[width=0.28\linewidth]{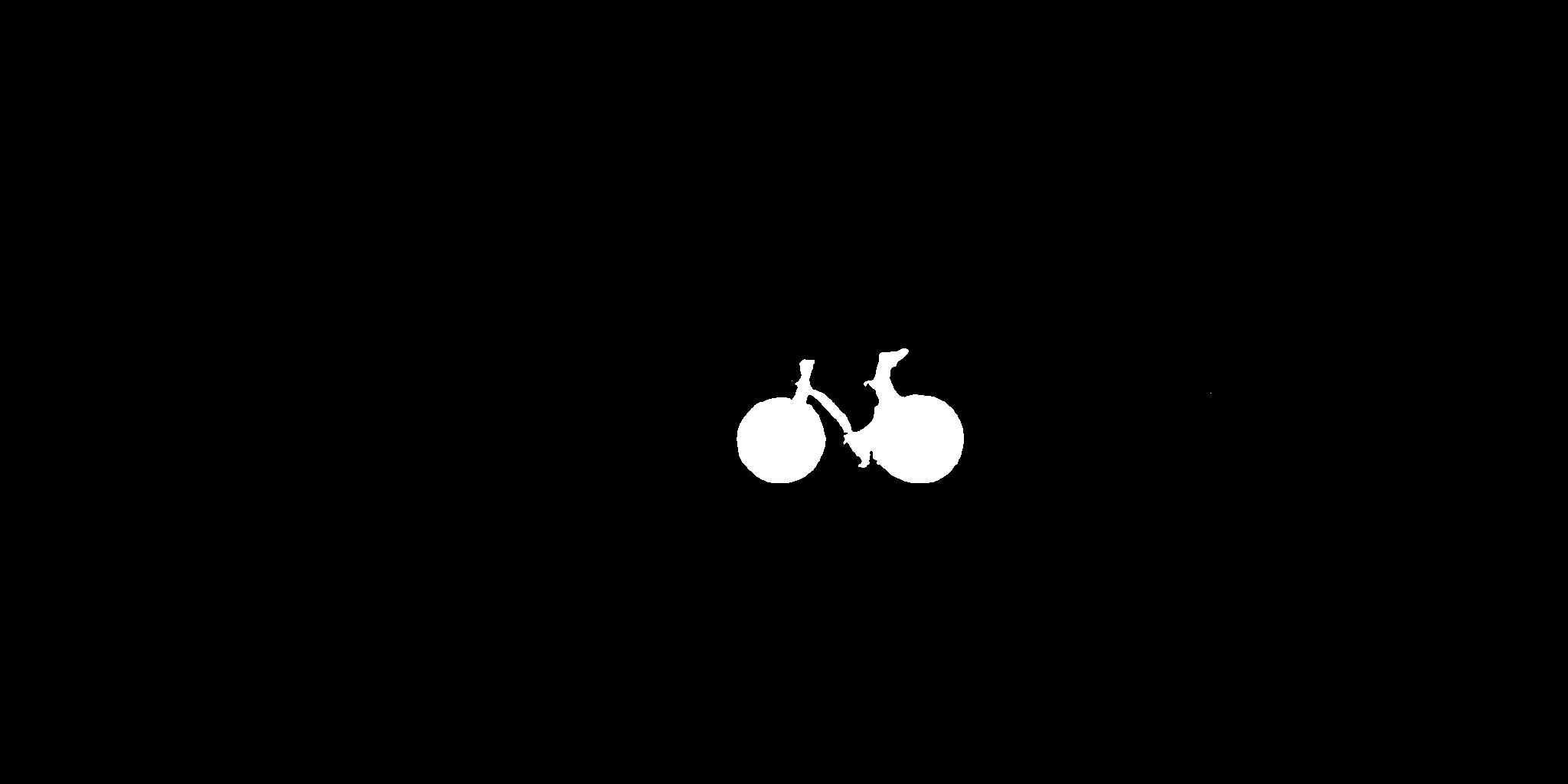}\\

\includegraphics[width=0.28\linewidth]{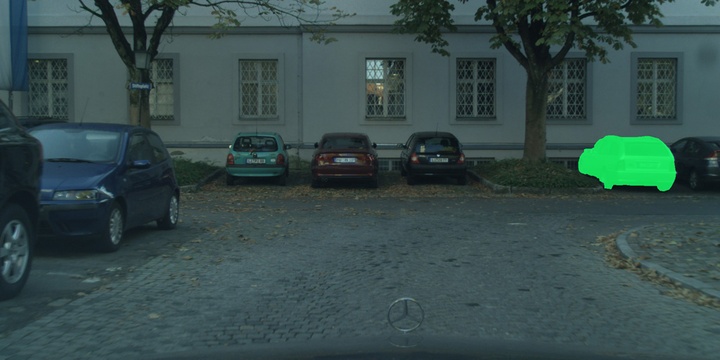}
\includegraphics[width=0.28\linewidth]{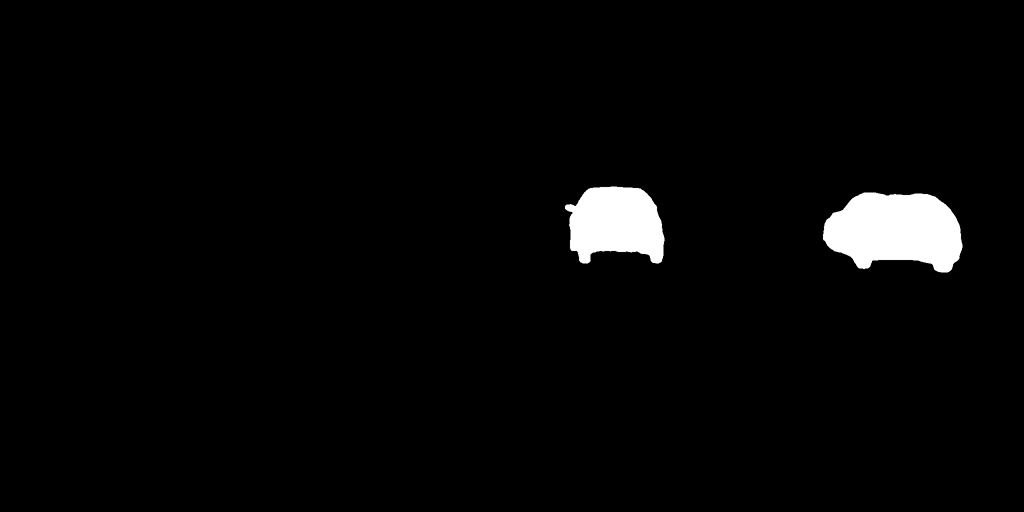}
\includegraphics[width=0.28\linewidth]{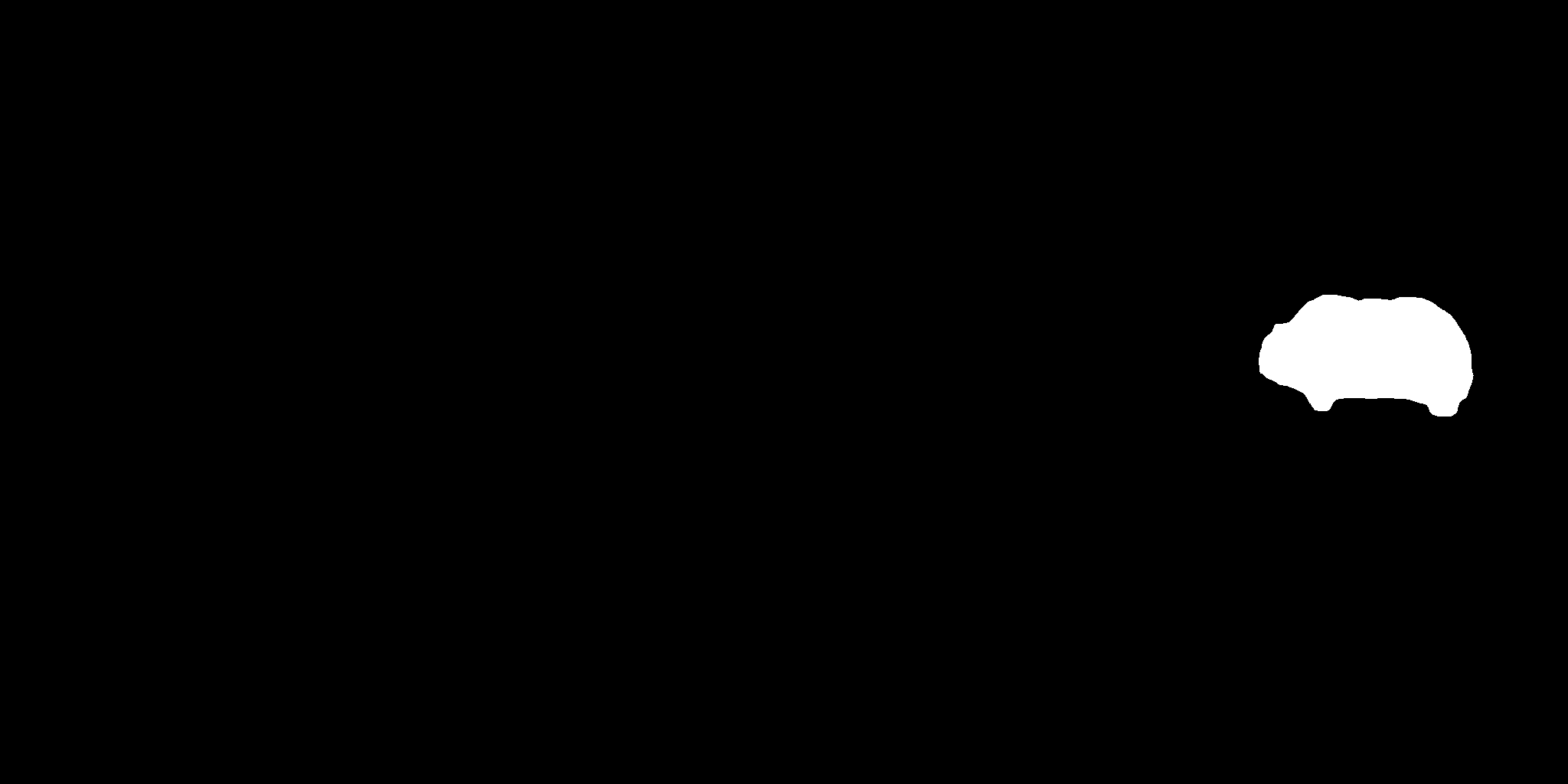}\\

\includegraphics[width=0.28\linewidth]{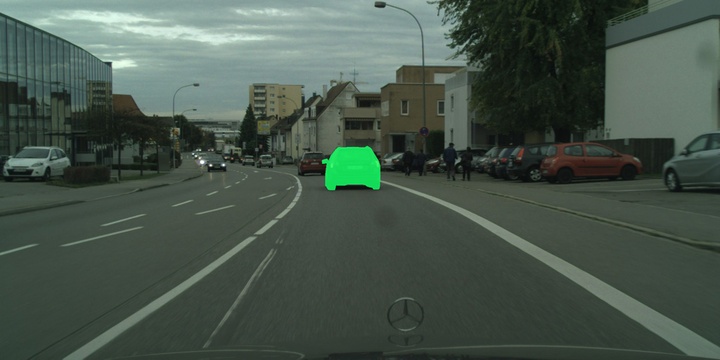}
\includegraphics[width=0.28\linewidth]{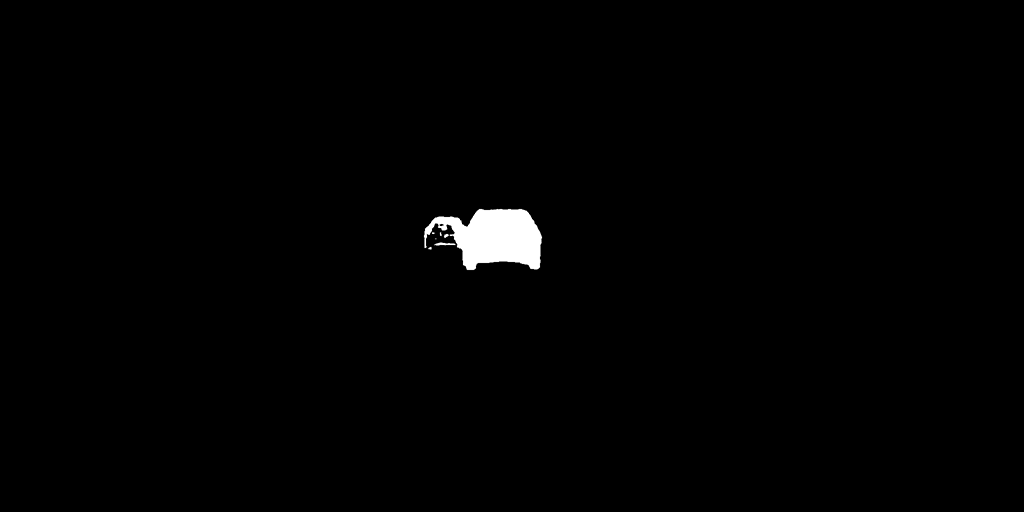}
\includegraphics[width=0.28\linewidth]{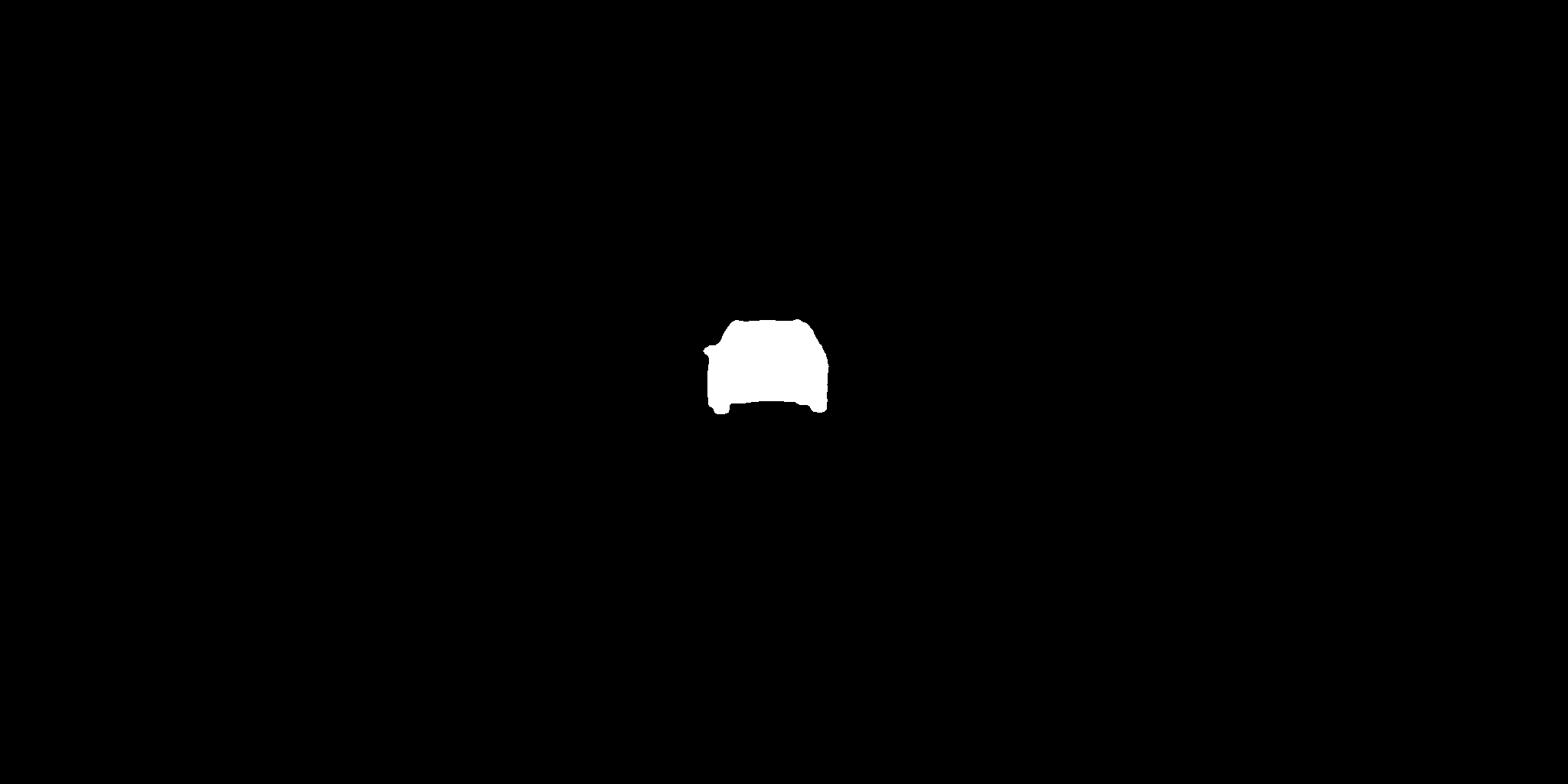}\\

\includegraphics[width=0.28\linewidth]{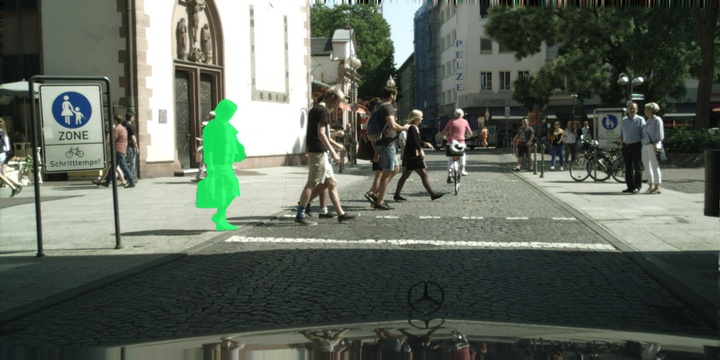}
\includegraphics[width=0.28\linewidth]{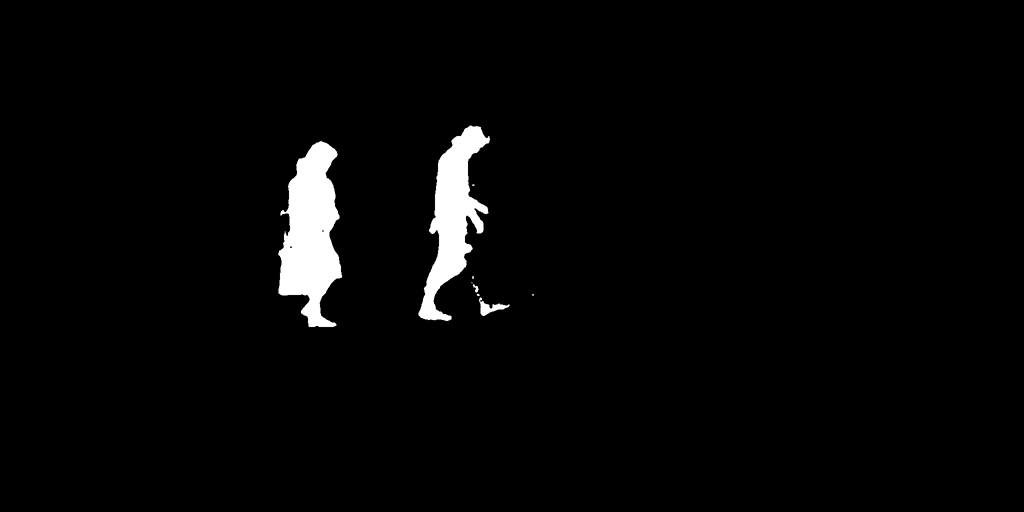}
\includegraphics[width=0.28\linewidth]{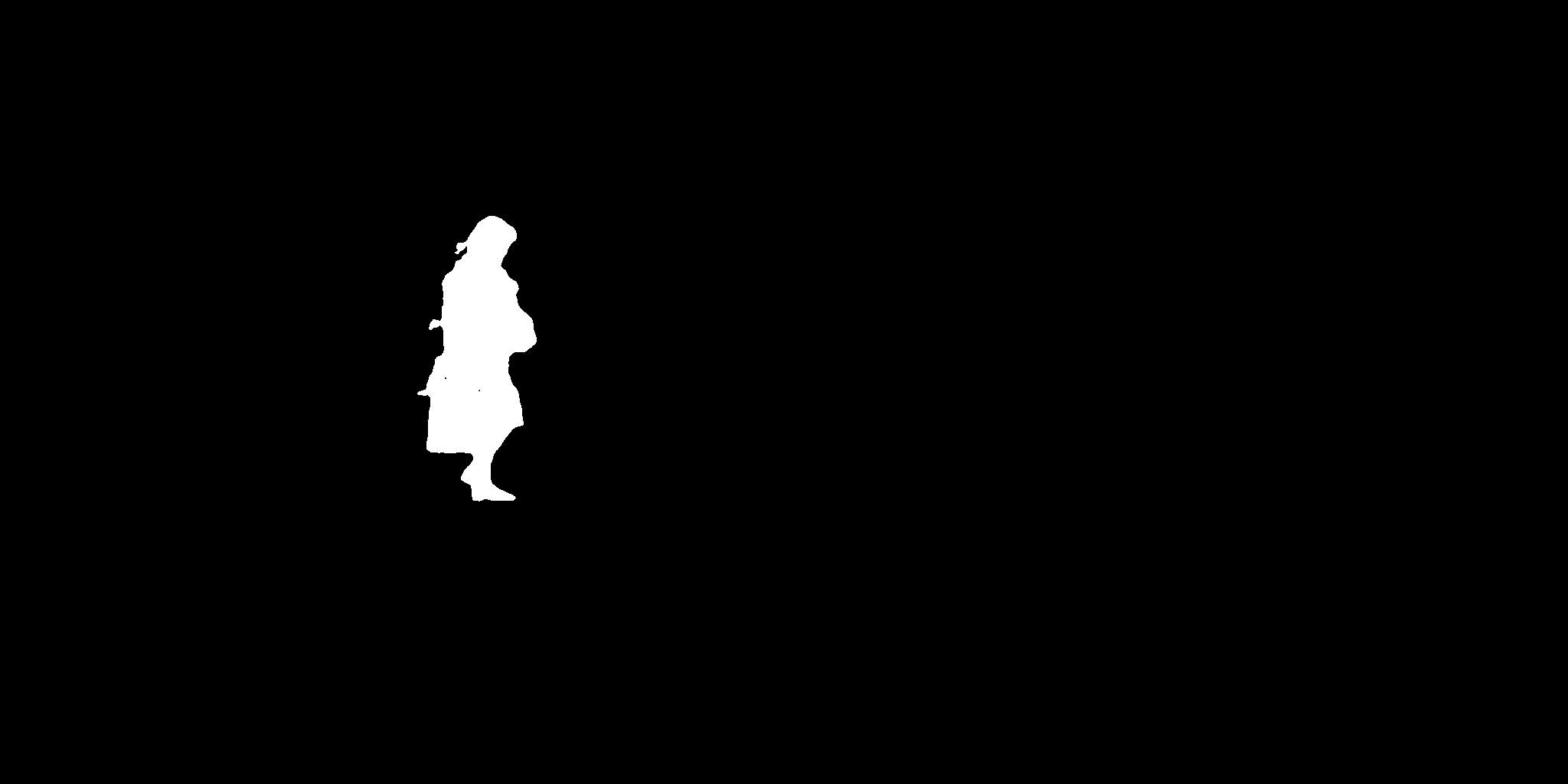}\\

\includegraphics[width=0.28\linewidth]{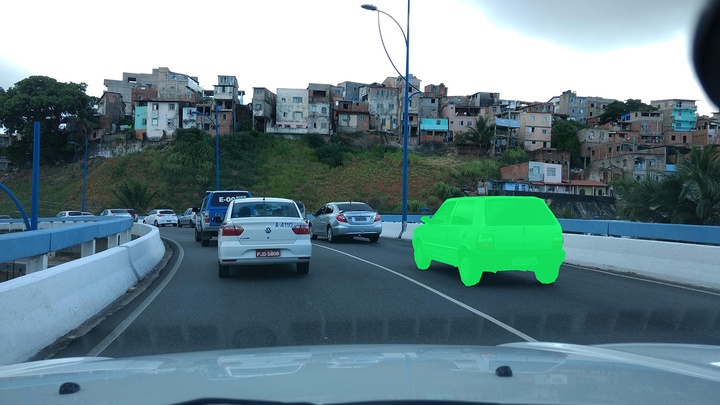}
\includegraphics[width=0.28\linewidth]{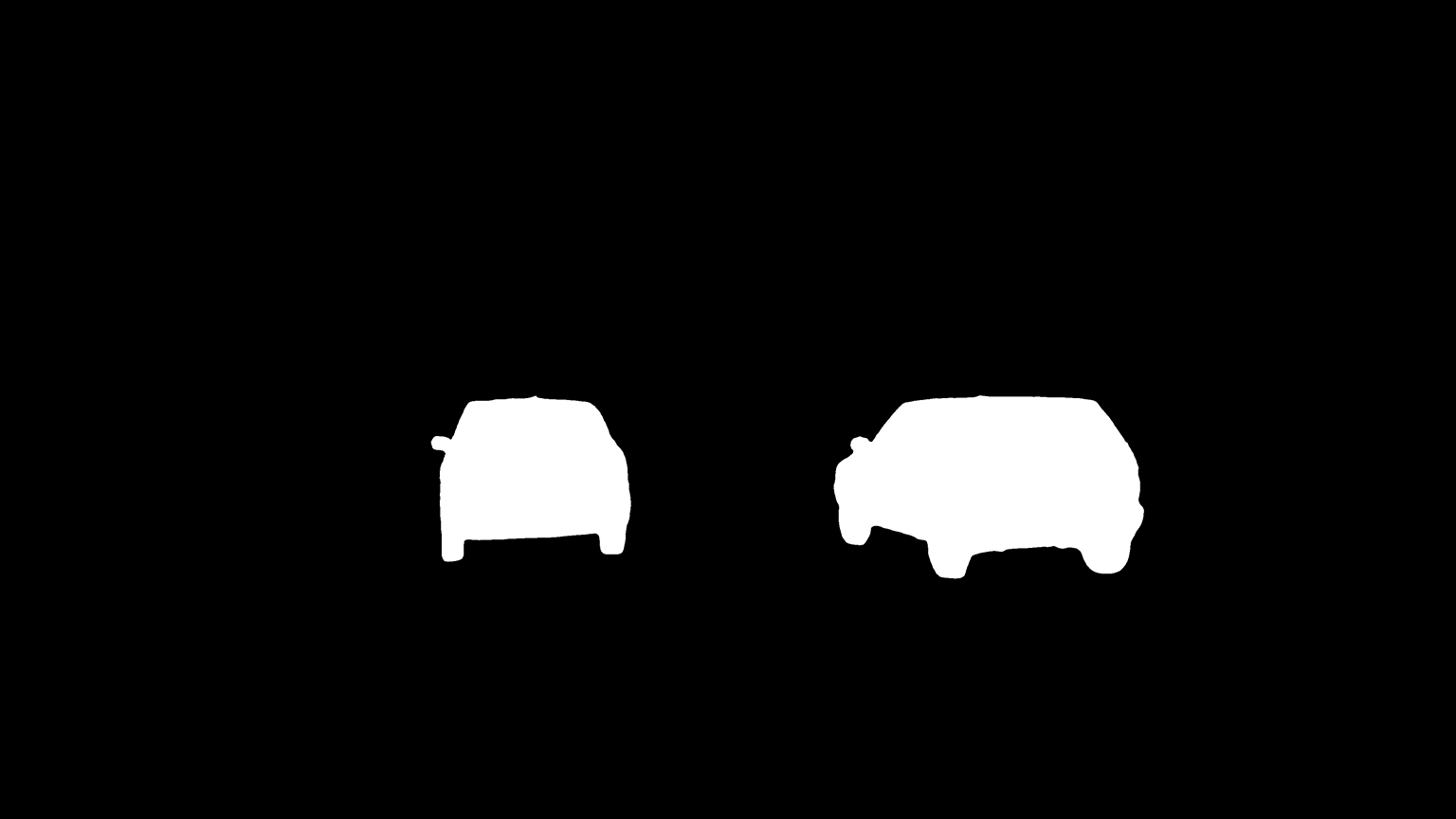}
\includegraphics[width=0.28\linewidth]{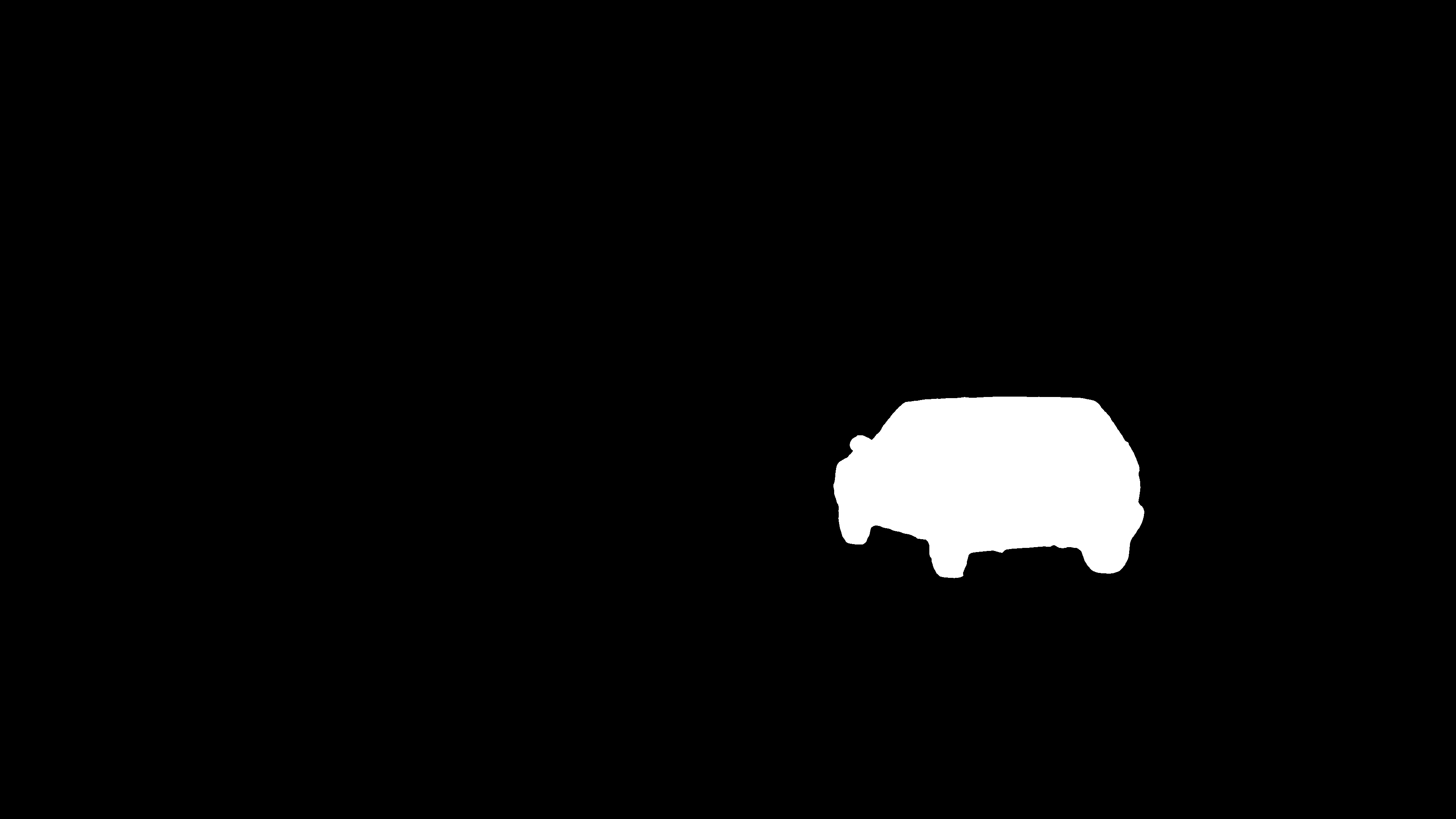}\\
\includegraphics[width=0.28\linewidth, trim={0 0 0 4.4720cm},clip]{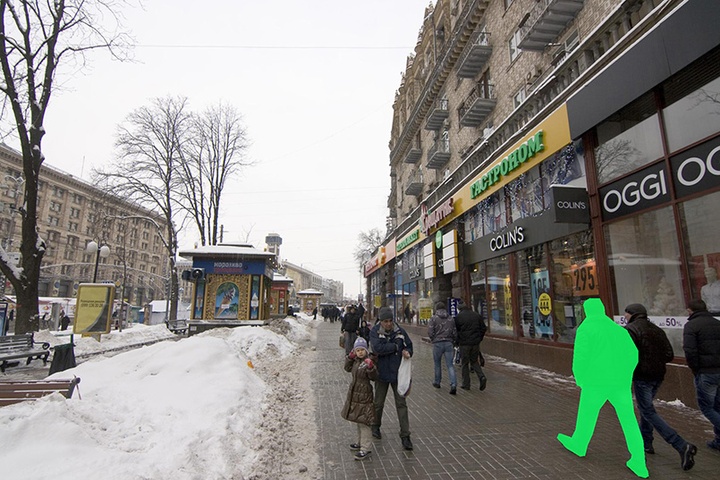}
\includegraphics[width=0.28\linewidth, trim={0 0 0 5cm},clip]{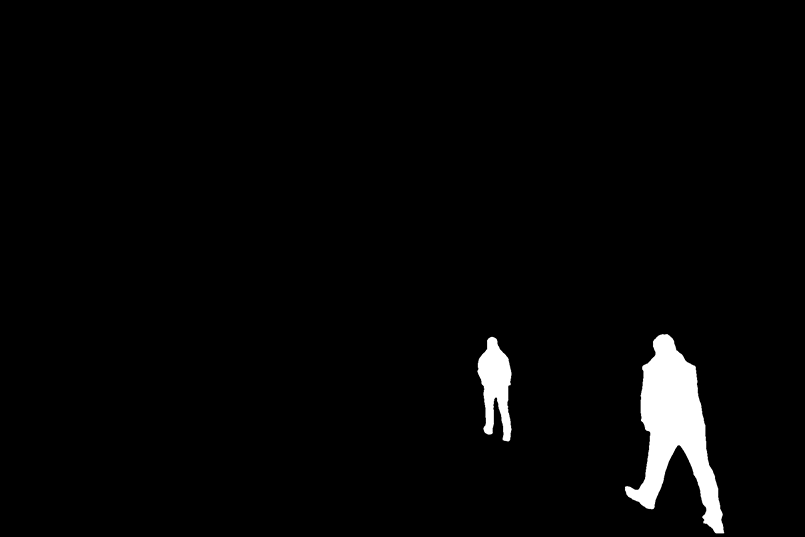}
\includegraphics[width=0.28\linewidth, trim={0 0 0 12.5cm},clip]{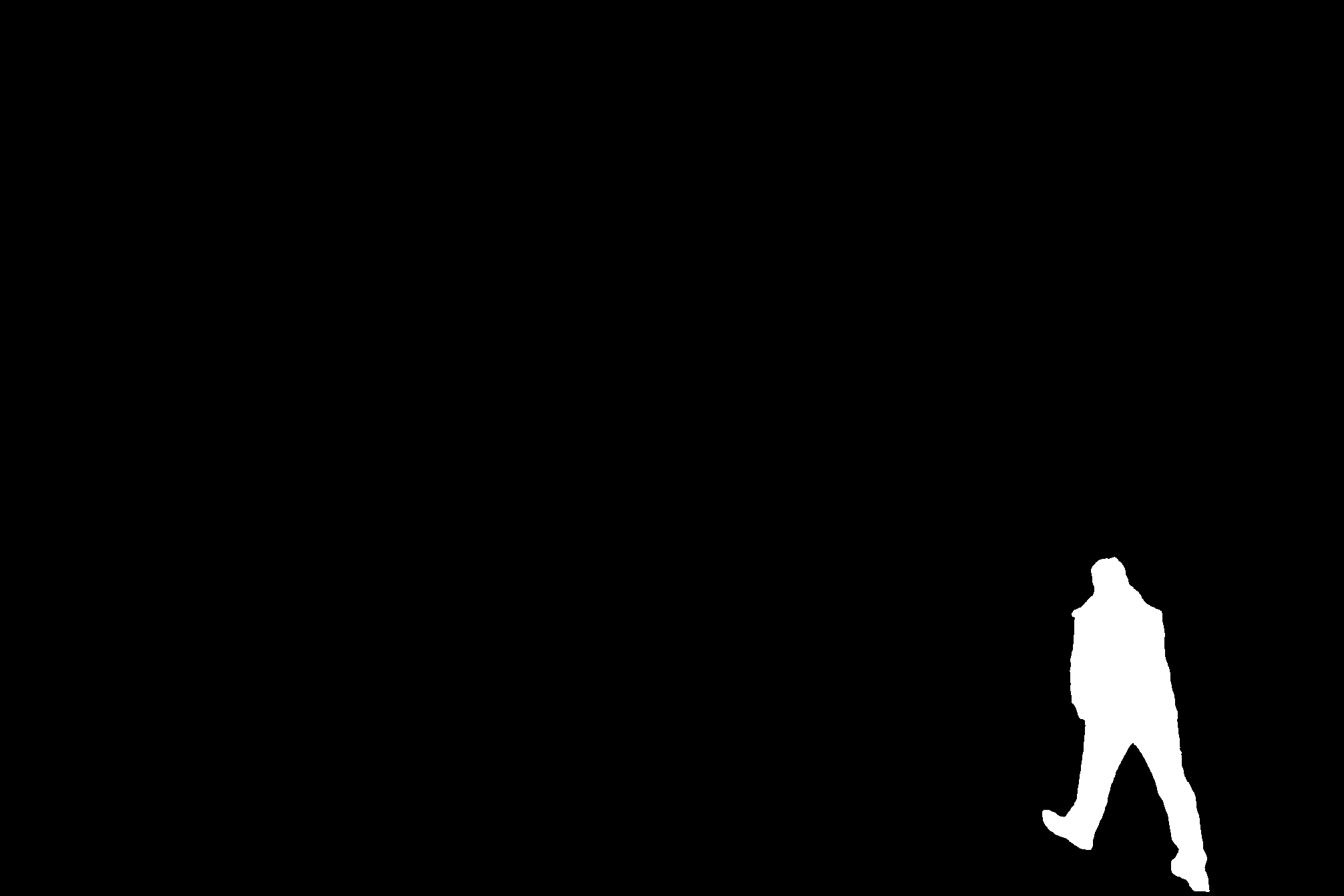}\\
\includegraphics[width=0.28\linewidth]{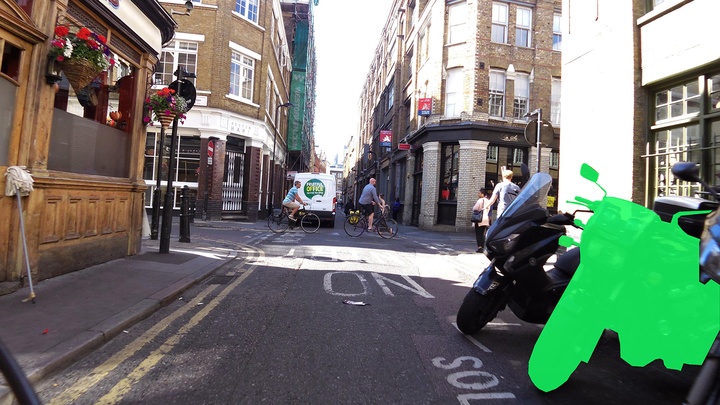}
\includegraphics[width=0.28\linewidth]{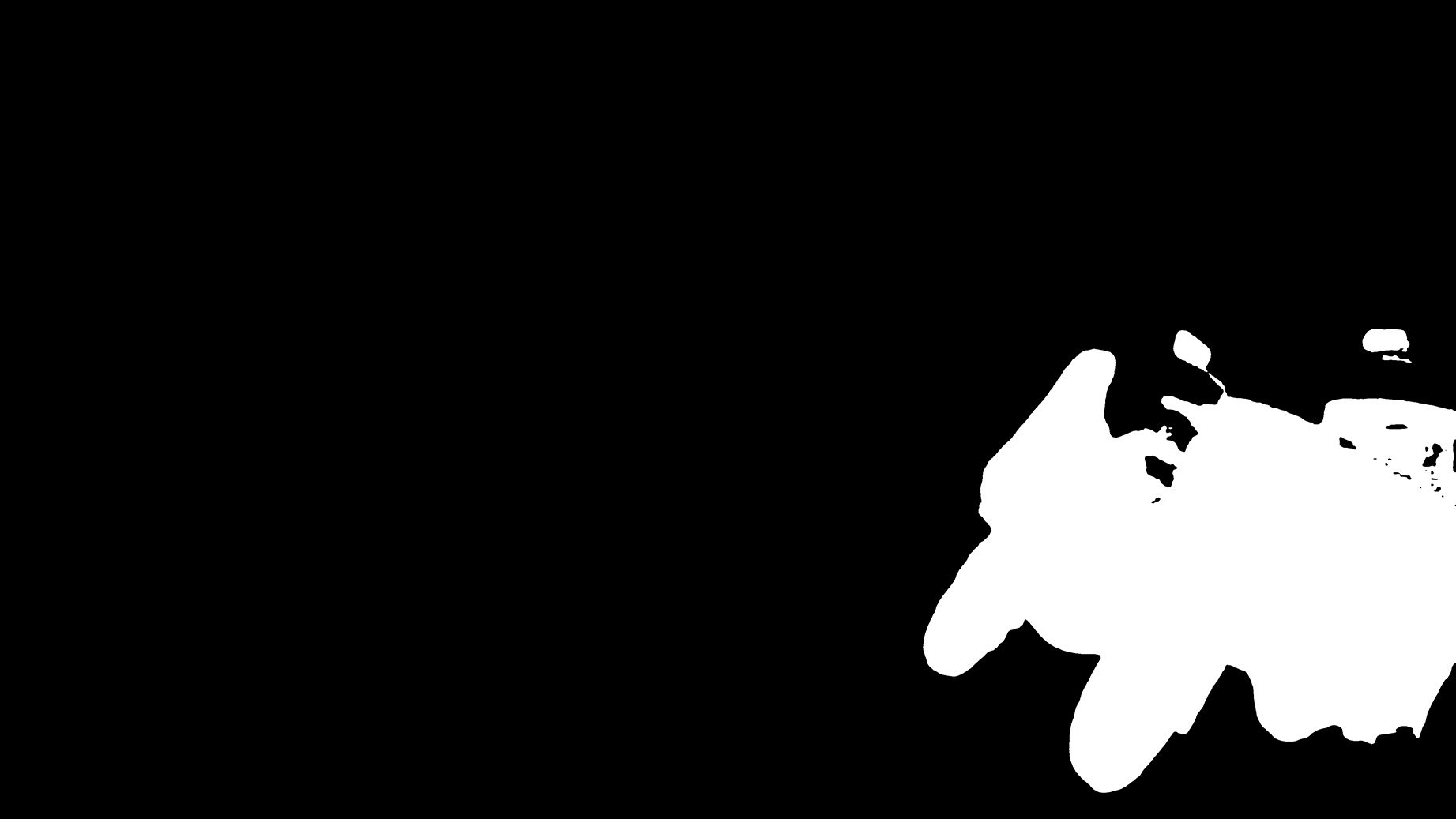}
\includegraphics[width=0.28\linewidth]{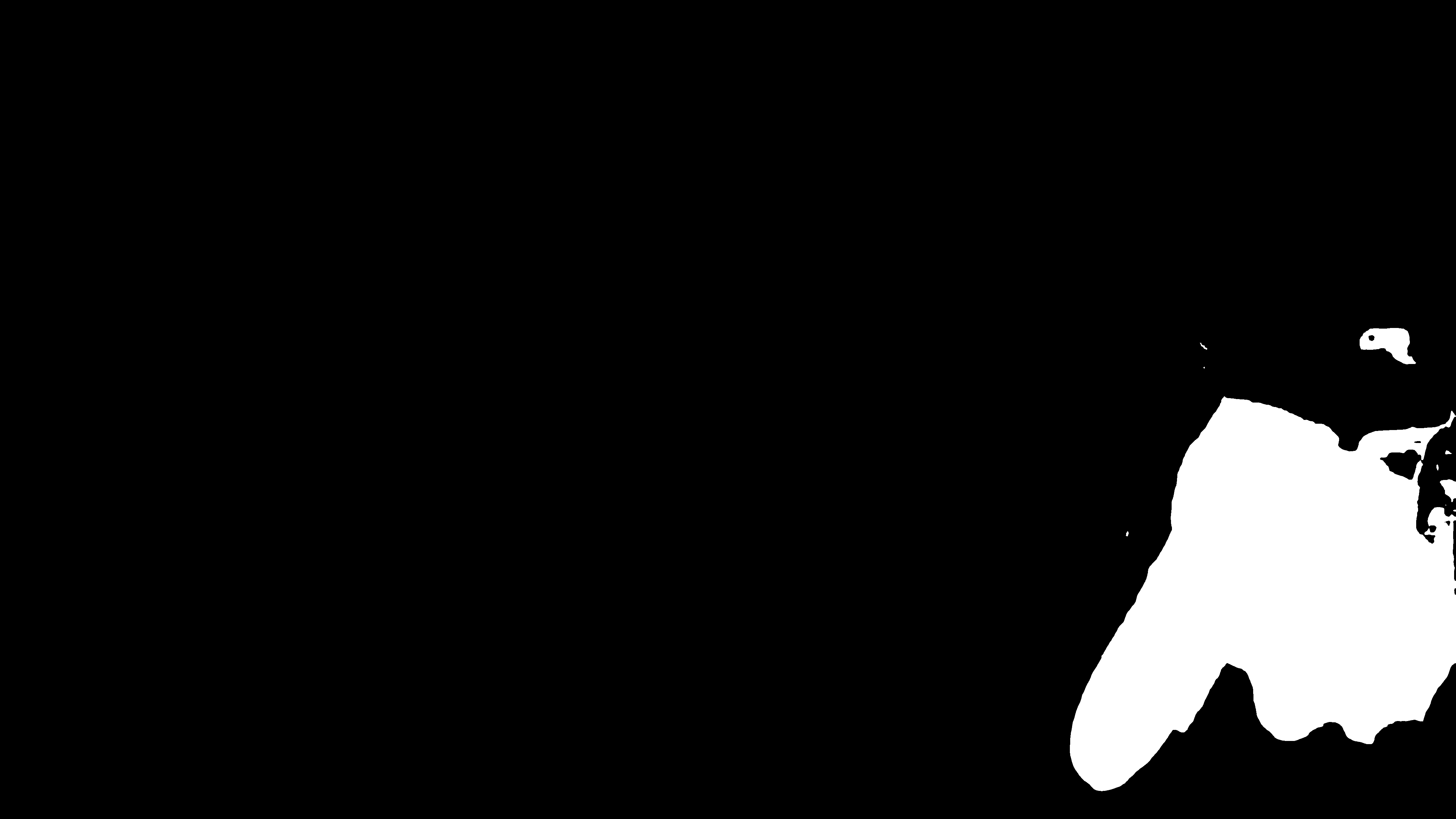}\\
\includegraphics[width=0.28\linewidth, trim={0 0 0 5.5130cm},clip]{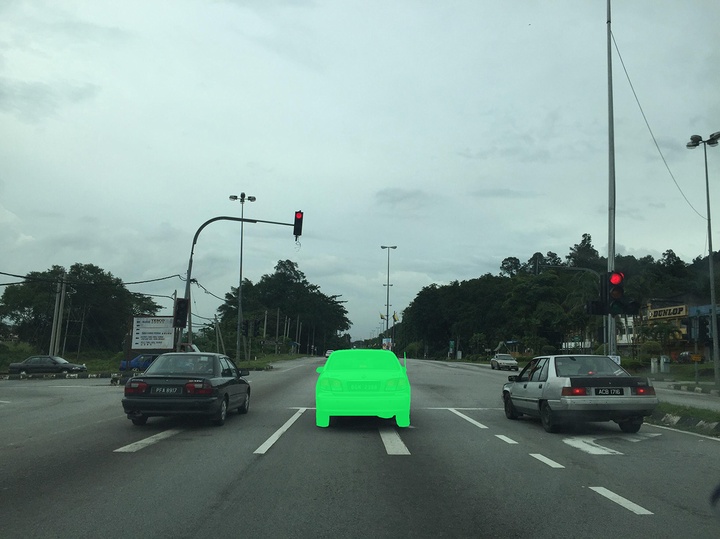}
\includegraphics[width=0.28\linewidth, trim={0 0 0 10cm},clip]{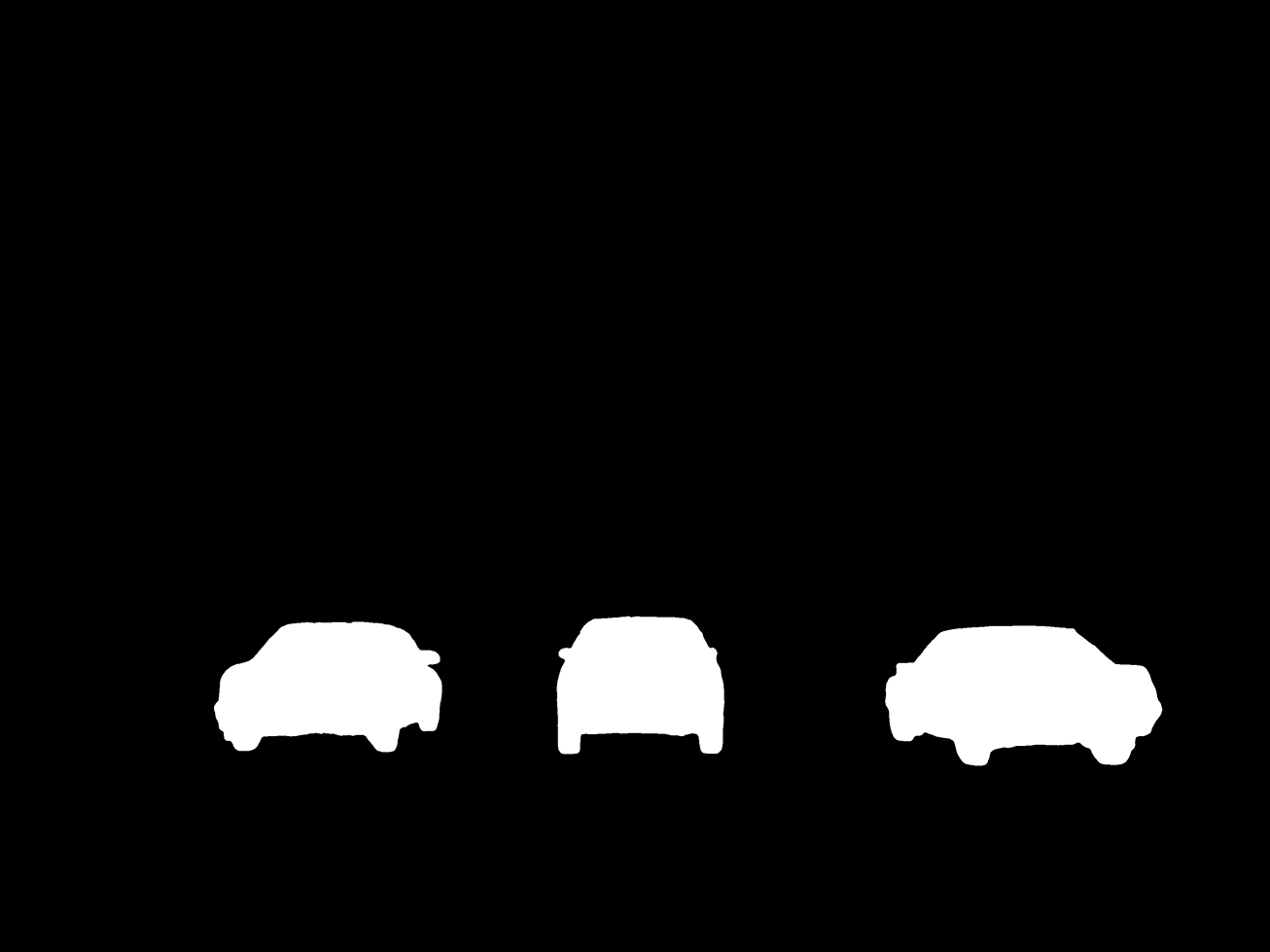}
\includegraphics[width=0.28\linewidth, trim={0 0 0 25cm},clip]{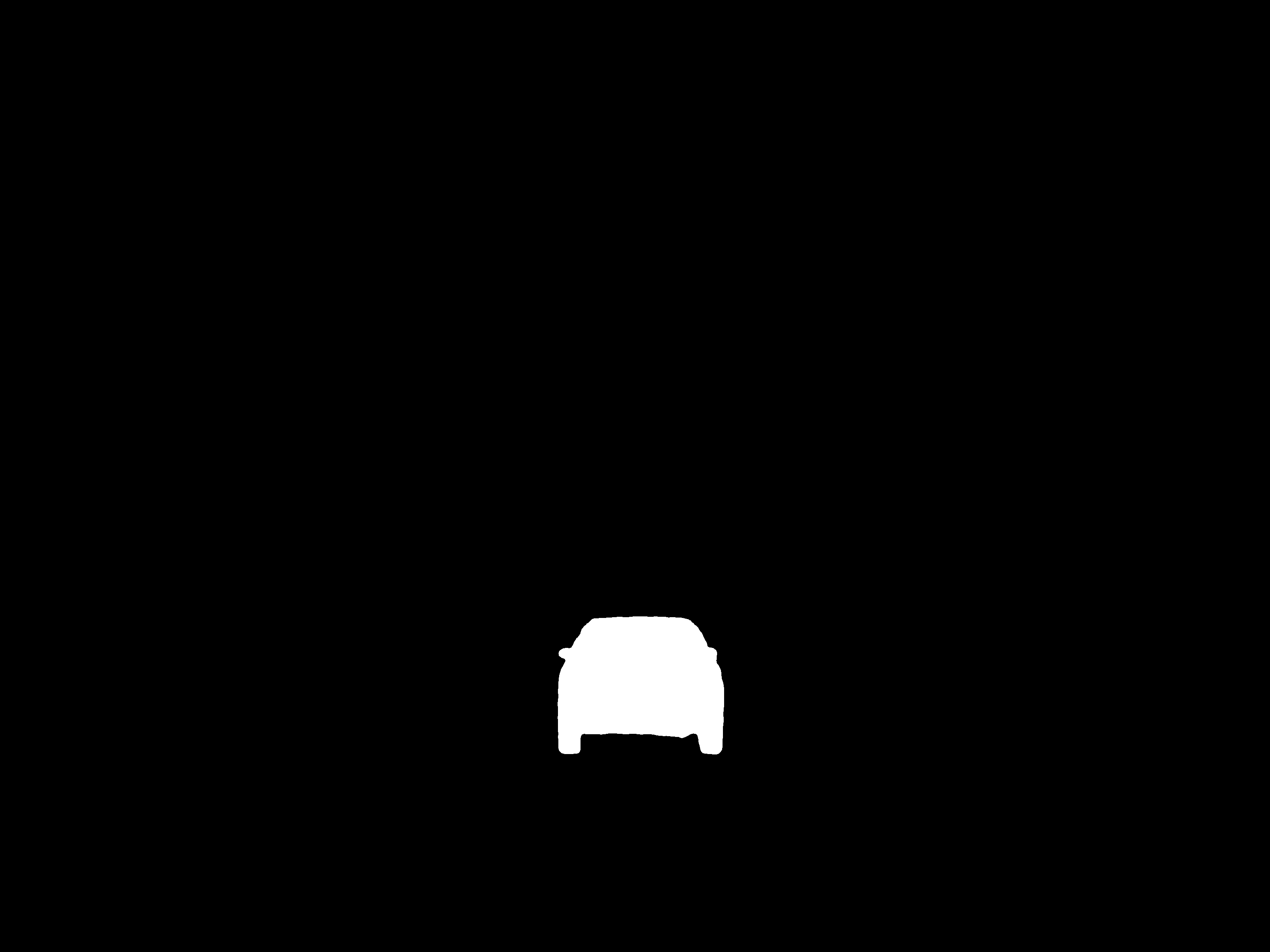}\\

 \begin{subfigure}[b]{0.28\textwidth}
     \centering
     \caption{Image with ground-truth segment}
 \end{subfigure}
 \begin{subfigure}[b]{0.28\textwidth}
     \centering
     \caption{Predicted segment \textbf{without IBS}}
 \end{subfigure}
  \begin{subfigure}[b]{0.28\textwidth}
     \centering
     \caption{Predicted segment \textbf{with IBS} (ours)}
 \end{subfigure}

\vspace{-10pt}
\caption{\textbf{Confusion problem for crop-based training of Mask2Former.} Predictions for individual thing instances with and without IBS. Top four images: Cityscapes \textit{val}; bottom four: Mapillary Vistas \textit{validation}. (b) The predictions by Mask2Former without IBS suffer from confusion, and (c) IBS largely solves this problem, leading to more accurate predictions. Full panoptic results for these images are shown in \Cref{fig:results_m2f_overall}.} 
\label{fig:results_m2f_problem}
\end{figure*}

\begin{figure*}[t]
\centering
\includegraphics[width=0.28\linewidth]{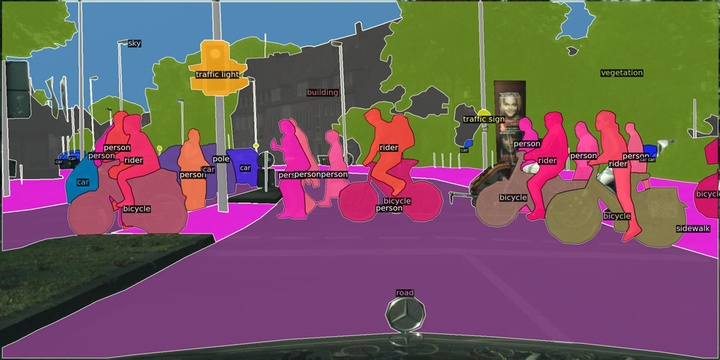}
\includegraphics[width=0.28\linewidth]{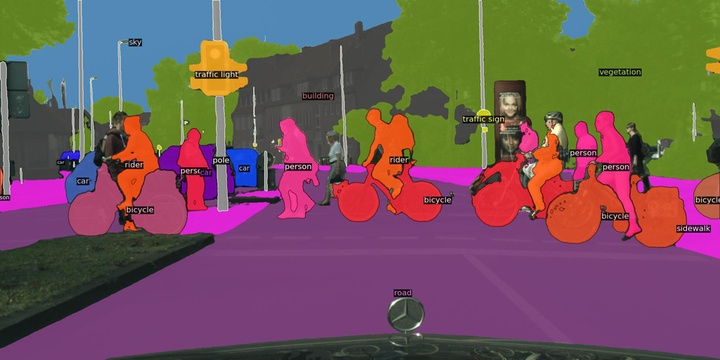}
\includegraphics[width=0.28\linewidth]{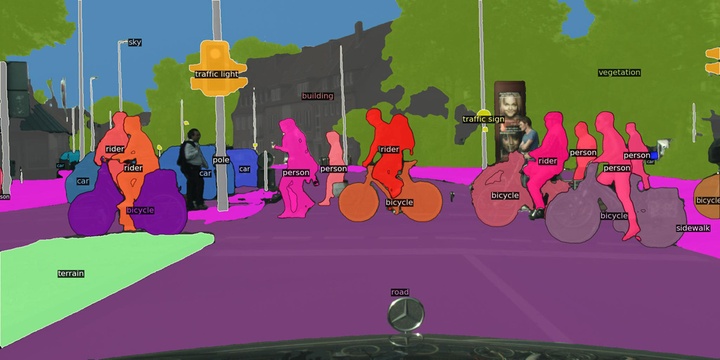}
\\

\includegraphics[width=0.28\linewidth]{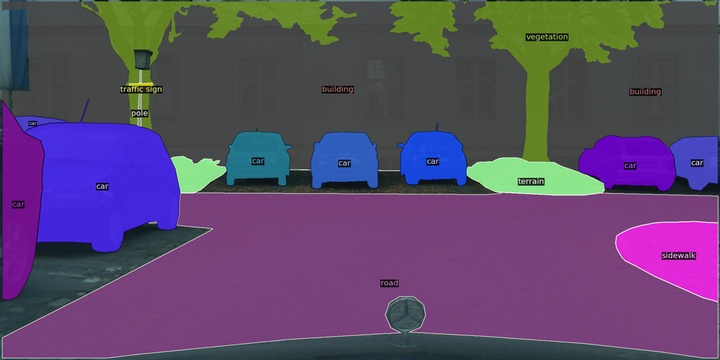}
\includegraphics[width=0.28\linewidth]{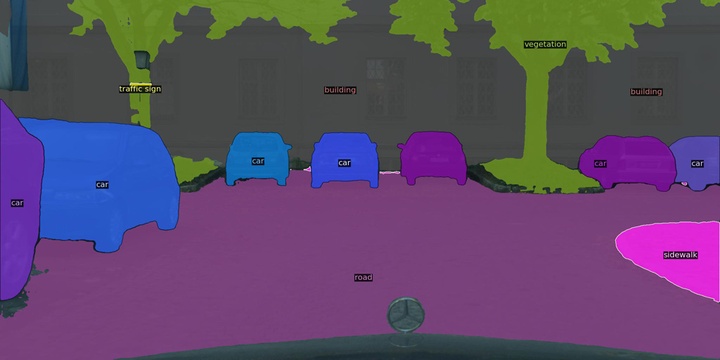}
\includegraphics[width=0.28\linewidth]{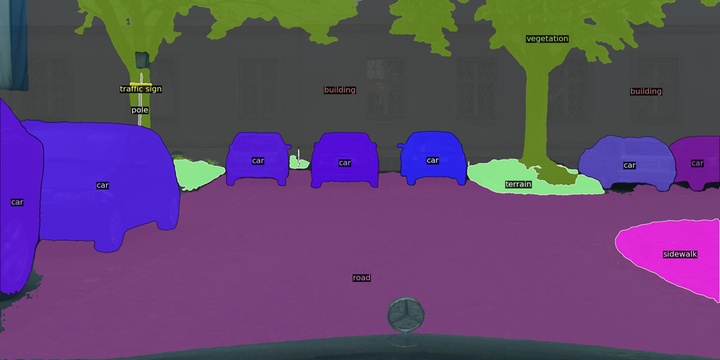}
\\

\includegraphics[width=0.28\linewidth]{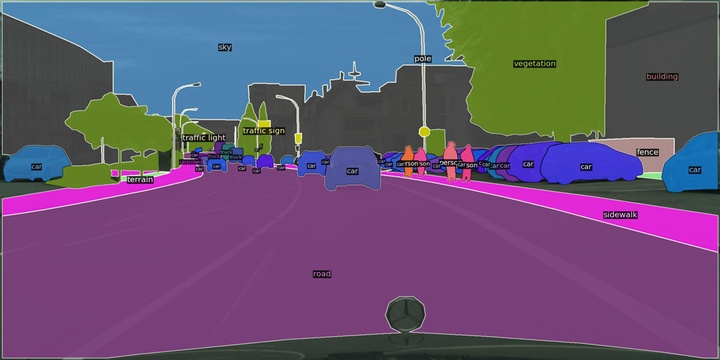}
\includegraphics[width=0.28\linewidth]{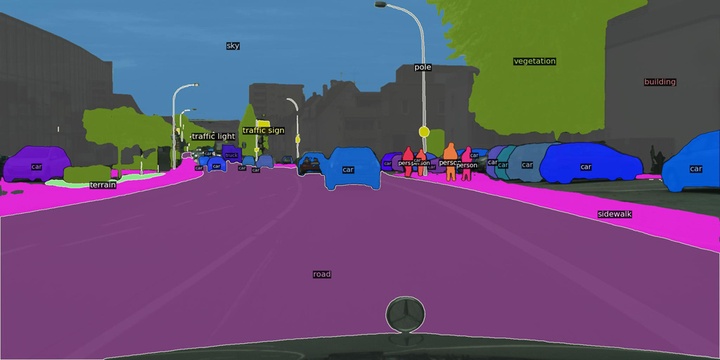}
\includegraphics[width=0.28\linewidth]{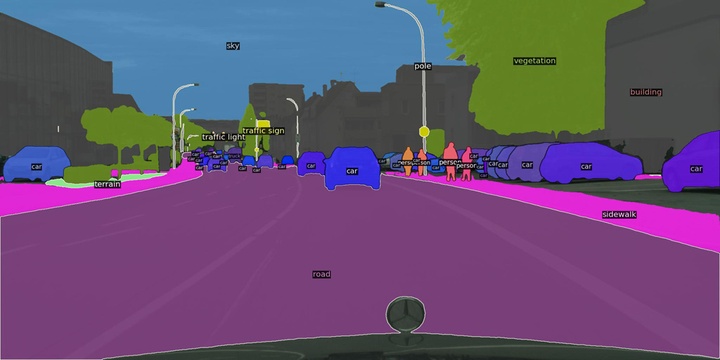}
\\

\includegraphics[width=0.28\linewidth]{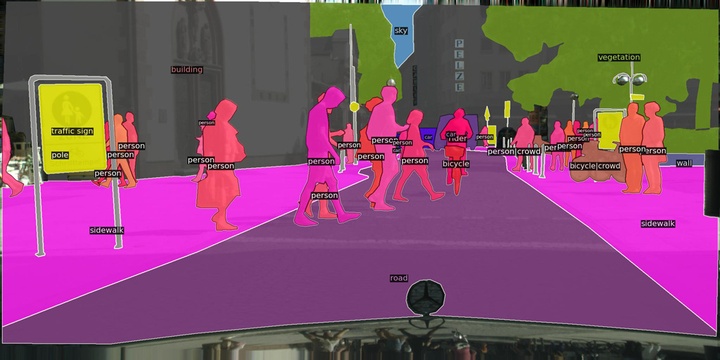}
\includegraphics[width=0.28\linewidth]{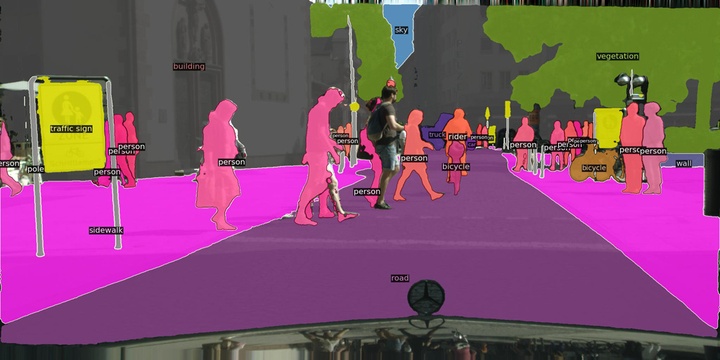}
\includegraphics[width=0.28\linewidth]{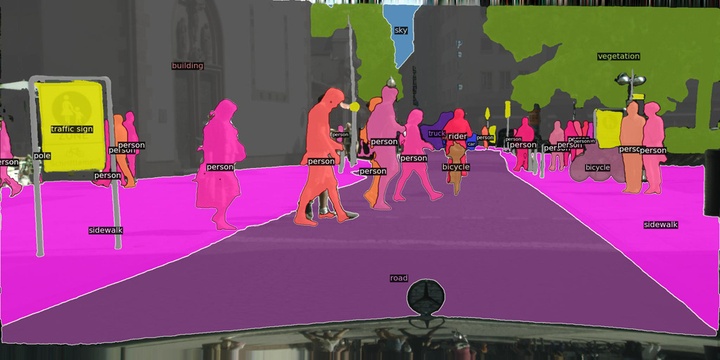}
\\

\includegraphics[width=0.28\linewidth]{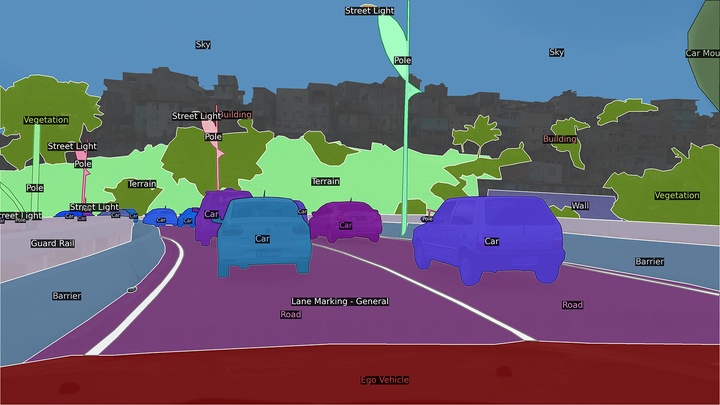}
\includegraphics[width=0.28\linewidth]{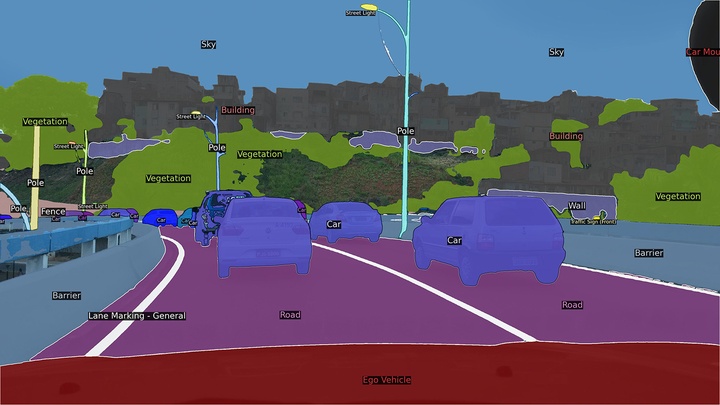}
\includegraphics[width=0.28\linewidth]{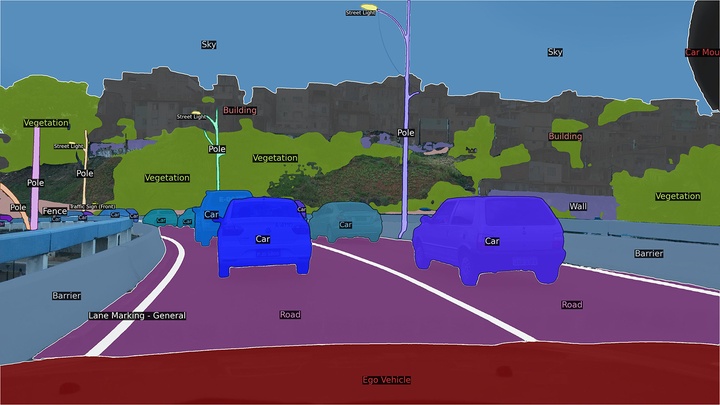}
\\

\includegraphics[width=0.28\linewidth, trim={0 0 0 4.4720cm},clip]{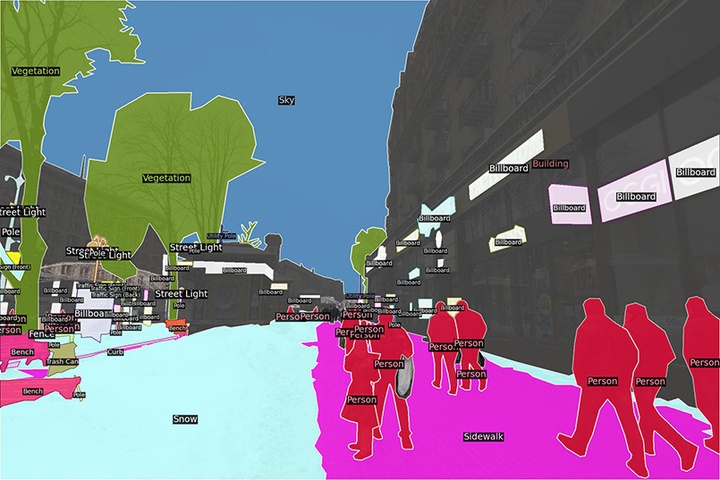}
\includegraphics[width=0.28\linewidth, trim={0 0 0 4.4720cm},clip]{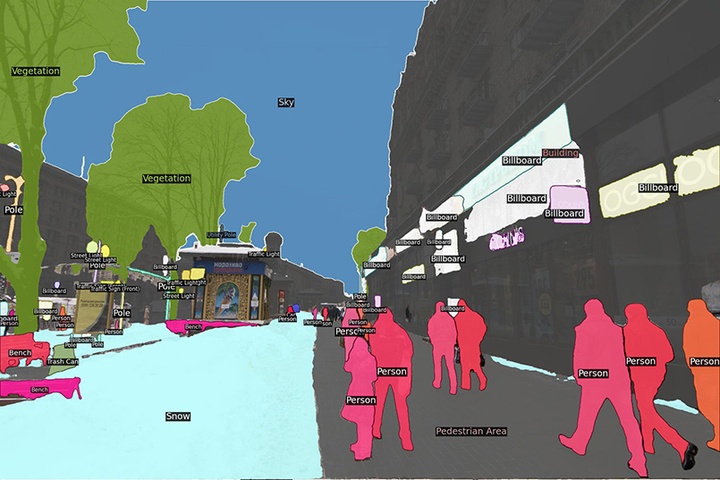}
\includegraphics[width=0.28\linewidth, trim={0 0 0 4.4720cm},clip]{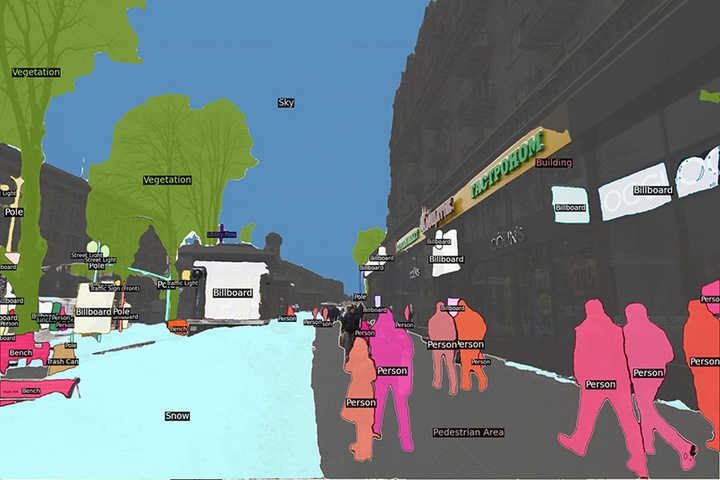}
\\

\includegraphics[width=0.28\linewidth]{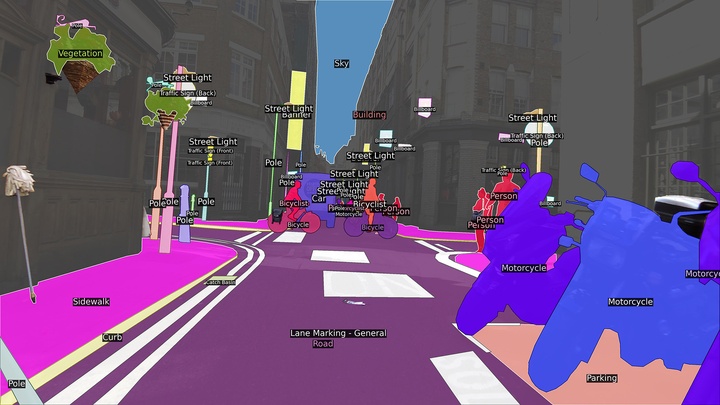}
\includegraphics[width=0.28\linewidth]{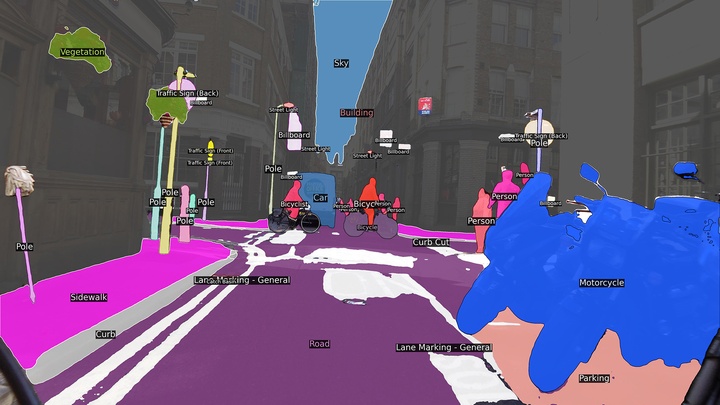}
\includegraphics[width=0.28\linewidth]{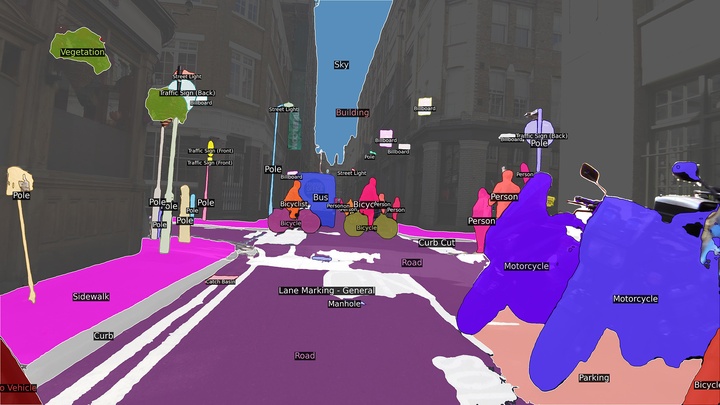}
\\

\includegraphics[width=0.28\linewidth, trim={0 0 0 5.5130cm},clip]{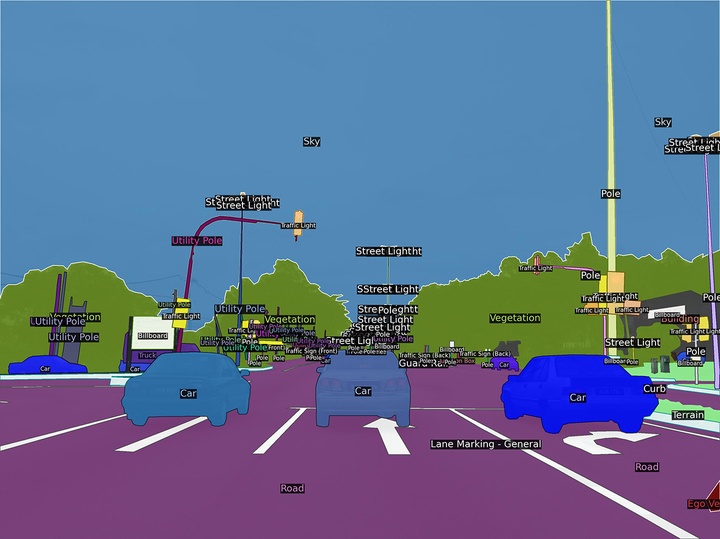}
\includegraphics[width=0.28\linewidth, trim={0 0 0 5.5130cm},clip]{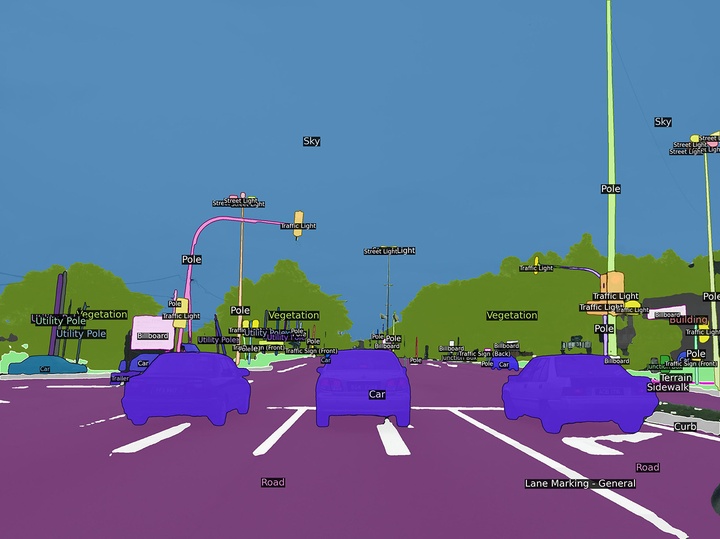}
\includegraphics[width=0.28\linewidth, trim={0 0 0 5.5130cm},clip]{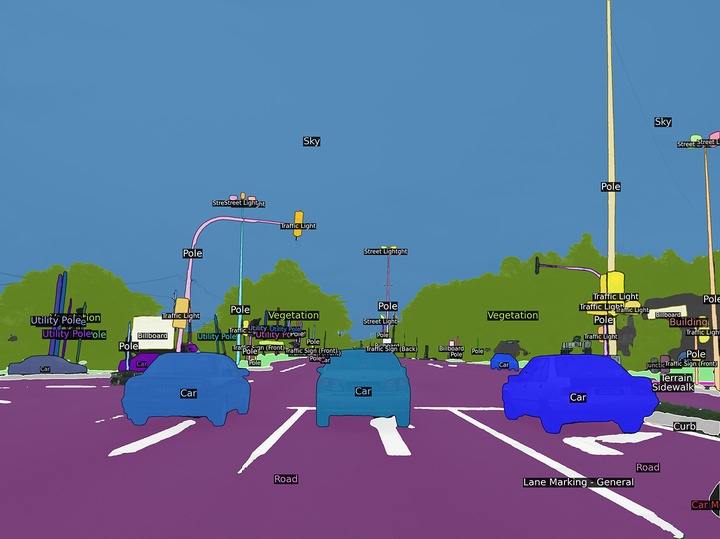}
\\

\begin{subfigure}[b]{0.28\textwidth}
 \centering
 \caption{Ground truth}
\end{subfigure}
\begin{subfigure}[b]{0.28\textwidth}
 \centering
 \caption{Without IBS}
\end{subfigure}
\begin{subfigure}[b]{0.28\textwidth}
 \centering
 \caption{With IBS (ours)}
\end{subfigure}

\vspace{-10pt}
\caption{\textbf{Intra-Batch Supervision on Mask2Former.} Top four images: Cityscapes \textit{val}; bottom four: Mapillary Vistas \textit{validation}. Each segment is indicated with a unique color and text label, so confusion can be observed when when multiple thing instances share a color or text label. Individual thing predictions for these images are shown in \Cref{fig:results_m2f_problem}. Best viewed digitally.}  
\label{fig:results_m2f_overall}
\end{figure*}

\begin{figure*}[t]
\centering
\includegraphics[width=0.28\linewidth]{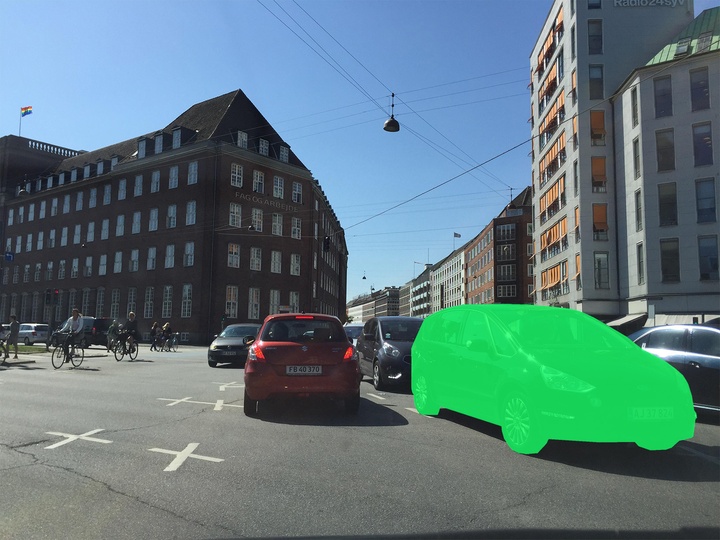}
\includegraphics[width=0.28\linewidth]{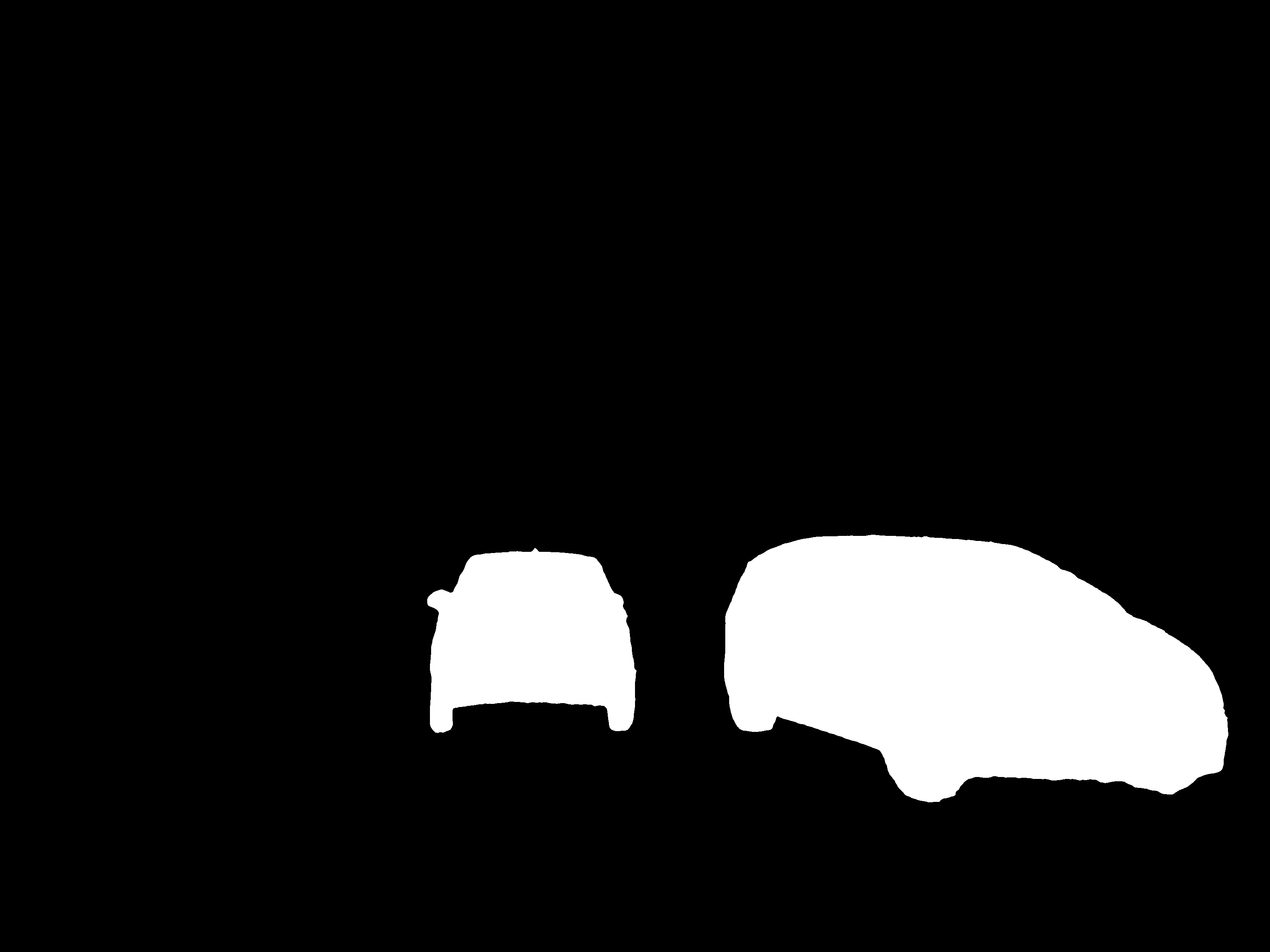}
\includegraphics[width=0.28\linewidth]{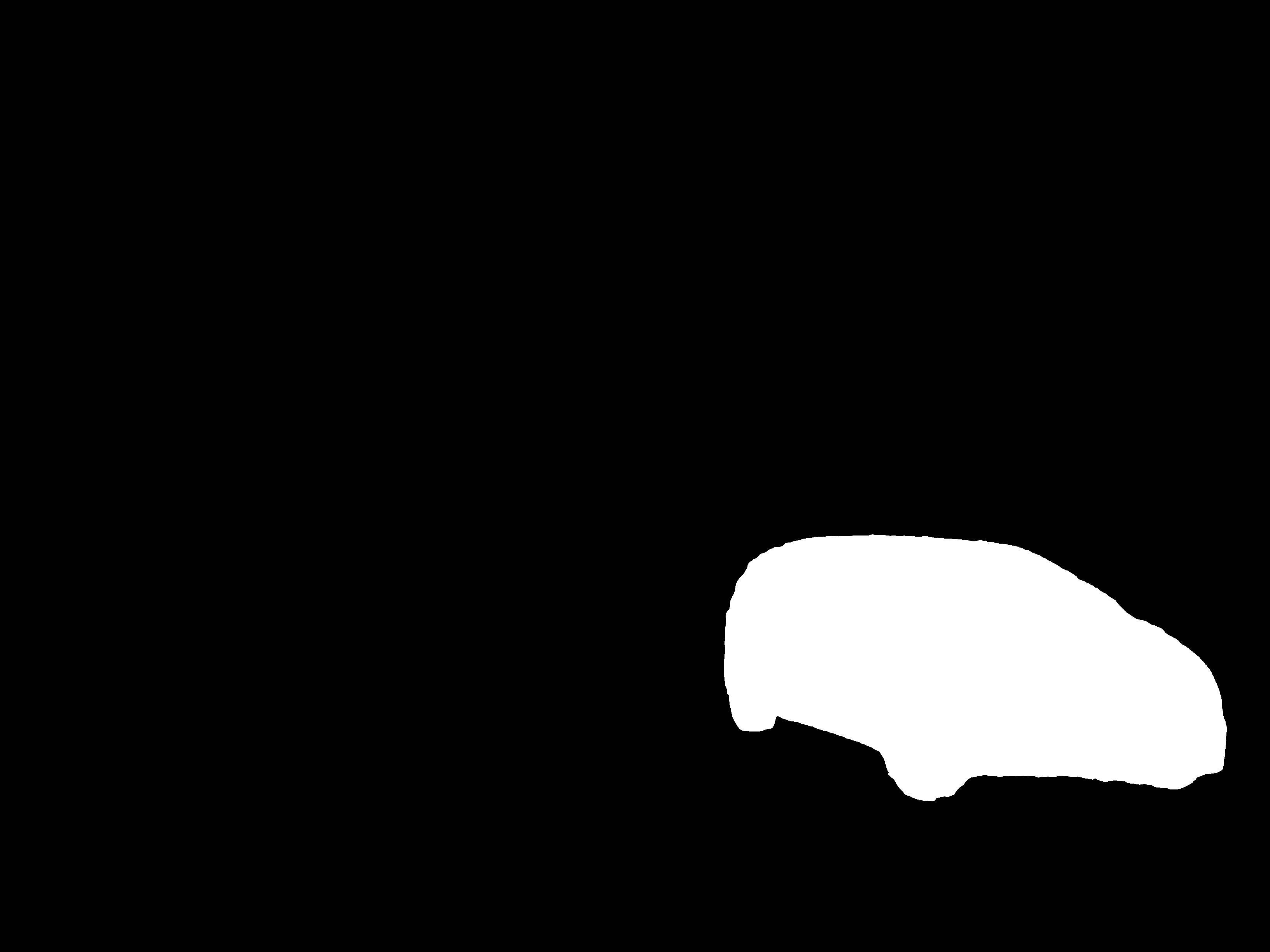}\\

\includegraphics[width=0.28\linewidth]{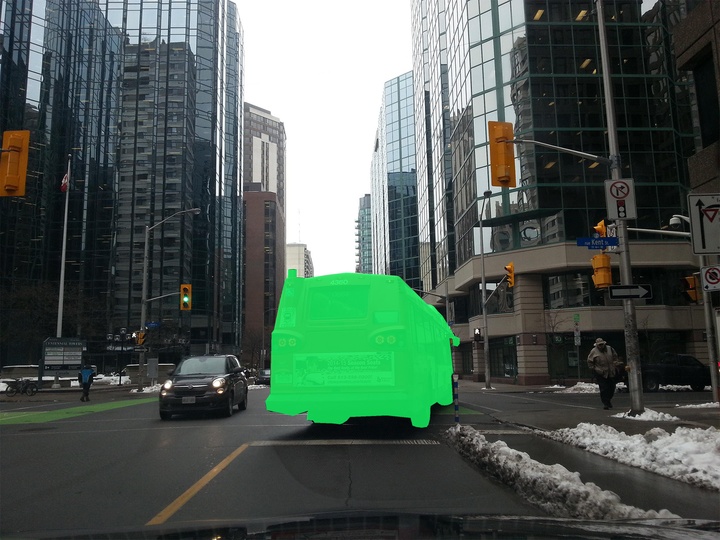}
\includegraphics[width=0.28\linewidth]{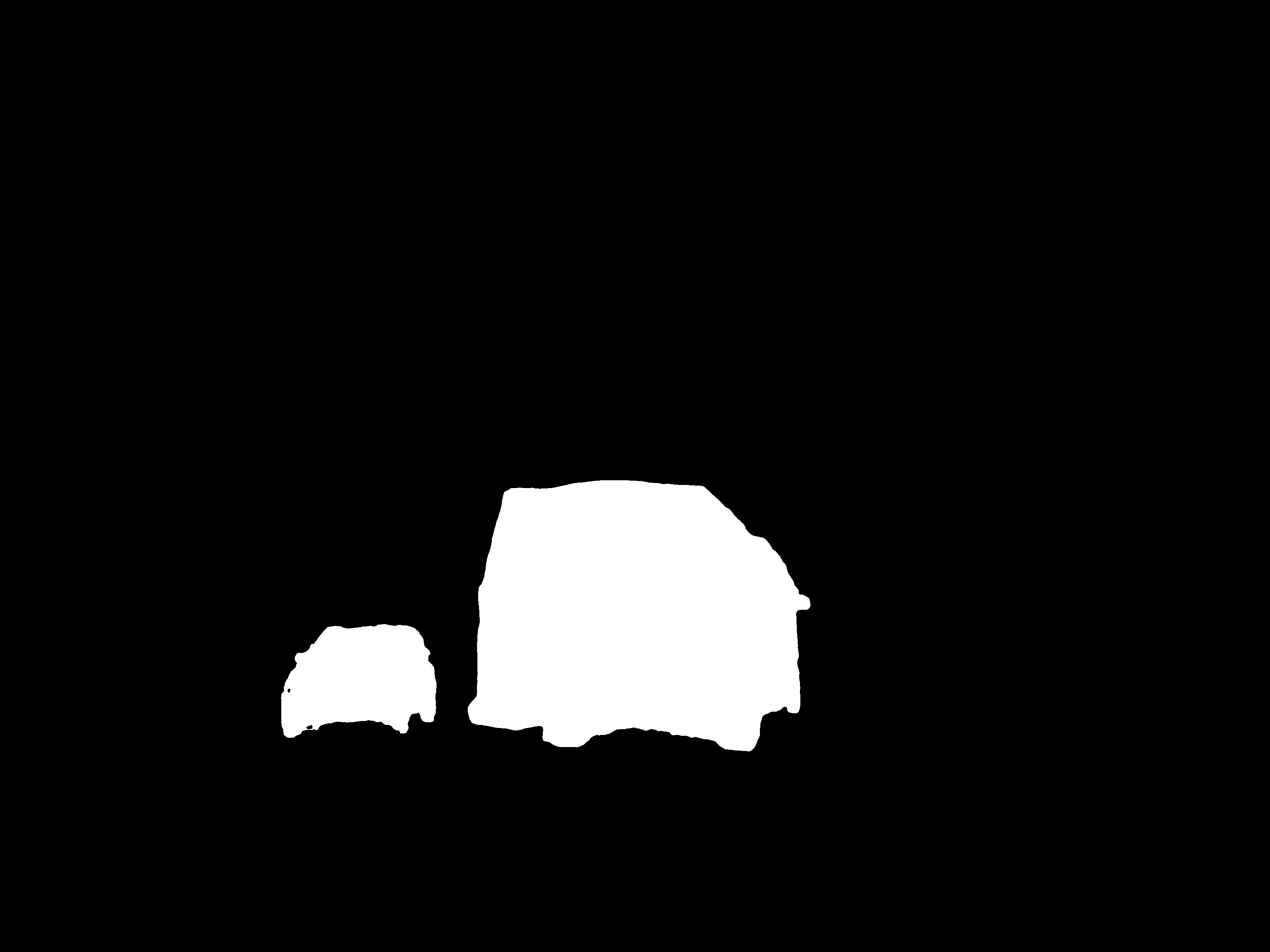}
\includegraphics[width=0.28\linewidth]{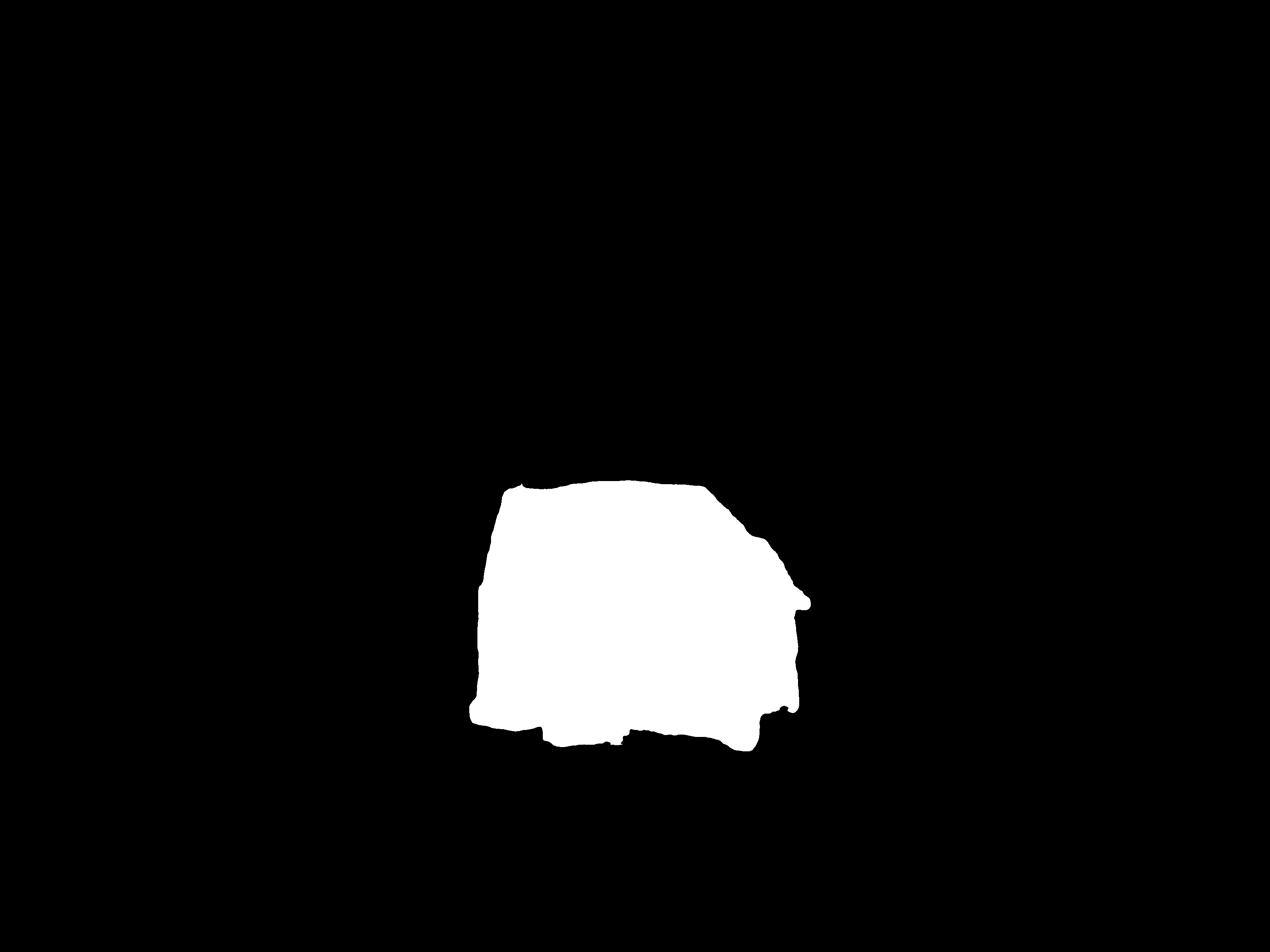}\\

\includegraphics[width=0.28\linewidth]{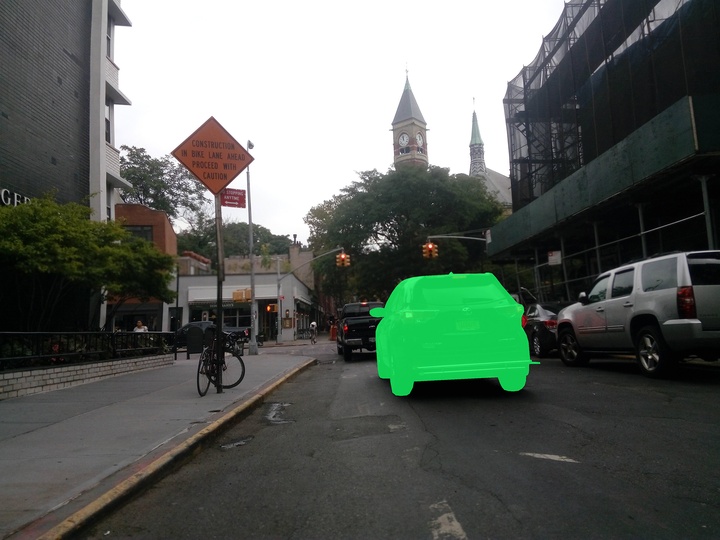}
\includegraphics[width=0.28\linewidth]{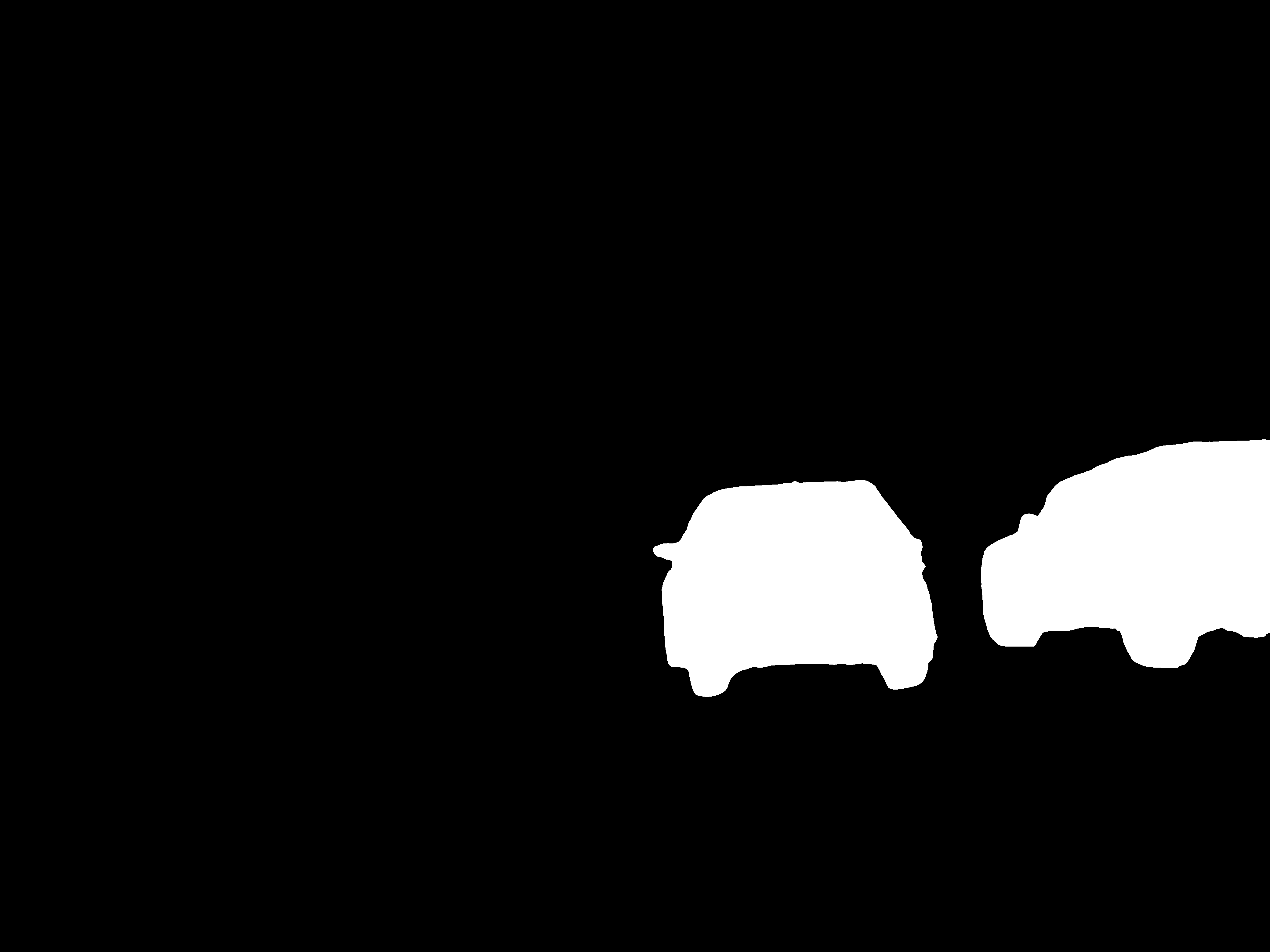}
\includegraphics[width=0.28\linewidth]{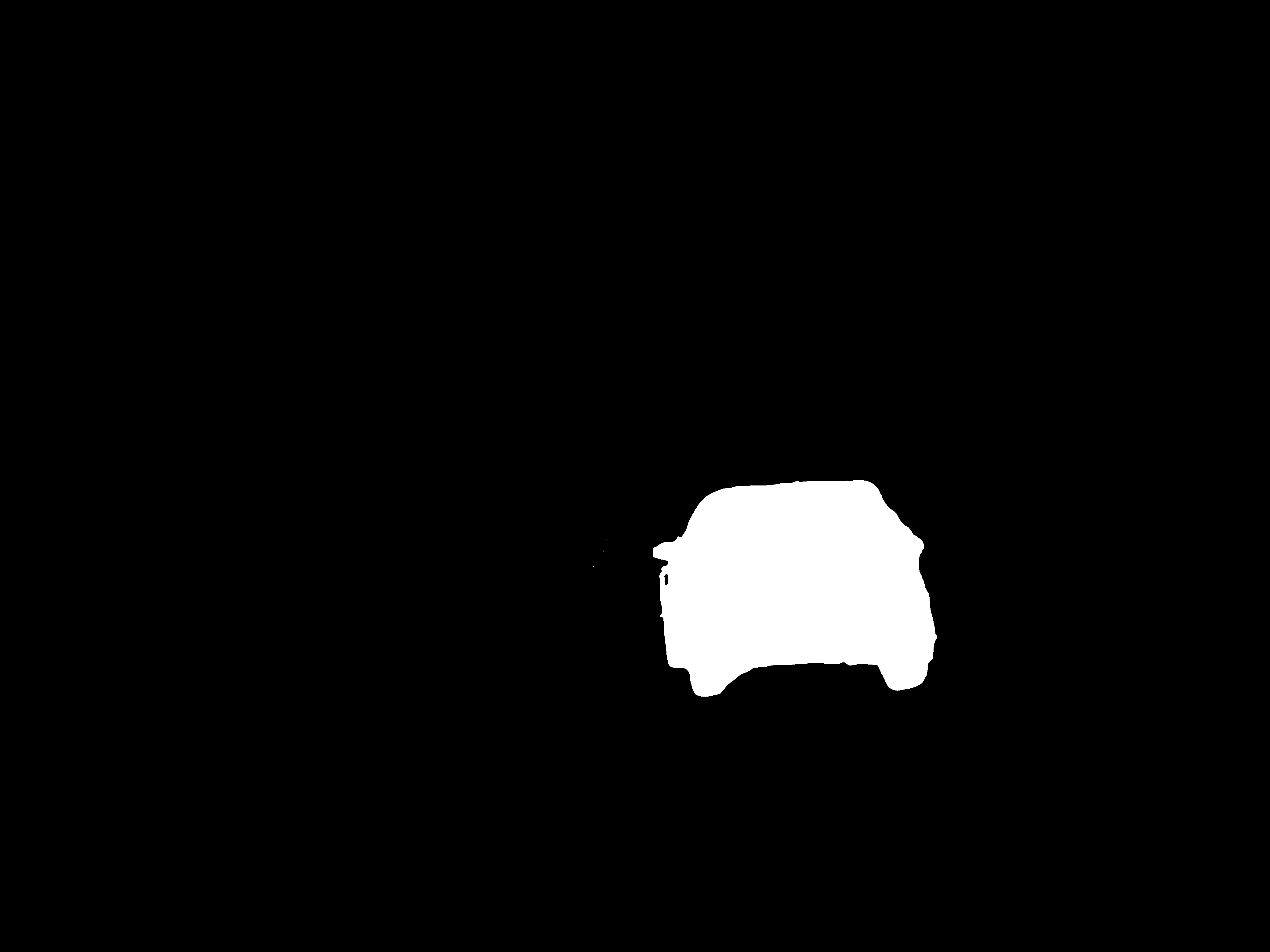}\\

\includegraphics[width=0.28\linewidth]{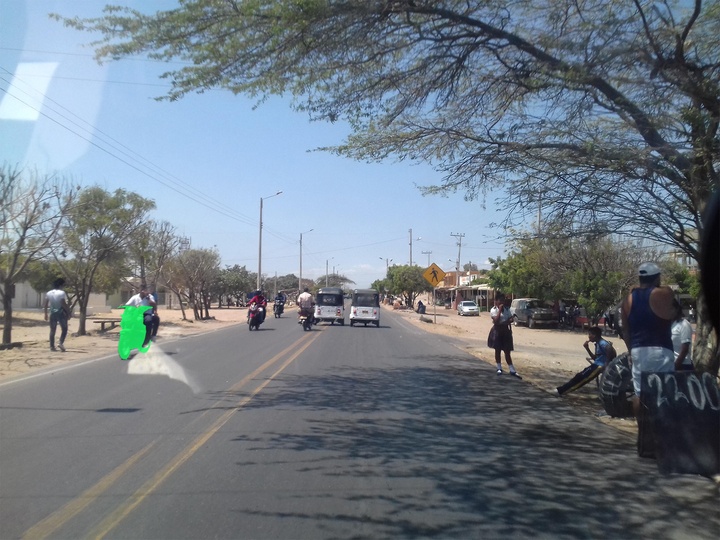}
\includegraphics[width=0.28\linewidth]{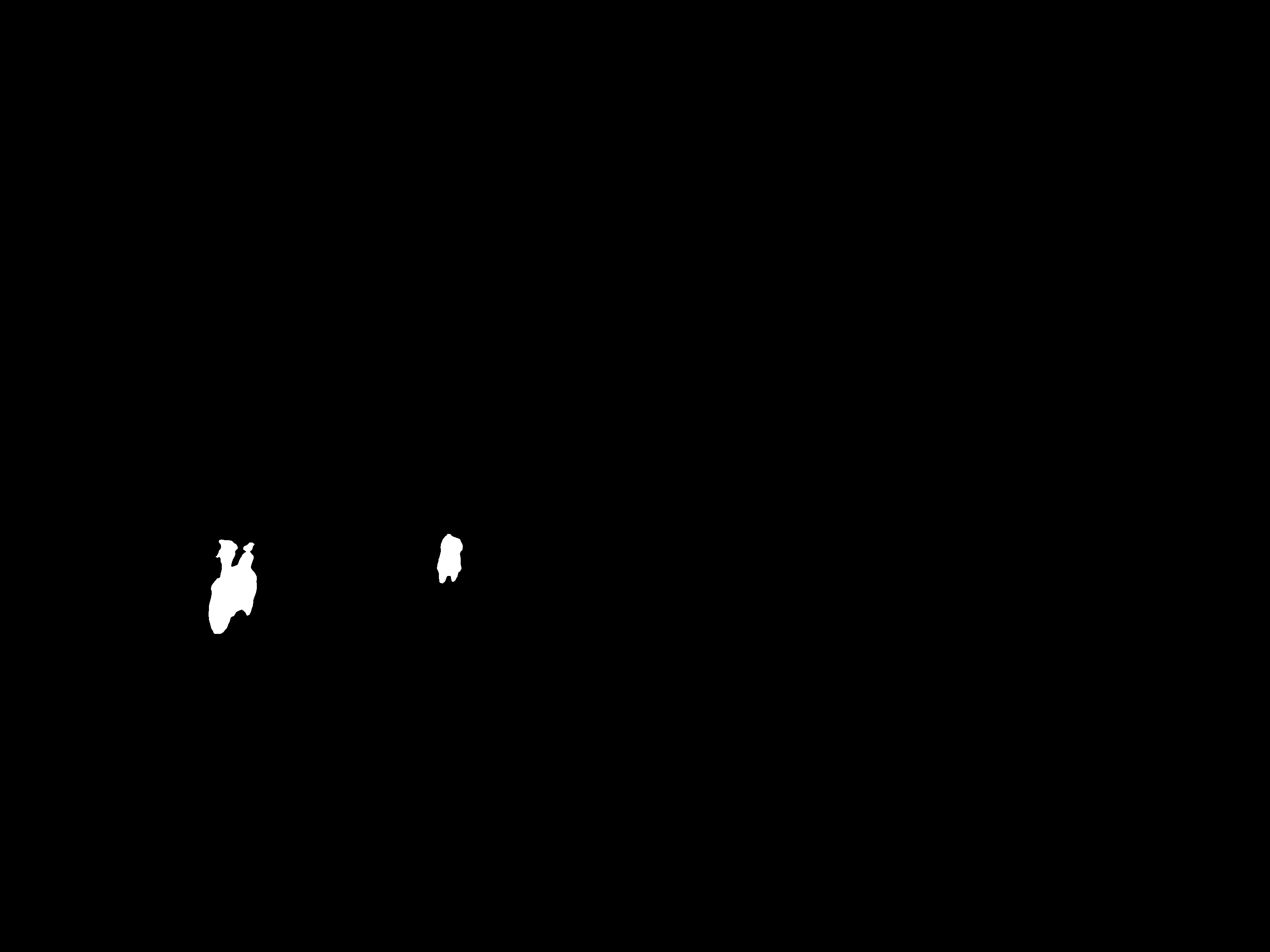}
\includegraphics[width=0.28\linewidth]{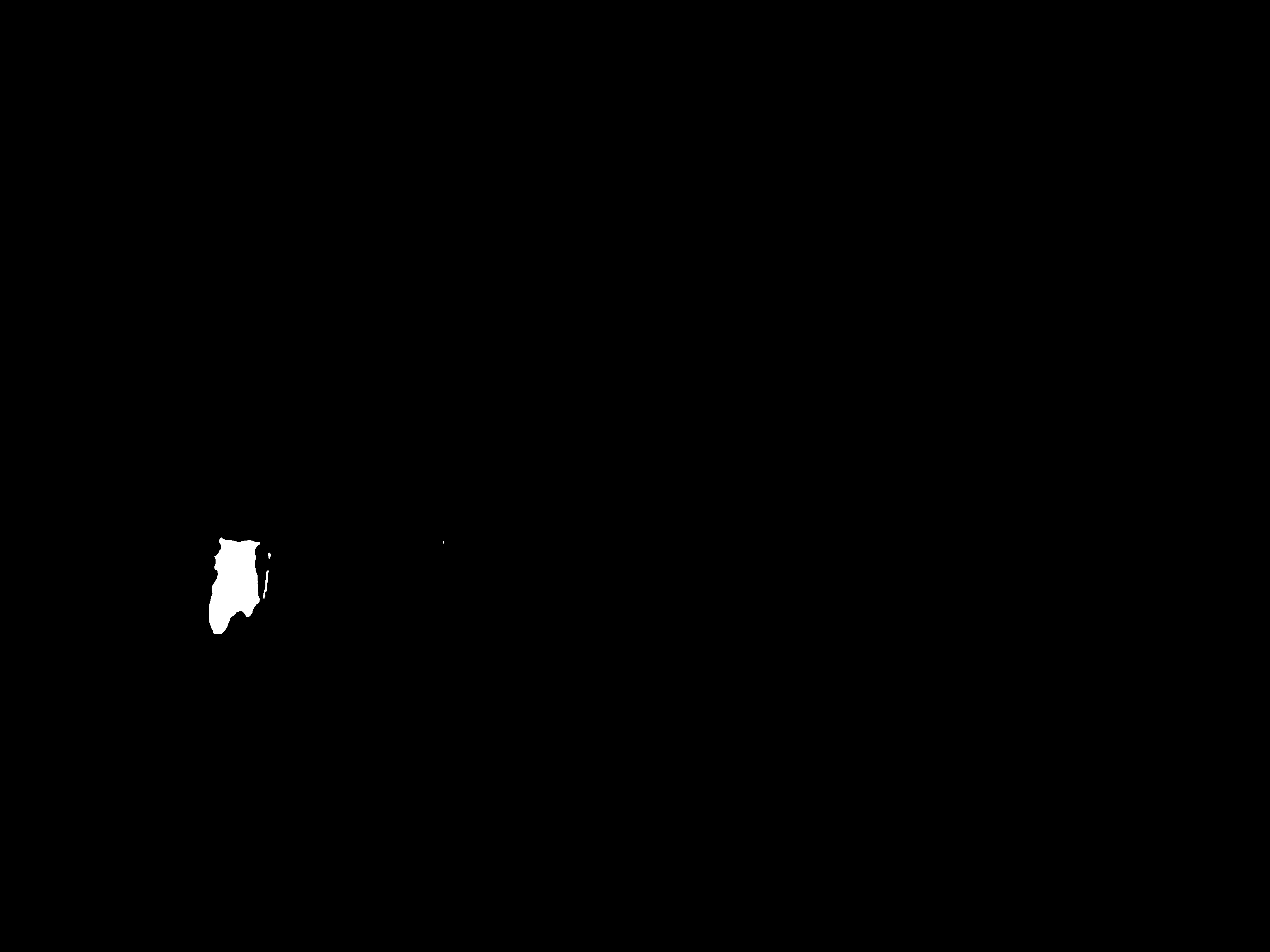}\\

 \begin{subfigure}[b]{0.28\textwidth}
     \centering
     \caption{Image with ground-truth segment}
 \end{subfigure}
 \begin{subfigure}[b]{0.28\textwidth}
     \centering
     \caption{Predicted segment \textbf{without IBS}}
 \end{subfigure}
  \begin{subfigure}[b]{0.28\textwidth}
     \centering
     \caption{Predicted segment \textbf{with IBS} (ours)}
 \end{subfigure}

\vspace{-10pt}
\caption{\textbf{Confusion problem for crop-based training of Mask2Former for instance segmentation.} Predictions for individual thing instances with and without IBS, on the Mapillary Vistas \textit{validation} set. (b) The predictions by Mask2Former without IBS suffer from confusion, and (c) IBS largely solves this problem, leading to more accurate predictions. Full instance segmentation results for these images are shown in \Cref{fig:results_m2f_inst_seg_overall}.}
\label{fig:results_m2f_inst_seg_problem}
\end{figure*}

\begin{figure*}[t]
\centering
\includegraphics[width=0.28\linewidth]{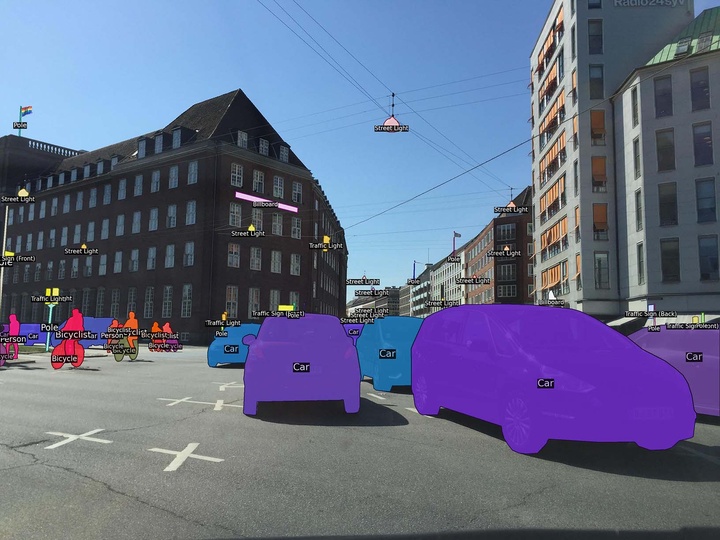}
\includegraphics[width=0.28\linewidth]{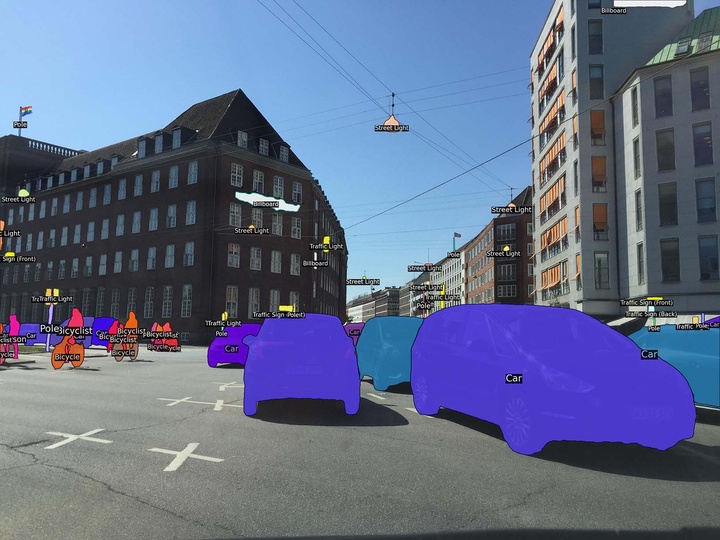}
\includegraphics[width=0.28\linewidth]{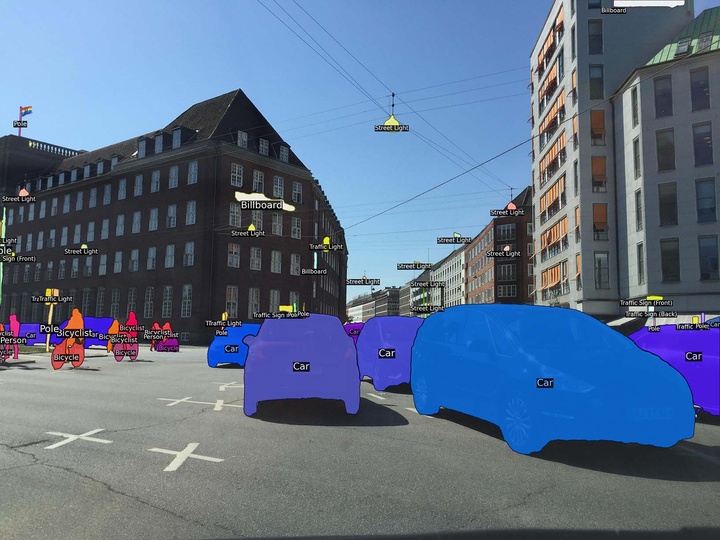}
\\

\includegraphics[width=0.28\linewidth]{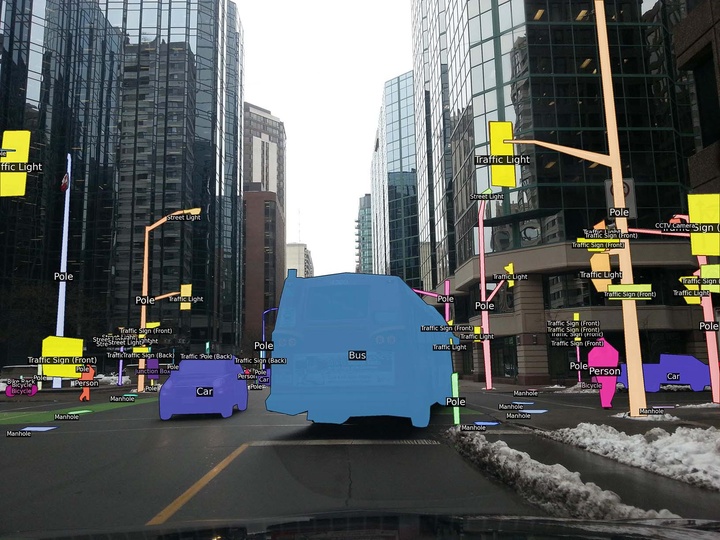}
\includegraphics[width=0.28\linewidth]{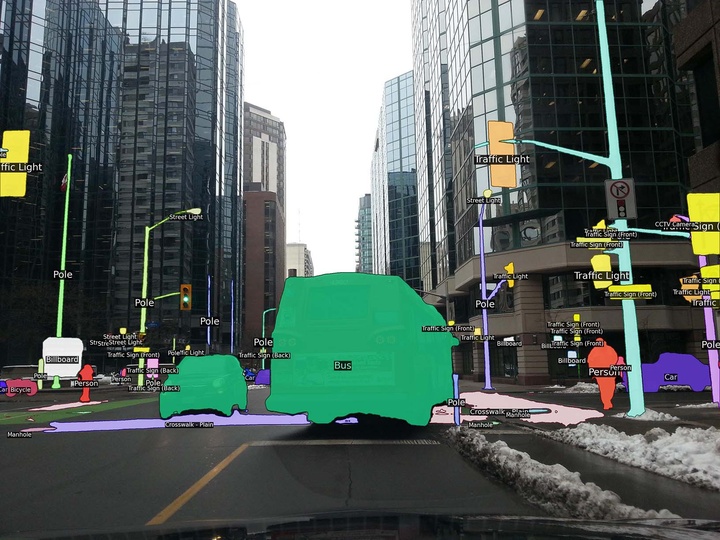}
\includegraphics[width=0.28\linewidth]{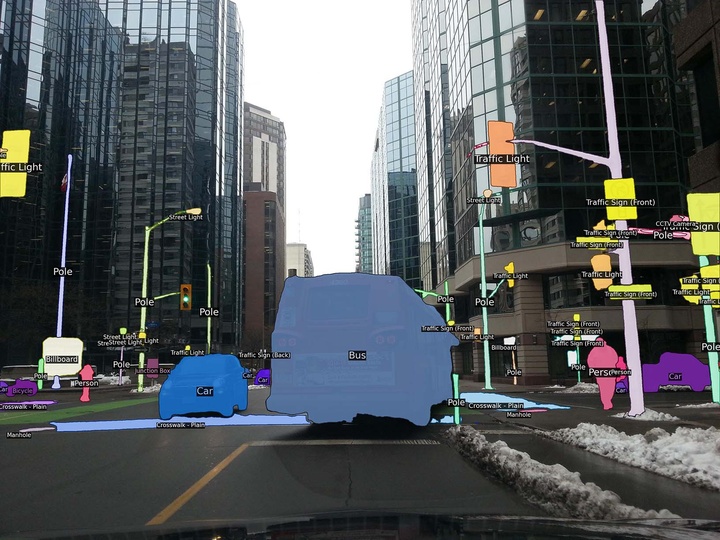}
\\

\includegraphics[width=0.28\linewidth]{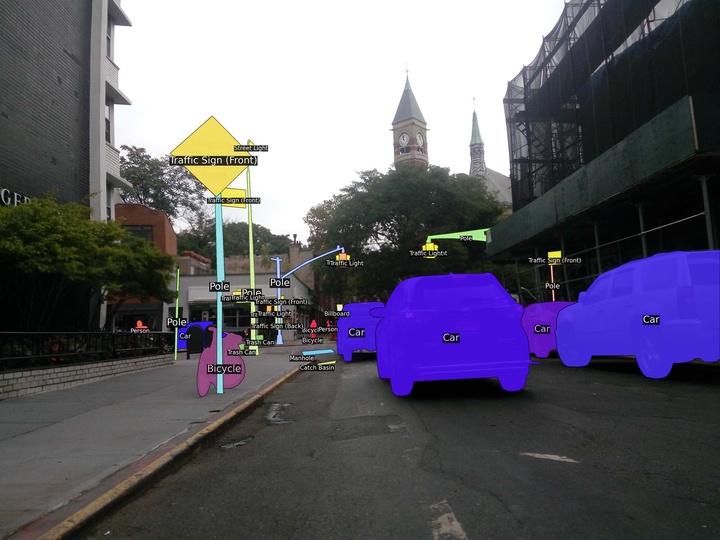}
\includegraphics[width=0.28\linewidth]{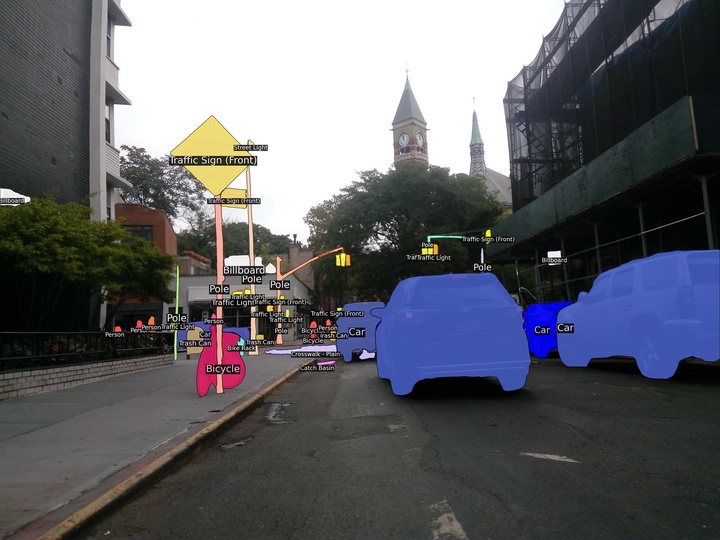}
\includegraphics[width=0.28\linewidth]{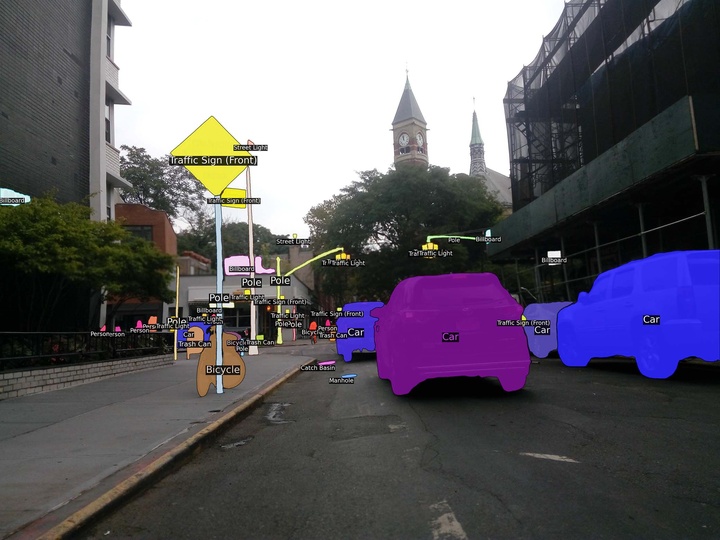}
\\

\includegraphics[width=0.28\linewidth]{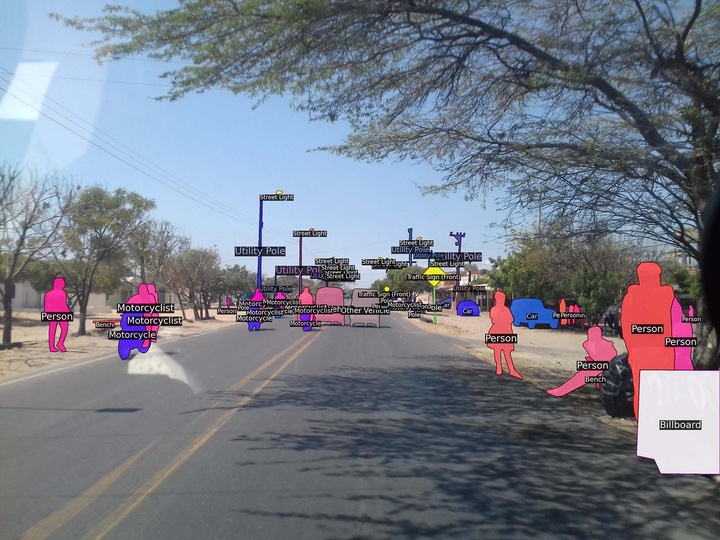}
\includegraphics[width=0.28\linewidth]{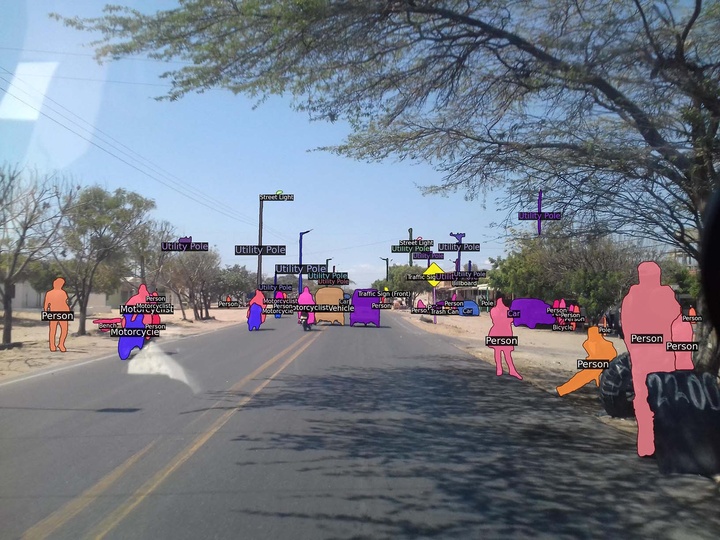}
\includegraphics[width=0.28\linewidth]{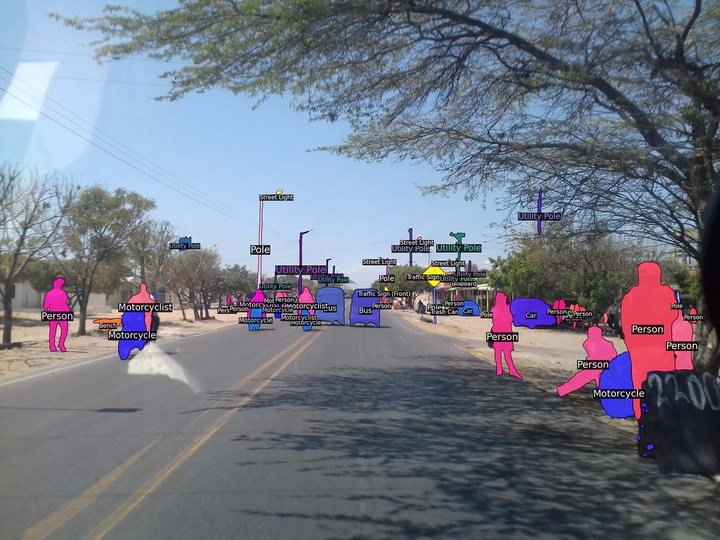}
\\

\begin{subfigure}[b]{0.28\textwidth}
 \centering
 \caption{Ground truth}
\end{subfigure}
\begin{subfigure}[b]{0.28\textwidth}
 \centering
 \caption{Without IBS}
\end{subfigure}
\begin{subfigure}[b]{0.28\textwidth}
 \centering
 \caption{With IBS (ours)}
\end{subfigure}

\vspace{-10pt}

\caption{\textbf{Intra-Batch Supervision on Mask2Former for instance segmentation.} Images from the Mapillary Vistas \textit{validation} set. Each instance is indicated with a unique color and text label, so confusion can be observed when when multiple instances share a color or text label. Individual thing predictions for these images are shown in \Cref{fig:results_m2f_inst_seg_problem}. Best viewed digitally.}  
\label{fig:results_m2f_inst_seg_overall}
\end{figure*}

To demonstrate the specific confusion problem, we show several examples of individual thing predictions by Panoptic FCN and Mask2Former in \Cref{fig:results_panfcn_problem,fig:results_m2f_problem,fig:results_m2f_inst_seg_problem}. In the predictions made by the networks \textit{without IBS}, it is clearly visible that the masks overlap multiple ground-truth thing instances -- which is what we call confusion -- and that this confusion mostly occurs between instances of the same class. In these figures, we also demonstrate that IBS solves this confusion problem to a great extent, resulting in much more accurate thing predictions. In~\Cref{fig:results_panfcn_overall,fig:results_m2f_overall,fig:results_m2f_inst_seg_overall}, we also provide the full panoptic and instance segmentation predictions for the same images and networks. This way, the impact of confusion on the overall result is visualized. Note that each instance should receive a unique color and text label in the visualized panoptic prediction, so a prediction in which two or more instances share a color and text label is a case of confusion. From these figures, it is also clear the networks \textit{with IBS} make considerably more accurate panoptic predictions, especially for large thing segments. Although there are still some small imperfections in the predicted masks, these predictions are considerably more suitable for downstream processing, as there are fewer grouped or missed objects.

\end{document}